\documentclass{bmvc2k}

\usepackage{tabularx,booktabs}
\usepackage{multirow}
\usepackage{lipsum}
\usepackage{graphicx}
\usepackage{color}
\usepackage{subfigure}
\usepackage{amssymb}
\usepackage{amsmath}
\usepackage{mathtools}
\usepackage{svg}
\usepackage{hyperref}

\usepackage{comment}

\title{An attention-driven hierarchical multi-scale representation for visual recognition}
\addauthor{Zachary Wharton}{zachary.wharton@go.edgehill.ac.uk}{0}
\addauthor{Ardhendu Behera}{https://computing.edgehill.ac.uk/~abehera/}{0}
\addauthor{Asish Bera}{beraa@edgehill.ac.uk}{0}
\addinstitution{
 Department of Computer Science\\
 Edge Hill University\\
 Ormskirk, Lancashire, UK
}

\runninghead{Wharton, Behera, Bera}{Attention-driven Hierarchical Multi-scale}

\begin{document}

\maketitle
\begin{abstract}
Convolutional Neural Networks (CNNs) have revolutionized the understanding of visual 
content. 
This is mainly due to their ability to break down an image into smaller pieces, extract multi-scale localized features and compose them to construct highly expressive representations for decision making. However, the convolution operation is unable to capture long-range dependencies such as arbitrary relations between pixels since it operates on a fixed-size window. Therefore, it may not be suitable for discriminating subtle changes (e.g. fine-grained visual recognition). To this end, our proposed method captures the high-level long-range dependencies by exploring Graph Convolutional Networks (GCNs), which aggregate information by establishing relationships among multi-scale hierarchical regions. These regions consist of smaller (closer look) to larger (far look), and the dependency between regions is modeled by an innovative attention-driven message propagation, guided by the graph structure to emphasize the neighborhoods of a given region. Our approach is simple yet extremely effective in solving both the fine-grained and generic visual classification problems. It outperforms the state-of-the-arts with a significant margin on three and is very competitive on other two datasets. 
\end{abstract}
\section{Introduction}
\label{sec:intro}
%
Convolution operation is the lifeblood of modern Convolutional Neural Networks (CNNs) used in computer vision to advance the popular tasks such as image/video recognition, object detection, visual-question answering, 
video forecasting, 
image-to-image translation and many more \cite{khan2020survey,khan2021transformers}. It has brought a significant breakthrough in most of these topics and is mainly due to its ability to effectively capture hidden patterns in the Euclidean space since an image can be represented as a regular grid. As a result, it captures meaningful local features by exploiting the shift-invariance and local connectivity. However, a well-known drawback of the convolution operation is, its inability to capture long-range dependencies between pixels in the image space as it operates on a fixed-size window. This is inappropriate in discriminating subtle variations in images, especially in fine-grained visual classification (FGVC) involving in recognition of different species of animals, various car/aeroplane models, different kinds of retail products, etc. To address this, researchers have explored informative object parts/regions \cite{bera2021attend, zhang2016weakly, yao2017autobd, he2017fine, ge2019weakly}, attention mechanisms to identify these salient regions \cite {behera2021context, liu2019bidirectional, liu2016fully, zheng2019looking}, local and non-local operations \cite{hu2019local, wang2018non} to capture discriminative information. However, the above approaches still focus on the Euclidean space while modeling/exploring/identifying salient regions/parts and aggregating interactions/similarities to discriminate subtle variations. In this work, we advance this approach by taking advantage of Graph Convolutional Networks (GCN), which collectively aggregate key information from graph structure by modeling long-range dependencies in a non-Euclidean domain. 

\subsection{Motivation} To discriminate subtle variations in visual features, many recent works have focused on both single-scale and multi-scales object parts and regions \cite{behera2021context, shroff2020focus, yang2021re, he2019and, eshratifar2019coarse2fine}. These are mainly focused on coarse-to-fine exploration by jointly integrating feature representation of regions at different scales. These have notably enhanced the recognition accuracy. However, for a better representation of visual-spatial structural relationships among regions, the hierarchical connection between regions should be considered so that the larger regions (see from far) pay more attention to the high-level shape and appearance, and the smaller ones (closer look) concentrate on detailed texture and parts information 
to capture subtle variations. As a result, 
it can provide a rich representation by jointly learning meaningful complementary information from multi-scale hierarchical regions that is applicable to both FGVC (coarse-to-fine) and generic visual classification (fine-to-coarse). We achieve this by a novel multi-scale hierarchical representation learning to boost the recognition accuracy by jointly integrating local (within a region), and non-local (between regions) information to capture long-range dependencies by exploring the graph structure to propagate information between regions within a layer, as well as between layers in the hierarchical structure. In the above SotA methods, the joint feature representation between regions is learned as a part of features. Whereas, we learn this using a graph structure-guided information propagation between regions represented as graph nodes. 
The information propagation is emphasized by learning the ``importance'' of neighboring regions of a given region. Furthermore, our approach does not require objects or parts bounding-box annotations. 

\subsection{Contributions} Our main contributions are: 1) An innovative multi-scale hierarchical representation learning is proposed for improving visual recognition. 
2) A more abstract and coarser representation of multiple graphs denoting the hierarchical structures is considered via spectral clustering-based regions aggregation. 3) A  novel  gated  attention mechanism is proposed to aggregate the cluster-level class-specific confidence. 
4) Analysis of the model on five datasets consisting of FGVC and generic visual classification, obtaining competitive results. 5) Visual analysis 
and ablation studies justify the effectiveness of our model.
\section{Related work}
Our work is the conflation of attention mechanism, multi-scale hierarchical representation, and GCN to improve visual recognition. A precise study on these key aspects is presented. 
\subsection{Generic visual recognition} CNNs have remarkably enhanced the performance of large-scale and generic image classification \cite{he2019rethinking, chollet2017xception,simonyan2014very}. 
Recently, wider and deeper networks with learned data augmentation strategies are used for performance gain \cite{cubuk2019autoaugment}. Researchers have also explored Knowledge distillation (KD) for visual classification to avoid complex and cumbersome models involving heavy computational overload and memory requirements \cite{zhang2019your, yun2020regularizing}. 
A group of small student networks jointly learn by optimizing three loss functions \cite{zhang2019your}. Similarly, the class-wise self-KD 
regularizes the dark knowledge and generates effective predictions 
for generic image classification and FGVC \cite{yun2020regularizing}.  We address the diverse visual recognition task by exploring hierarchical relations by modeling long-range dependencies among multi-scale regions that self-guides the network to learn jointly. 

\subsection{Hierarchical and Multi-scale methods for FGVC} Weakly supervised FGVC methods are widely investigated to enhance the model's discriminative capability by exploring the hierarchical and multi-scale structures in images. These include but not limited to 
hierarchical pooling \cite{ding2021ap, cai2017higher, yu2018hierarchical, chen2018fine}, multi-scale representation \cite{fu2017look, yang2018learning, zheng2019learning} and context encoding \cite{behera2021context}.  
In \cite{cai2017higher}, higher-order intra- and inter-layer relations are exploited to integrate hierarchical convolutional features 
from various semantic parts at multiple scales. 
A cross-layer hierarchical bilinear pooling is explored in \cite{yu2018hierarchical} to learn complementary information 
from multiple intermediate convolutional layers. Similarly, a hierarchical semantic embedding is proposed in \cite{chen2018fine} to model structured correlation 
at various levels in the hierarchy. 
In \cite{ding2021ap}, a spatial pyramidal hierarchy-driven top-down and bottom-up attentions is described to combine high-level semantics and low-level features. This multi-level 
feature descriptor aims to discriminate subtle variances. A multi-agent cooperative learning (Navigator-Teacher-Scrutinizer) is aimed to locate informative regions at multi-scale via self-supervision \cite{yang2018learning}. 
These discriminative regions at various scales and ratios 
are used to generate final feature map for the FGVC. Similarly, the progressive-attention CNN locates object-parts at multi-scale, and learns a rich part hierarchy via generating multiple local attention maps \cite{zheng2019learning}. 
These maps are further refined and optimized in a mutual reinforcement way. Likewise, recurrent attentional CNN recursively locates informative regions from coarse-to-fine and learns region-based feature representation at multiple scales 
\cite{fu2017look}. Inspired by these progress, we establish an attention mechanism to capture dependencies between multi-scale regions hierarchically to learn discriminative feature representation at multiple granularity to model subtle changes.

\subsection{Graph Convolutional Networks (GCN)} 
 Motivated by CNN, there is a growing interest to learn high-level representation using GCN in non-Euclidean space \cite{wu2020comprehensive}. 
 GCN in conjunction with attention mechanism has proven its effectiveness in modeling long-range dependency. 
 Focusing on the node's most relevant hidden representations in a graph, a 
 multi-head-attention \cite{vaswani2017attention} is exploited to stabilize the learning task, leveraging the second-order hierarchical pooling 
 \cite{wang2020second}.  
In \cite{lee2019self}, a self-attention is introduced for hierarchical pooling to learn structural information using the node features and graph topology.   
 Similarly, a multi-layer differentiable pooling is proposed to learn hierarchical structures of graphs \cite{ying2018hierarchical}. 
 Recently, spectral clustering is adapted in \cite{bianchi2020spectral} for node clustering to improve the accuracy. A graph-based relation discovery (GRD) approach that applies a collaborative learning strategy is proposed in 
 \cite{zhao2021graph} to learn higher-order feature relationships to enhance the FGVC accuracy. 
 Similarly, GCN is also adapted to advance FGVC by learning the category-specific semantic correspondence using the latent attributes between image-regions \cite{wang2020category}. Lately, a criss-cross graph propagation sub-network is built to find out the correlation between discriminative regions \cite{wang2020graph} in solving FGVC. However, both methods compute a fewer regions per image (e.g. 4), which may render a sub-optimal solution. To address this, we propose a multi-scale hierarchical representation learning, which is guided by attention and graph structure to enhance image recognition task through weighted message aggregation. 
\begin{figure}[t]
\centering
\includegraphics[width=0.95\textwidth] {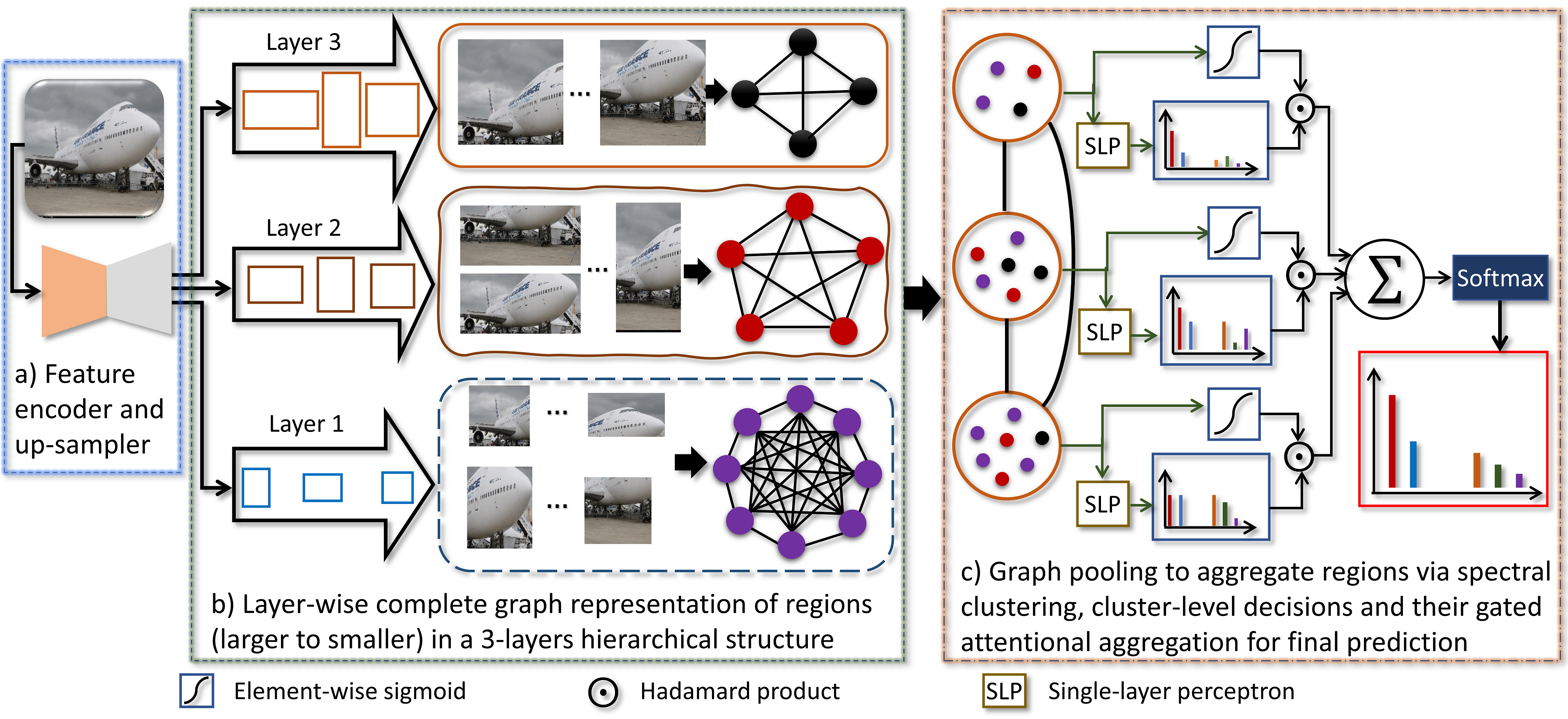} \hspace{-0.3cm}
\vspace{-.3cm}
\caption{a) CNN backbone as an encoder followed by an up-sampler to increase spatial resolution. b) Layer-wise graph representation using multi-scale regions to capture long-range dependencies using GCN's aggregation scheme in which the feature of a node is computed by recursively aggregating and transforming features from its neighboring nodes. c) Graph coarsening 
by grouping the similar nodes using spectral clustering, followed by cluster-level predictions and their gated attentional aggregation for making a final decision.} 
\label{fig:full_model}
\end{figure}
\section{Proposed approach}
Our model's architecture 
is shown in Fig. \ref{fig:full_model}. It extracts high-level convolutional features using a CNN 
and then pools features from it by considering hierarchical (e.g. small $\rightarrow$ medium $\rightarrow$ large $\rightarrow$ full image) regions. These regions are linked using a graph structure, leveraging GCN to propagate information in addressing the shortcomings of prior region-based approaches 
to enhance the discriminability for visual classification.   
%
\subsection{Problem formulation}
A set of $N$ images $I=\{I_n|n=1\dots N\}$ and their respective class labels are given to train an image classifier. The aim is to learn the classifier's mapping function $\mathbf{F}$ that predicts $\hat{y}_n=\mathbf{F}\left(I_n\right)$ to match the true label $y_n$. During training, it learns $\mathbf{F}$ by minimizing a loss $\mathcal{L}\left(y_n,\hat{y}_n\right)$ between the true and the predicted label. In our case, the function $\mathbf{F}$ is an end-to-end deep network in which we introduce a simple yet effective modification to advance the visual recognition. It focuses on aggregating information from a given region by attending over its neighbors by exploring a novel attention mechanism. Thus, 
$\mathbf{F}$ consists of:\vspace{-0.3 cm}
\begin{equation}
    \mathbf{F} = \text{Softmax}\left(\sum_{l=1}^{L}\sum_{r=1}^{R_l}\phi_l(I_n(A_{r,l});\theta_l)\right),
    \label{eq_1}
    \vspace{- 0.3 cm}
\end{equation} 
where $l=1\dots L$ is the layers within the hierarchical structure, and $R_l$ is the number of regions in $l^{th}$ layer. $\phi_l$ measures the contribution of image-region $A_{r,l}$, representing $r^{th}$ region located in the $l^{th}$ layer i.e., $I_n(A_{r,l})$ is the hierarchical representation of image $I_n$. $\theta_l$ is the corresponding $l^{th}$ layer's parameter and is learned via end-to-end fashion as follows. 
%
\subsection{Hierarchical multi-scale regions} \label{sec:regions}
Our hierarchical multi-scale regions approach is motivated by recent region-based approaches \cite{zhang2016weakly, yao2017autobd, he2017fine, ge2019weakly, behera2020regional, behera2021context} in solving FGVC. However, it is different since we use smaller (look closer) to larger (look from far) regions in a hierarchical fashion. Whereas, the regions in \cite{zhang2016weakly, yao2017autobd, he2017fine, ge2019weakly} are considered as object proposals from a detector and the approaches in \cite{behera2020regional, behera2021context} use multi-scale regions by exploring HOG 
cells and blocks. Our regions within a layer in the hierarchy have the same area but with different aspect ratios (Fig. \ref{fig:full_model}b). For example, given width $w$ and height $h$, there are at least 3 different regions with the same area: 1) width = $w$ and height = $h$, 2) width = $h$ and height = $w$, and 3) width = height = $\sqrt{w\times h}$. The aim is to represent an image with a rich discriminative descriptor by considering multiple hierarchical overlapped regions that are not only applicable for advancing FGVC involving large inter-class similarities and great intra-class variations but also pertinent to solve the generic visual classification with distinctive categories (e.g. Caltech-256 \cite{griffin2007caltech}, CIFAR-100 \cite{krizhevsky2009learning}).
%
\subsection{Graph-based region's feature representation}
For a given image $I_n$, there are $L$ layers within the hierarchical structure. In each layer $l$ ($l=1\hdots L$), there are $R_l$ regions representing a complete graph, 
resulting in $L$ numbers of graphs ($G=\{G_1,G_2,\hdots ,G_L\}$) as shown in Fig. \ref{fig:full_model}b. A node in graph $G_l$ denotes a region $r=1\hdots R_l$ with feature $f_{r,l}$, resulting in $R_l$ nodes. The aim is to update $f_{r,l}$ of the region/node $r$ by considering a non-local operation, capturing long-range dependencies in the image space. This way, it is able to capture interactions between regions regardless of their locations in the image space. We achieve this operation by exploring the graph convolution that transforms feature $f_{r,l}$ to $f_{r,l}^{'}$, leveraging different weights to different neighborhood regions to address the drawbacks of the existing approaches in encoding long-range dependencies. We revisit the message passing algorithm in GCN \cite{kipf2016semi} to get a feature representation of a given node $i$ by considering its neighbors using a simple propagation rule i.e. $f_i=\sigma \left(\sum_j\frac{1}{c_{i,j}}f_j W\right)$, where $j$ represents neighboring nodes of $i$ and $c_{i,j}$ is a normalization constant for the edge ($i,j$), which originates from the symmetrically normalized adjacency matrix $D^{-\frac{1}{2}}\mathcal{M}D^{-\frac{1}{2}}$ of the GCN. $D$ is the diagonal node degree matrix of the graph adjacency matrix $\mathcal{M}$, $\sigma(.)$ is a nonlinear activation (e.g. Sigmoid or ReLU), and $W$ is a learnable weight matrix. GCN aggregates information from the neighboring nodes that neither ``remember'' important nodes nor ``emphasize'' them by giving higher weights throughout the learning. We address this shortcoming by specifying different weights to the neighborhood nodes via the attention mechanism \cite{BahdanauCB14, vaswani2017attention}. Specially, our method is inspired by the shared attention mechanism in \cite{velivckovic2017graph} to perform self-attention in graph nodes. For a given graph $G_l$, it is computed as:
\begin{equation}
\begin{aligned}
 \alpha_{r,r'} &= \frac{\text{exp}(e_{r,r'})}{\sum_{k \in R_l}\text{exp}(e_{r,k})}  ; 
  \text{ } e_{r,r'} = \text{LeakyReLU}\left(\mathbf{W_a}^T([Wf_{r}||Wf_{r'}])\right) ; \\ f_{r}^{'} &=  \overset{H}{\underset{h=1}{\Bigm\Vert}}\sigma\left(\sum_{r' \in  R_l}\alpha_{r,r'}^h W_h f_{r^{'}} + b_h\right) ;
  \text{ } f_{r} = \text{ }\psi(I_n(A_{r})),
\end{aligned}
\label{eqn:attn}
\end{equation}     
where $\psi$ is the feature representation of region $A_r$ in image $I_n$, and $\sigma$ is a nonlinear activation function. In our experiments, $\psi$ refers to local operation (within a region) of global average pooling, and $\sigma$ denotes Exponential Linear Unit (ELU) \cite{clevert2015fast}. The normalized attention coefficients $\alpha_{r,r'}$ indicate the ``importance'' of region $r'$ to region $r$  ($r,r'\in \{R_l\}$). $W$ is the shared weight matrix 
and is used for the linear transformation of each region. $\mathbf{W_a}$ is the weight matrix of the attention mechanism \cite{velivckovic2017graph} consisting of a single-layer feedforward neural network with the LeakyReLU as a nonlinear activation function. $||$ represents concatenation 
to extend single attention to \textit{multi-head attention} (similar to \cite{vaswani2017attention}) to stabilize the learning process \cite{velivckovic2017graph}. 
We follow this suggestion to 
use $H$ independent attention heads, 
where $\alpha_{r,r'}^h$ are the normalized attention coefficients from the $h^{th}$ attention head with the respective linear transformation's weight matrix $W_h$ and bias $b_h$. 
We compare the performance of concatenation with averaging, and experimentally find that the former is better than the latter aggregation. 

The above process is applied to each graph $G_l$ within the hierarchical representation consisting of $L$ 
graphs corresponding to $L$ layers within the structure. The learnable parameters in (\ref{eqn:attn}) 
are graph-specific i.e., not shared among graphs representing different layers, but shared between nodes within a graph. The next objective is how to combine these graphs to make the region-level classification decision. We achieve this by exploring graph pooling, which allows a graph
to learn more abstract representations of the input graphs increasingly by summarizing local components, 
and is similar to the pooling operations in CNNs. We are inspired by the recent advancement in model-based pooling methods \cite{bianchi2020spectral,lee2019self,gao2019graph,ying2018hierarchical} that perform pooling through a learnable function. 
In this work, we adapt the spectral clustering-based graph pooling \cite{bianchi2020spectral}, which computes the soft clustering of the input graphs by aggregating regions belonging to the same cluster using a multi-layered perceptron (MLP) with softmax activation function. The MLP maps each region feature $f_{r,l}^{'}$ $\in$ $\{f^{'}\}$ in graph $G_l$ into the $i^{th}$ row of a soft cluster assignment matrix $\mathbf{S}$ $\in$ $\mathbb{R}^{R\times K}$ i.e. $\mathbf{S}$=$\text{MLP}(f^{'};\theta)$, where $R=\sum_{l=1}^LR_l$ is the total number of regions within the hierarchical structure and $K$ is the target number of clusters ($K=3$ in Fig. \ref{fig:full_model}c). The softmax activation guarantees the value of $s_{i,j}\in \mathbf{S}$ within $[0,1]$, resulting in soft cluster assignment. As a result, we are able to combine $L$ number of graphs 
to a single complete graph (Fig. \ref{fig:full_model}c) with $K$ nodes in which each node represents a cluster of closely related regions for making the final prediction. 
\subsection{Gated attention for prediction}
Given a graph representation, the image classification can be linked to either node-level \cite{kipf2016semi} or graph-level \cite{ying2018graph}. 
At node-level, a single label to a node is assigned and is based on its high-level feature. 
Whereas, a compact representation of the graph by combining pooling and readout operations is often used in graph-level \cite{bianchi2020spectral,lee2019self, ying2018hierarchical, zhang2018end} for classifying the entire graph. A significant drawback of the node-level classification is its inability to propagate hierarchical information to facilitate decision-making. Similarly, graph-level approaches aggregate node representations before applying classification. As a result, the node-level decision making capability is overlooked. It is worth mentioning that node-level decisions have some relationships with the graph-level predictions \cite{holtz2019multi}.  
Thus, we use a novel gated attention mechanism to aggregate the node-level class-specific confidence instead of node-level feature representation for the graph-level classification. First, we compute the node-level class-specific confidence using a shared 
single-layered perceptron (SLP) applied to each node $k$ i.e., $\beta_k = f_k W_1 + b_1$, parameterized by a weight matrix $W_1$ and bias $b_1$, resulting in output $\beta_k \in \mathbb{R}^{1\times C}$ where $C$ is the number of classes. The dimension of $f_k$ is the same as the $f^{'}_{r,l}$ as $f_k$ represents a cluster of regions using soft clustering, as mentioned before.
Second, for graph-level predictions, we define a soft attention mechanism that decides which node's class-specific confidences are relevant to the current graph-level task and is computed as: 
\begin{equation}
    y_n = \text{Softmax}\left(\sum_{k=1}^{K}\beta_k \odot \sigma(W_2 f_k + b_2)\right),
    \label{eqn:pred}
\end{equation}
where $\sigma(.)$ is 
the sigmoid activation and $\odot$ is a Hadamard product. The shared weight matrix $W_2$ and bias $b_2$ are the parameters of the corresponding linear transformation. 
%
\section{Experiments}
We consider both fine-grained (Aircraft-100 \cite{maji13fine-grained}, Oxford Flowers-102 \cite{Nilsback08}, and Oxford-IIIT Pets-37 \cite{parkhi12a}) and generic (CIFAR-100 \cite{krizhevsky2009learning} and Caltech-256 \cite{griffin2007caltech}) visual classification tasks to demonstrate our method under general situations of data diversity using five  
benchmark datasets. 
It is worth mentioning that CIFAR-100 dataset consists of tiny (32$\times$32 pixels) RGB images in comparison to the rest. We compare our approach with a wide variety of strong baselines and past methods. We follow the standard train/test split described in the respective datasets, and use the conventional top-1 accuracy (\%) for the evaluation. We implement our model in TensorFlow 2.0. We use Stochastic Gradient Descent (SGD) optimizer with a learning rate of $10^{-5}$. 
The models are trained for 200 epochs with a batch size of 8 on a 
NVIDIA Quadro RTX 8000 GPU (48 GB memory).

\subsection{Implementation details} For a fair comparison \cite{behera2021context, yu2018bisenet}, we use the lightweight Xception \cite{chollet2017xception} as a backbone CNN and the output is 6$\times$ upsampled using a bilinear upsampler. Pre-trained ImageNet weights are used to initialize Xception for quicker convergence \cite{he2019rethinking}. Regions of the fixed area but with different aspect ratios corresponding to a given hierarchical layer are generated using the region proposal algorithm in \cite{behera2020regional}. A single-layer GCN (Fig. \ref{fig:full_model}b) is used for the layer-wise graph representation of a set of regions. The GCN employs $H=3$ attention heads in 
(\ref{eqn:attn}) with a per head output feature size of 512, resulting in final feature dimension of $512\times3=1536$. 
A dropout rate of 0.2 is applied to the normalized attention coefficients $\alpha_{r,r'}$ in (\ref{eqn:attn}). This signifies each region in a given layer is exposed to a stochastically sampled neighborhood in every training iteration. A single-layer MLP is used in spectral clustering-based soft clustering. The number of cluster $K$ is empirically evaluated and is found to be dataset-dependent. For a 3-layer hierarchical structure (Fig.\ref{fig:full_model}b), there are 52 possible regions (layer 1: 25, layer 2: 22, and layer 3: 5) generated using the region proposal in \cite{behera2020regional}. Input images are resized to 256$\times$256, and data augmentation of random rotation and random zoom are used. Finally, random cropping is applied to select the final size of 224$\times$224.
\begin{table*}[t]
\small
    \centering
\begin{tabular}{p{0.90 cm}p{3.8 mm}|p{1.0 cm}p{4 mm}|p{1.1 cm}p{4 mm}||p{1.46cm} p{3.9mm}|p{1.15cm}p{0.8cm}}
        \hline
        \multicolumn{2}{c}{Aircraft \cite{maji13fine-grained}} & \multicolumn{2}{c}{Flowers \cite{Nilsback08}} & \multicolumn{2}{c}{Pets \cite{parkhi12a}} & \multicolumn{2}{c}{CIFAR-100 \cite{krizhevsky2009learning}} & \multicolumn{2}{c}{Caltech-256 \cite{griffin2007caltech}}\\
        Method & Acc & Method & Acc & Method & Acc & Method & Acc & Method & Acc$\pm$std  \\
        \hline
S3N\cite{ding2019selective} & 92.8  & InAt\cite{xie2016interactive}  & 96.4 & One\cite{xie2015image}&90.0  & OCN\cite{wang2020orthogonal} & 80.1 &  IFK\cite{perronnin2010improving} &47.9$\pm$0.4 \\
    
    MC$^{\color{blue}*}$\cite{chang2020devil} & 92.9 & SJF$^{\color{blue}*}$\cite{ge2017borrowing}  & 97.0 & FOA\cite{zhang2015fused} & 91.4 &  DML$^{\color{magenta}\dagger}$\cite{zhang2018deep} & 80.3  & CLM\cite{wang2016towards} &53.6$\pm$0.2\\
    
    DCL\cite{chen2019destruction} & 93.0 & OPA$^{\color{blue}*}$\cite{peng2017object} & 97.1 & NAC\cite{simon2015neural} & 91.6 & SD$^{\color{magenta}\dagger}$\cite{xu2019data} & 81.5 & FV\cite{sanchez2013image} &57.3$\pm$0.2 \\
    
    GCL\cite{wang2020graph} & 93.2 & CL$^{\color{blue}*}$\cite{barz2020deep} &	97.2 & TL$^{\color{blue}*}$\cite{xiao2015application} & 92.5 & WRN40\cite{zagoruyko2016wide} & 81.7 & ZFN\cite{zeiler2014visualizing} & 74.2$\pm$0.3\\
    
    PMG\cite{du2020fine} & 93.4 & PMA$^{\color{magenta}\mathsection}$\cite{song2020bi} & 97.4 & InAt\cite{xie2016interactive} & 93.5 & WRN28\cite{zhong2020random} & 82.3 &  VGG$^{\color{cyan}\ddagger}$\cite{simonyan2014very} & 86.2$\pm$0.3 \\
    
     CSC\cite{wang2020category}  & 93.8 & DST$^{\color{blue}*}$\cite{cui2018large} & 97.6 & OPA$^{\color{blue}*}$\cite{peng2017object} & 93.8 & BOT$^{\color{magenta}\dagger}$\cite{zhang2019your} & 83.5 & L$^2$SP\cite{xuhong2018explicit} & 87.9$\pm$0.2 \\
     
    GRD\cite{zhao2021graph} & 94.3 & MC$^{\color{blue}*}$\cite{chang2020devil} & 97.7 & GPipe$^{\color{blue}*}$\cite{huang2019gpipe} & 95.9 & Augment\cite{cubuk2019autoaugment} & 89.3 & VSVC\cite{zhang2019multi} & 91.4$\pm$0.4 \\
   
    CAP\cite{behera2021context} & \textbf{94.9} & CAP\cite{behera2021context}  &97.7 & CAP\cite{behera2021context}  &97.3 &  GPipe$^{\color{blue}*}$\cite{huang2019gpipe} & \textbf{91.3} & CPM$^{\color{blue}*}$\cite{ge2019weakly} & 94.3$\pm$0.2 \\
        \hline
    Baseline & {79.5} & {Baseline}  &{91.9}  & {Baseline} & {91.0}  & {Baseline} &{80.9} & {Baseline} &{72.2$\pm$0.4} \\
    \textbf{Ours} & \textbf{94.9} & \textbf{Ours} & \textbf{98.7} & \textbf{Ours} & \textbf{98.1} & \textbf{Ours} & 83.8 & \textbf{Ours} & \textbf{96.2}$\pm$\textbf{0.1} \\
        \hline
    \end{tabular}
\caption{Accuracy (\%) comparison with the most recent top eight SotA methods. {\color{blue}*} involves transfer/joint learning strategy for objects/patches/regions involving more than one dataset (target and secondary). {\color{magenta}$^\mathsection$} uses additional textual description. {\color{magenta}$^\dagger$} applies self-knowledge distillation. {\color{cyan}$^\ddagger$} combines VGG-16 and VGG-19. The {baseline} accuracy is without our attention-driven hierarchical multi-scale  representation.
    }
    \label{tab:all_results} 
\end{table*}
\vspace{- 0.5 cm}
\subsection{Comparison with the state-of-the-arts (SotA)}
The performance of our model is compared with the recent eight SotA methods in Table \ref{tab:all_results}. Based on the dataset characteristics, we divide our comparison into two groups: 
\vspace{- 0.4 cm}
\subsubsection{Fine-grained image classification} 
We compare with the recent methods that avoid object/parts bounding-box annotations as followed in our work. 
Our method outperforms these existing methods on Flowers (1.0\%), and Pets (0.8\%) datasets. 
It attains 94.9\% accuracy on Aircraft akin to CAP \cite{behera2021context}. For a fair comparison with CAP, our method achieves the best performance on all three fine-grained datasets using Xception as a backbone. CAP's accuracy using Xception are: 94.1\% on Aircraft, 97.7\% on Flowers, and 97.0\% on Pets. Using the same backbone, our method gains a clear margin on Aircraft (0.8\%), Flowers (1.0\%) and Pets (1.1\%). We also achieve the SotA results compared to the others based on the attention mechanism \cite{behera2021context, peng2017object, chang2020devil, xiao2015application}, GCN \cite{wang2020category, wang2020graph, zhao2021graph}, and low-rank discriminative bases \cite{wang2020weakly}. Moreover, our approach gives superior performance over these existing methods (marked {\color{blue}*} in Table \ref{tab:all_results}) without leveraging secondary datasets. For example, ImageNet is used in \cite{chang2020devil} for joint learning (Flowers: 97.7\%), and pre-trained sub-networks or even higher image resolution (e.g. 448$\times$448 in PMG \cite{du2020fine}, 480$\times$480 in GPipe \cite{huang2019gpipe}, 
MC \cite{chang2020devil}, etc.) is considered in improving FGVC accuracy. Moreover, CSC (Aircraft: 93.8\%) \cite{wang2020category} uses three-modules and its graph representation interacts with a fewer number of patches (2-4). For a higher number of regions, it 
fails to learn spatial contextual information resulting in lower accuracy. Likewise, DFG \cite{wang2020weakly} also suffers from the same scalability problem in their graph-structure. Our model is scalable to any number of regions without increasing the model parameters since the graph node's parameters are shared. 
This clearly reflects our model's ability to capture subtle changes, avoiding any additional resource/constraint to enhance accuracy.
\vspace{- 0.2 cm}
\subsubsection{Generic image classification}
AutoAugment \cite{cubuk2019autoaugment} used PyramidNet (26M parameters) with ShakeDrop regularization to achieve 89.3\% accuracy on CIFAR-100 following a resource intensive (4 GPUs) training. 
It is improved to 91.3\% using a giant AmoebaNet-B (557M) in GPipe \cite {huang2019gpipe}. Whereas, AutoAugment's accuracy degrades to 82.9\% using Wide-ResNet WRN-28-10 (36.5M). Method in \cite{zagoruyko2016wide} attains 81.7\% with WRN-40-10, and drops to 81.2\% with WRN-28-10. Thus, the backbone CNN plays a vital role for performance gain and are susceptible to higher network depth, architecture, and capacity. It is also evaluated in BOT \cite{zhang2019your}, showing the lighter versions of ResNet/WRN achieve lower performance (ResNet152 (60M): 82.3\%, WRN44-8: 82.6\%) than their best accuracy (83.5\%) using PyramidNet101-240. 
Contrarily, we include our novel module on the top of Xception (22.9M), attaining better performance (83.8\%) with lesser computational overhead and complexity. In addition, we achieve 2.2\% improvement over GPipe on both Aircraft (92.7\%) and Pets (95.9\%). Likewise, the gains over AutoAugment are: Aircraft 2.2\%, Flowers 3.3\%, and Pets 9.1\%. Our method suppresses many SotA methods and attains better accuracy with a novel attention-driven hierarchical multi-scale representation using GCN, 
with 36.1M parameters that is 0.4M lesser than WRN-28-10. Hence, our end-to-end lightweight model performs comparatively better than the methods used heavier and deeper backbones with resource-intensive training.

We achieve 96.2\% on Caltech-256 and is better (1.9\%) than the recent best CPM \cite{ge2019weakly} that uses additional training (secondary) data from ImageNet for selective-joint fine-tuning. 
Also, CPM follows a complex and multi-step training mechanism. However, its accuracy is 93.5\% with only the target dataset. Moreover, our model gains 4.8\% improvement over the recent VSVC \cite{zhang2019multi} that combines multi-view information. Thus, it clearly reflects the efficacy of our method for generic visual recognition on top of the FGVC.
\begin{figure*}[t]
\centering
\subfigure[]{\includegraphics[width=0.2\textwidth] {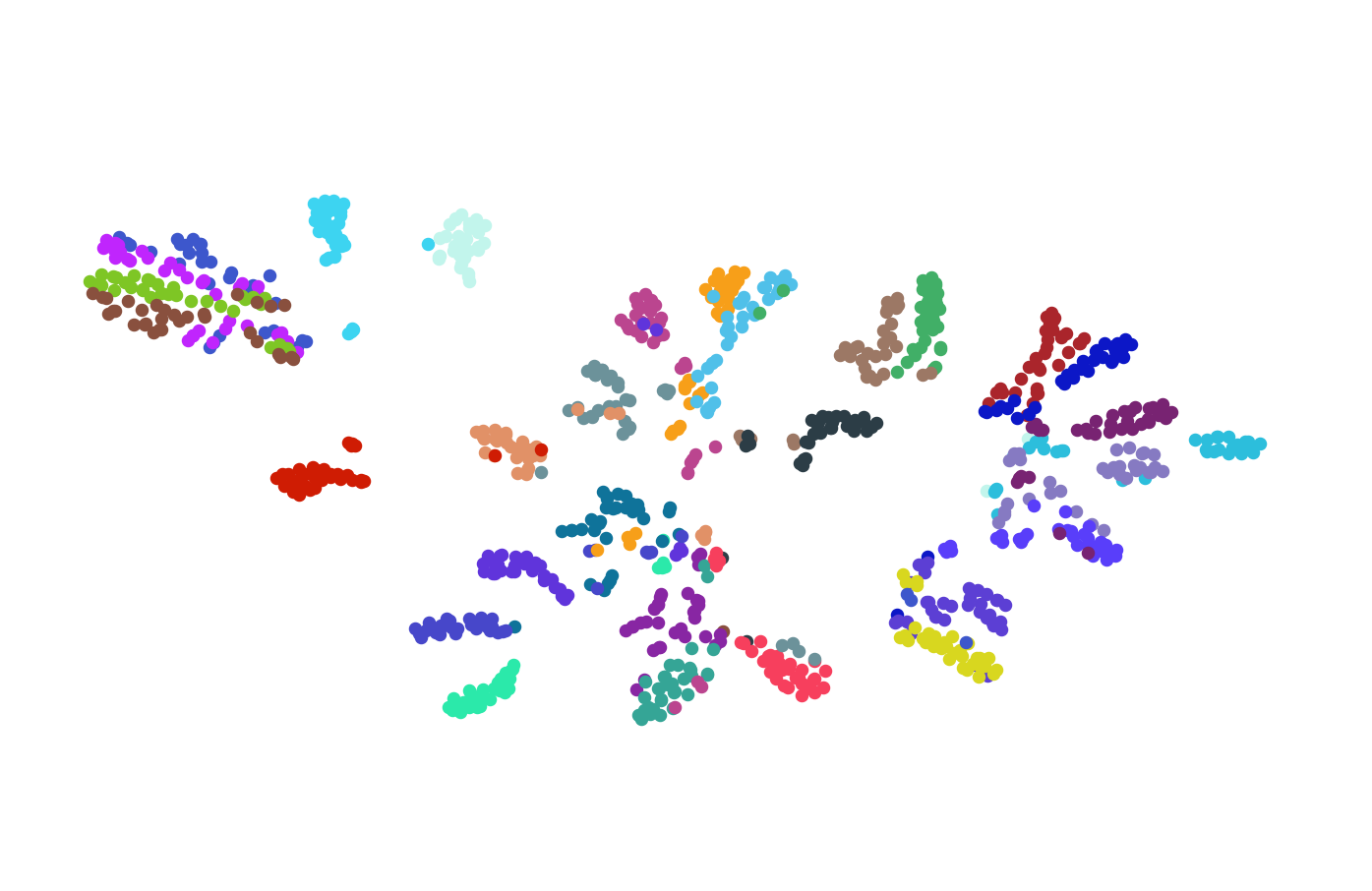}\hspace{-.8em}
    \includegraphics[width=0.2\textwidth] {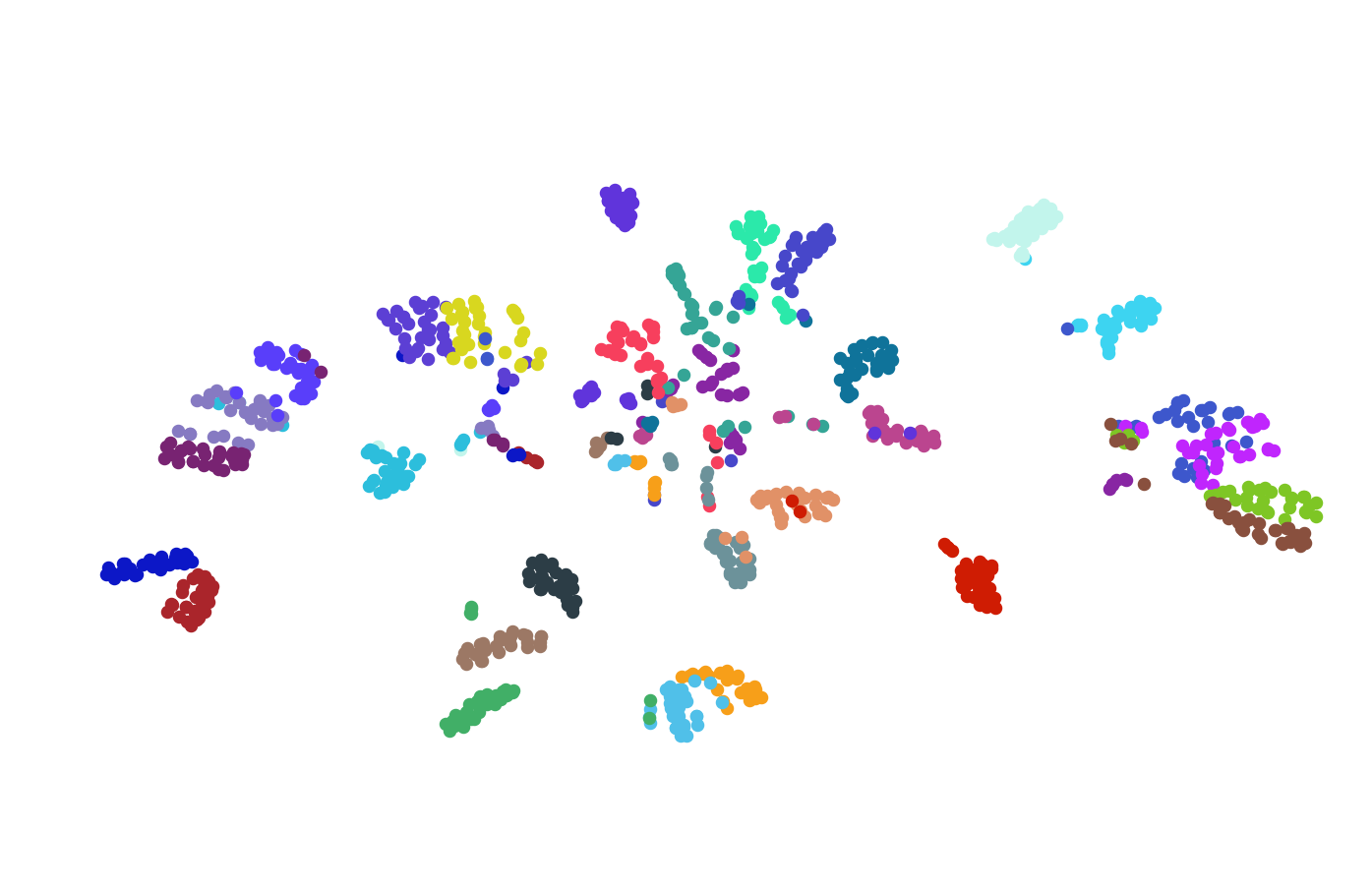}\hspace{-0.8em}
    \includegraphics[width=0.2\textwidth] {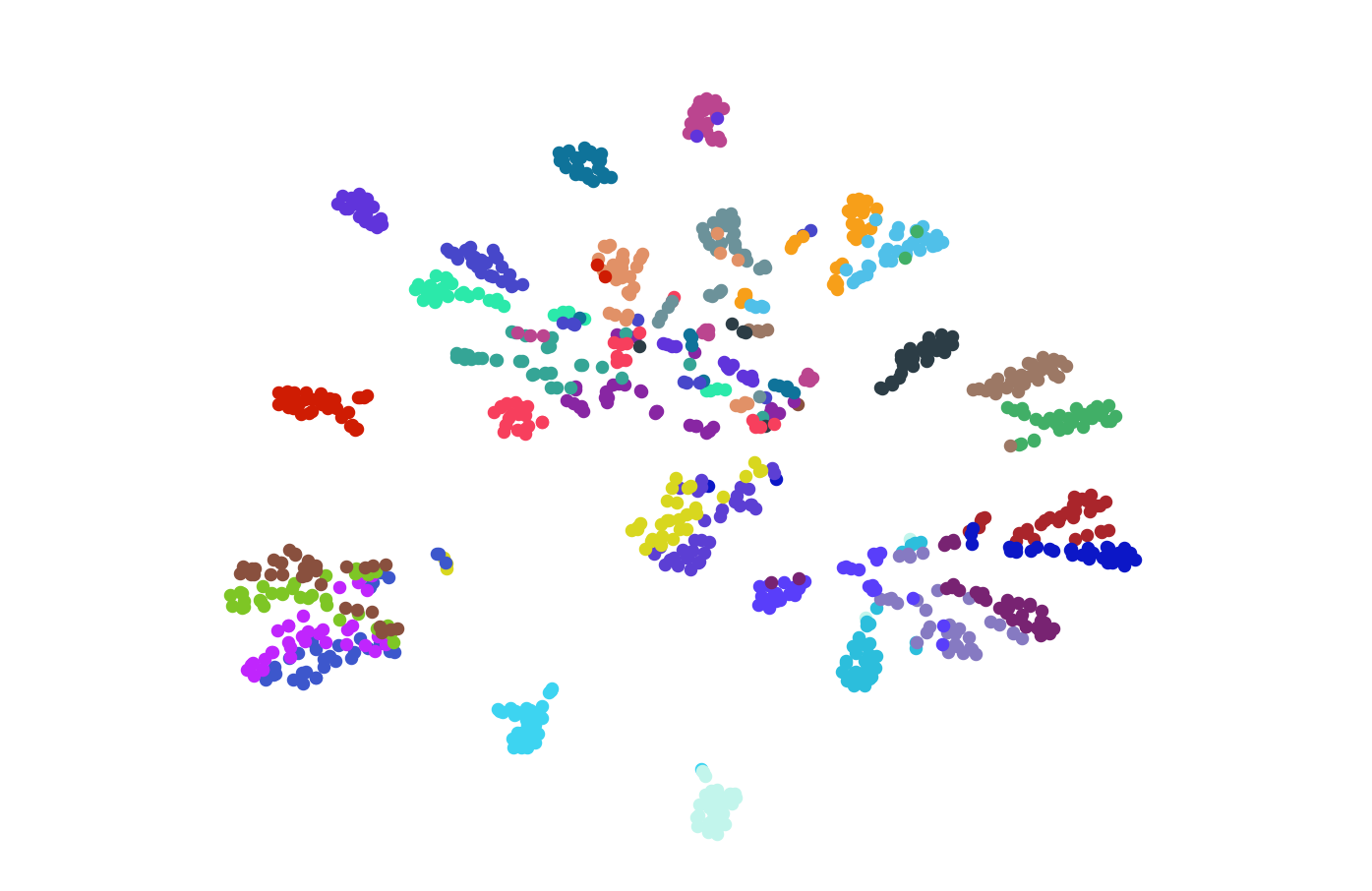}\hspace{-1.8em}
    \includegraphics[width=0.2\textwidth] {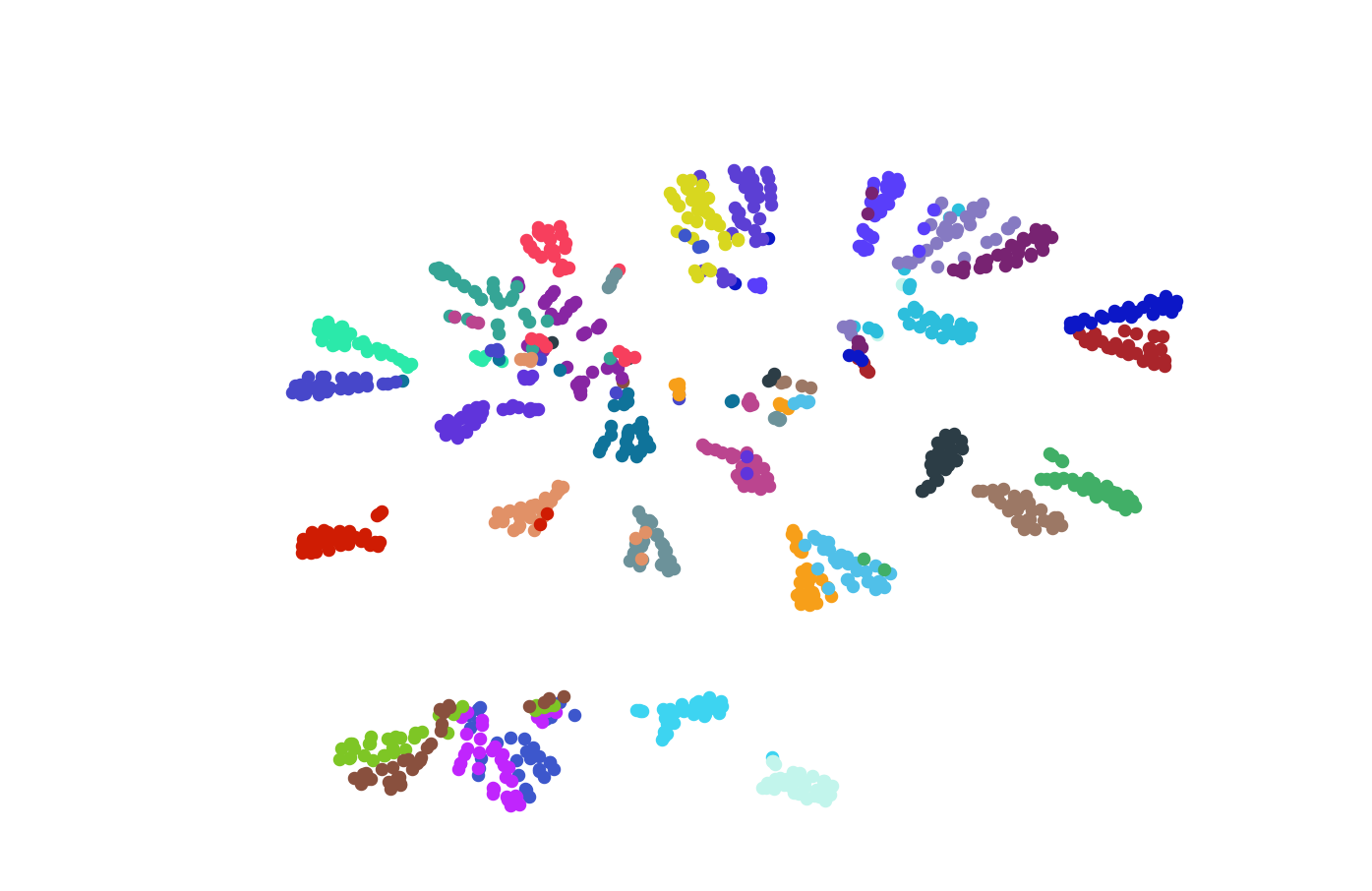}\hspace{.2em}
    }
    
\subfigure[]{\includegraphics[width=0.2\textwidth] {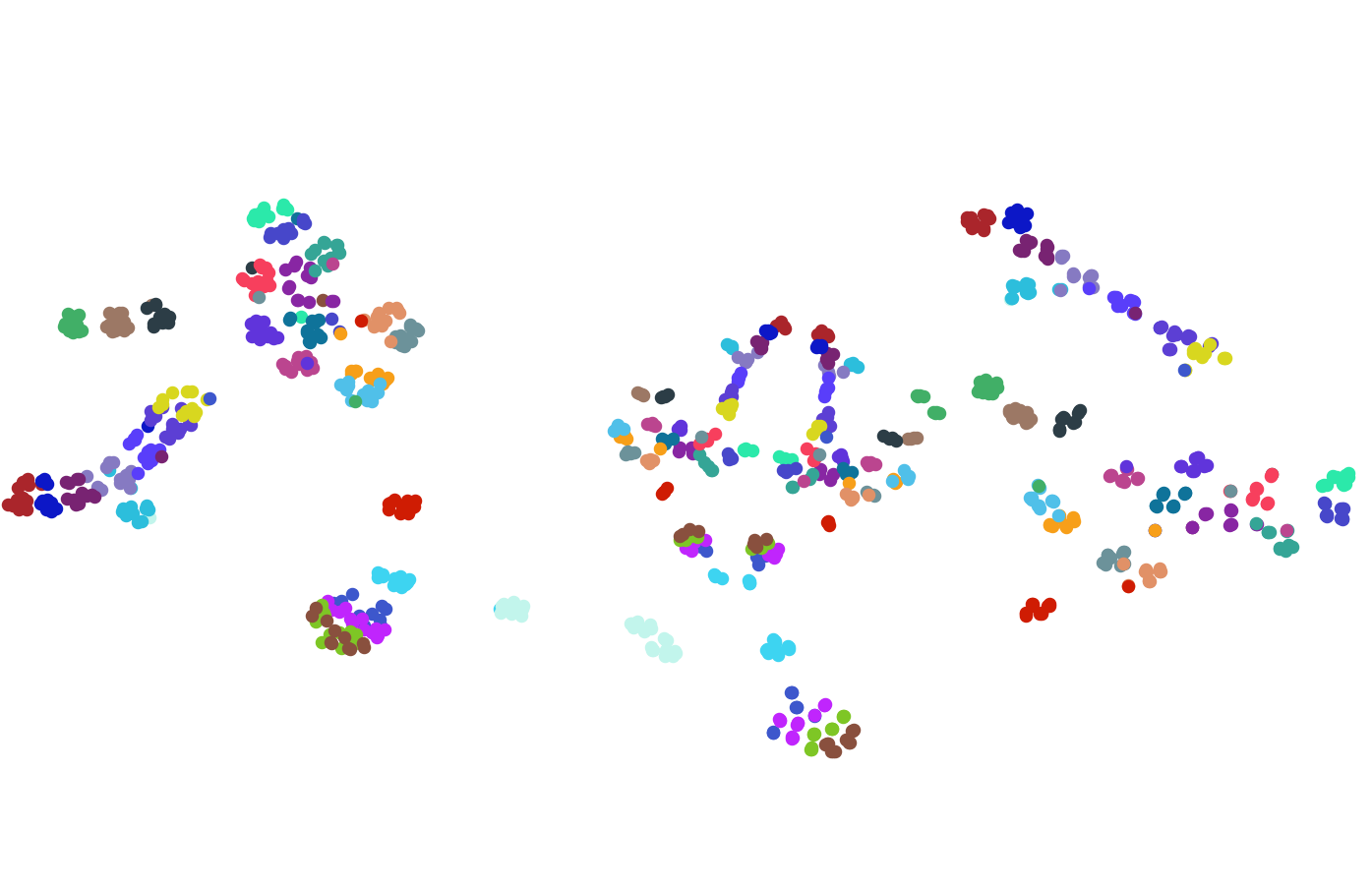}\hspace{-.3em}
    \includegraphics[width=0.2\textwidth] {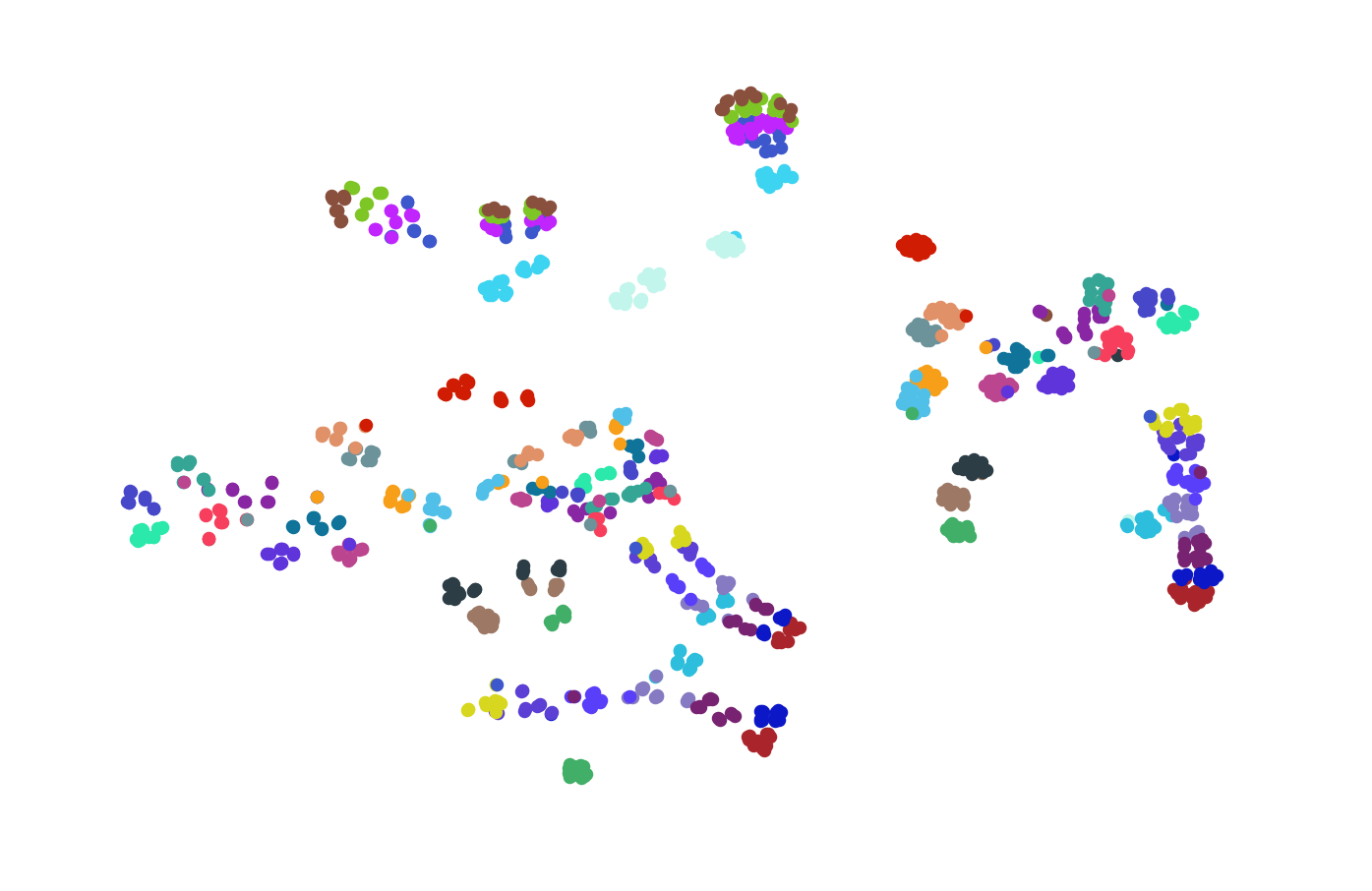}\hspace{-1.5em}
    \includegraphics[width=0.2\textwidth] {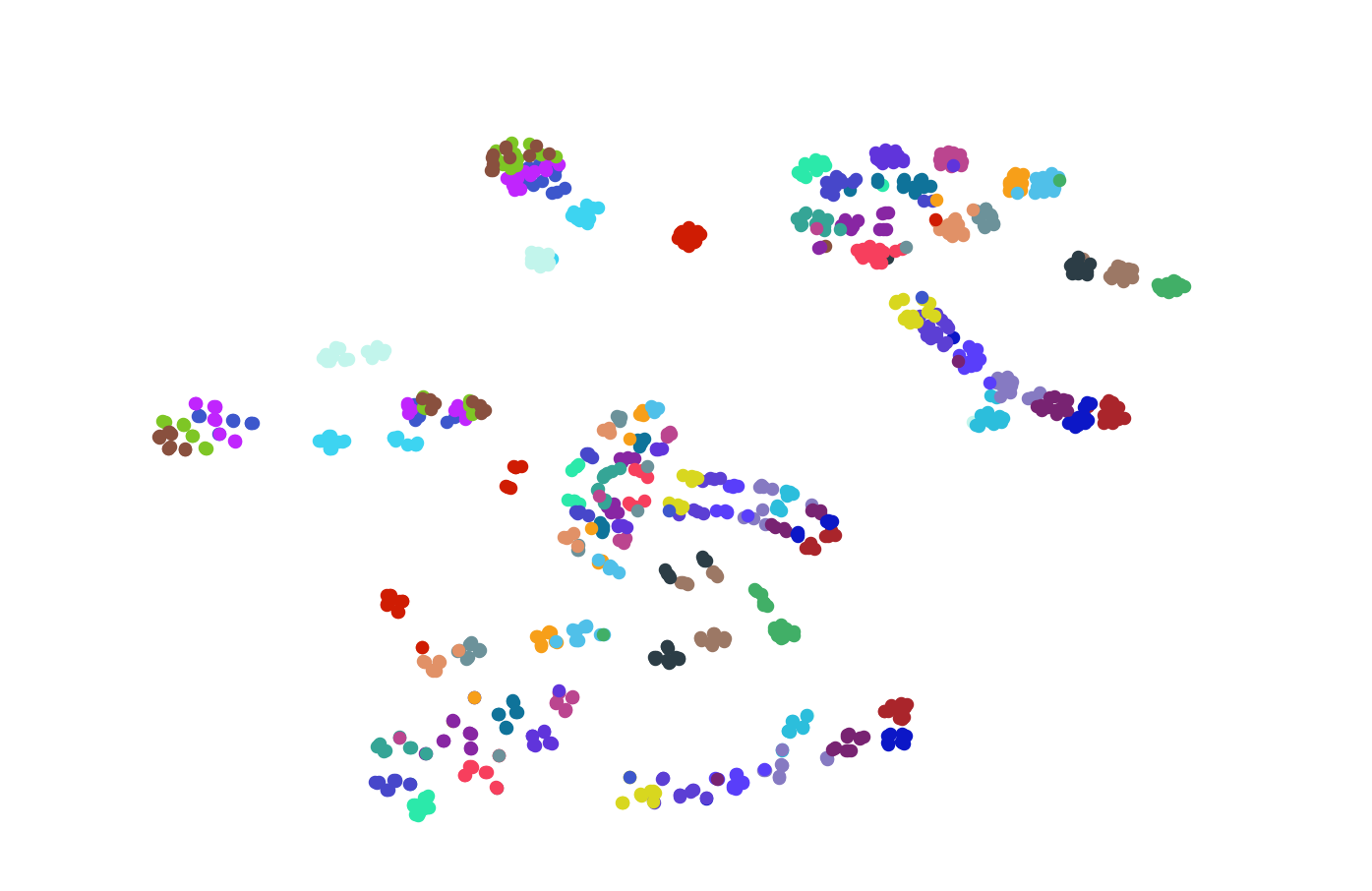}\hspace{-2em}
    \includegraphics[width=0.2\textwidth] {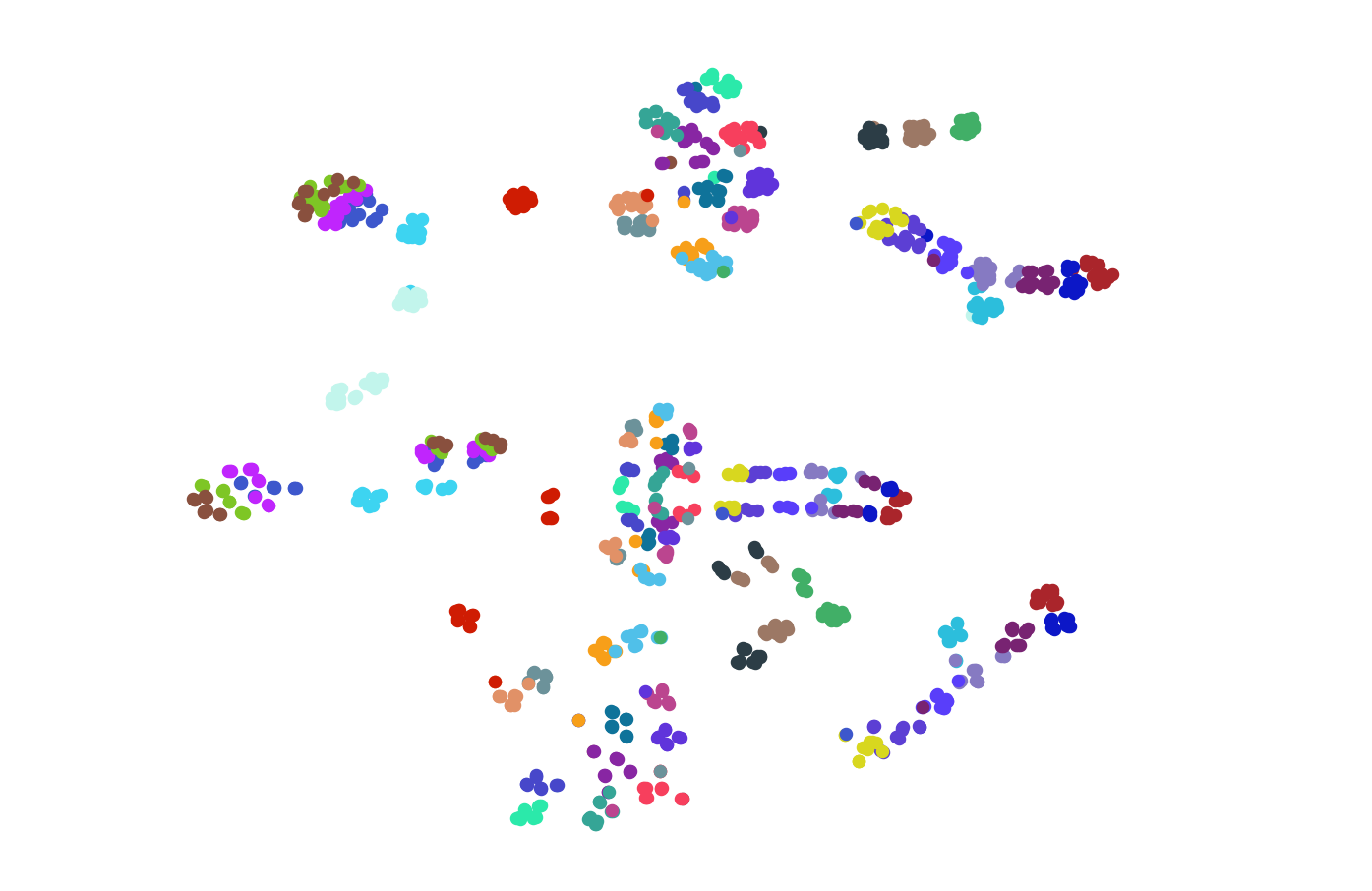}
    }\\ \vspace{-1.2em}
    
\subfigure[]{\includegraphics[width=0.2\textwidth] {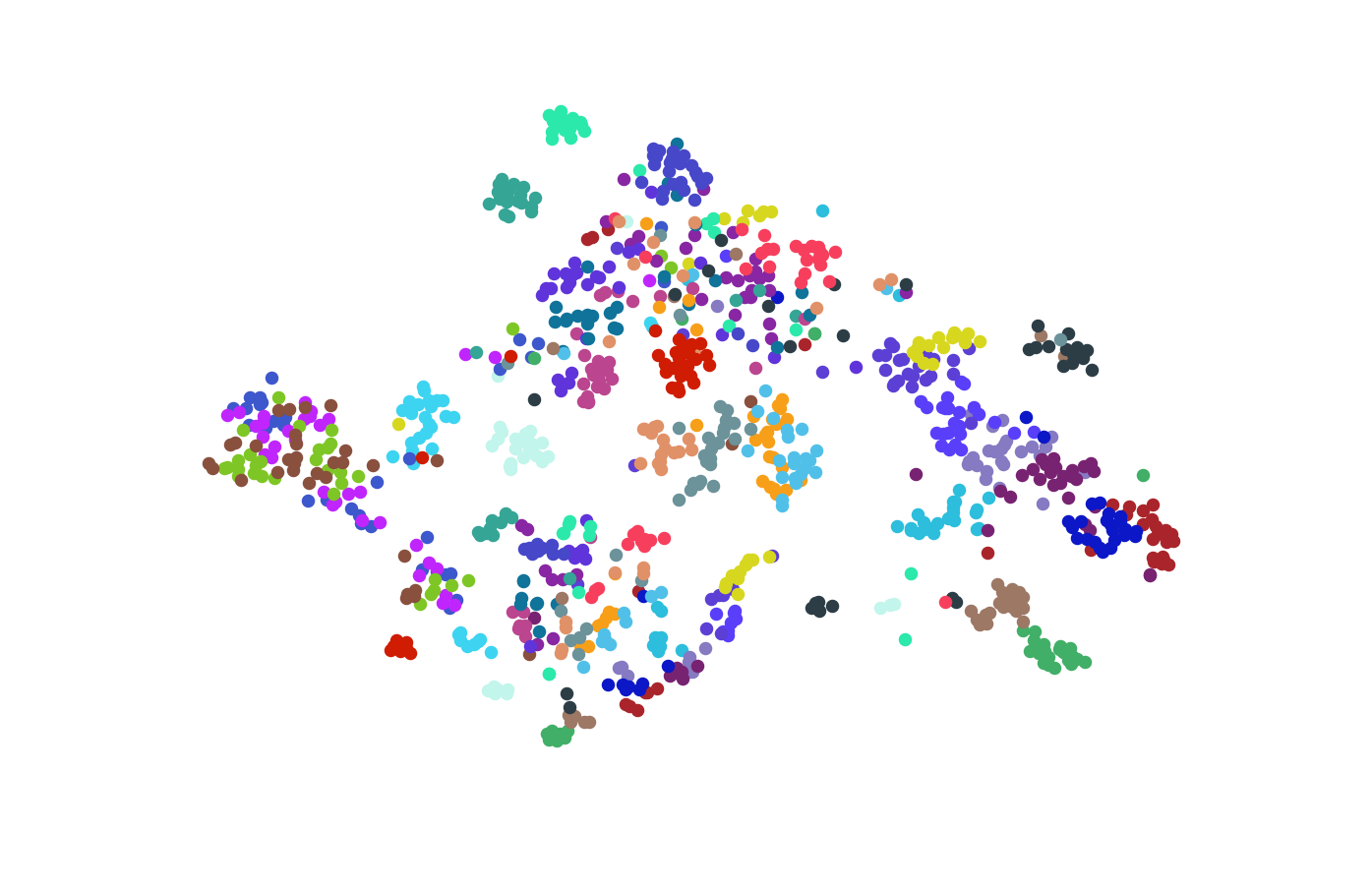}\hspace{-1.5em}
    \includegraphics[width=0.2\textwidth] {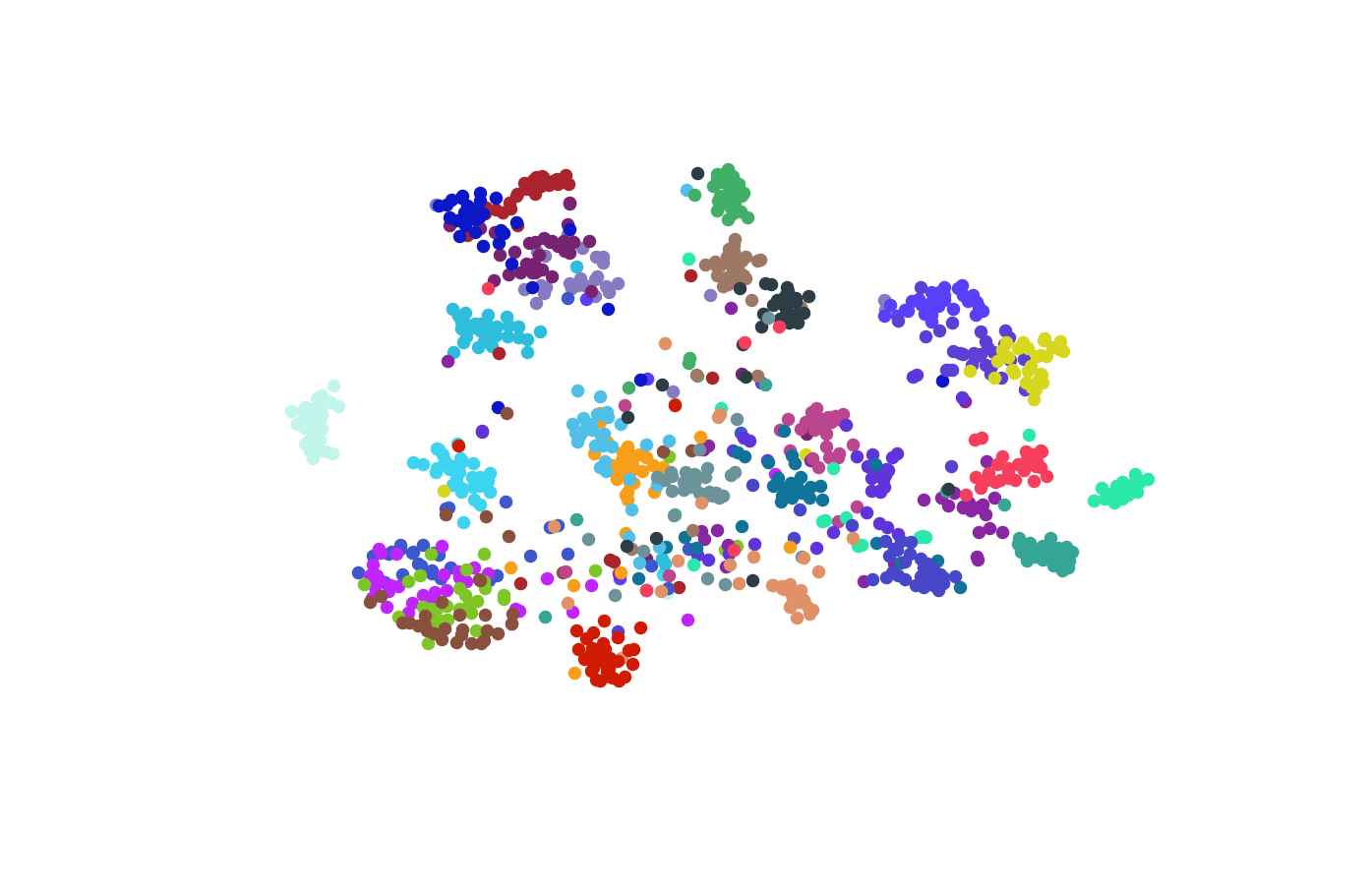}\hspace{-1.5em}
    \includegraphics[width=0.2\textwidth] {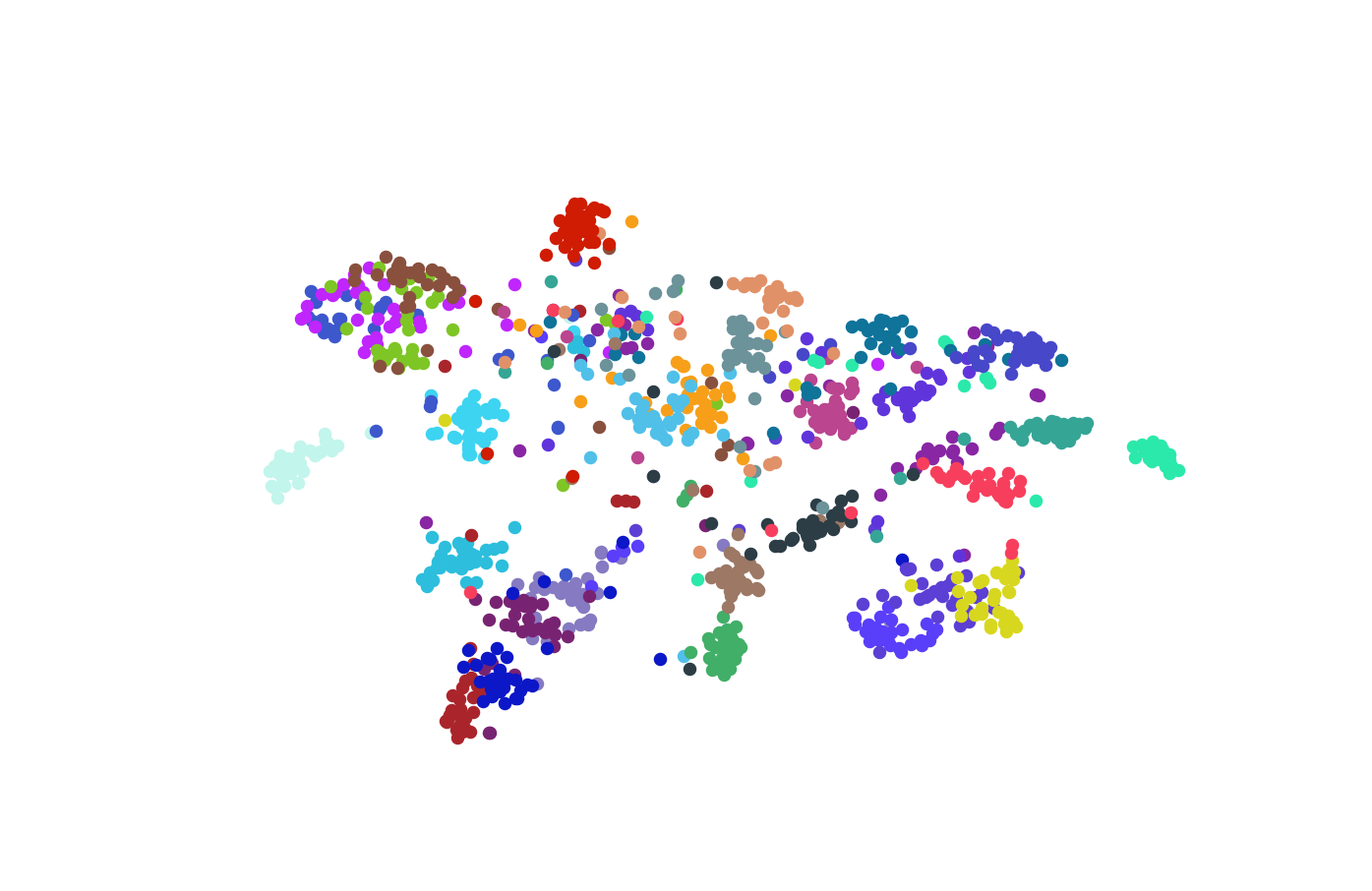}\hspace{-1.5em}
    \includegraphics[width=0.2\textwidth] {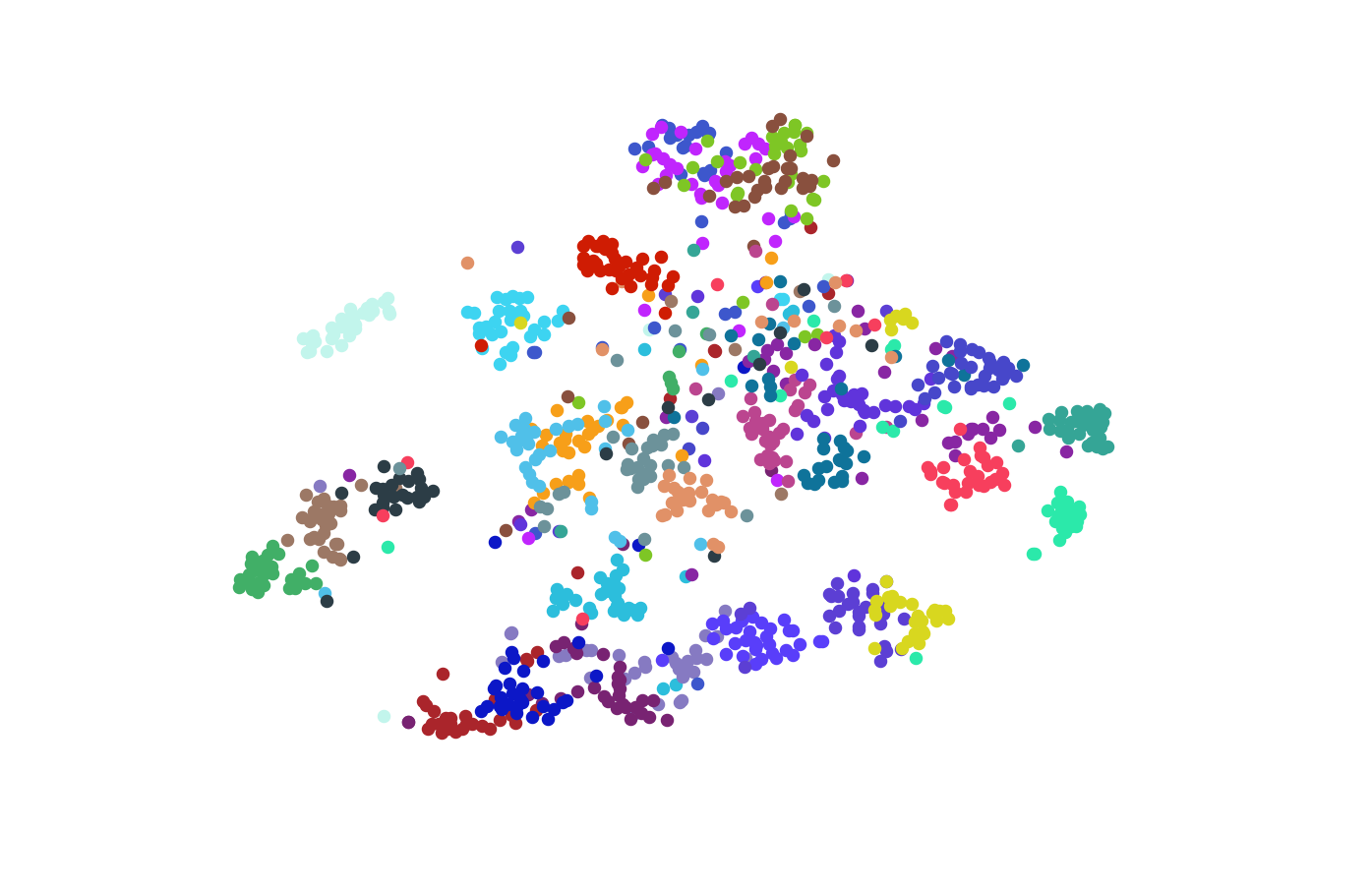} 
    }
    
\subfigure[]{\includegraphics[width=0.25\textwidth] {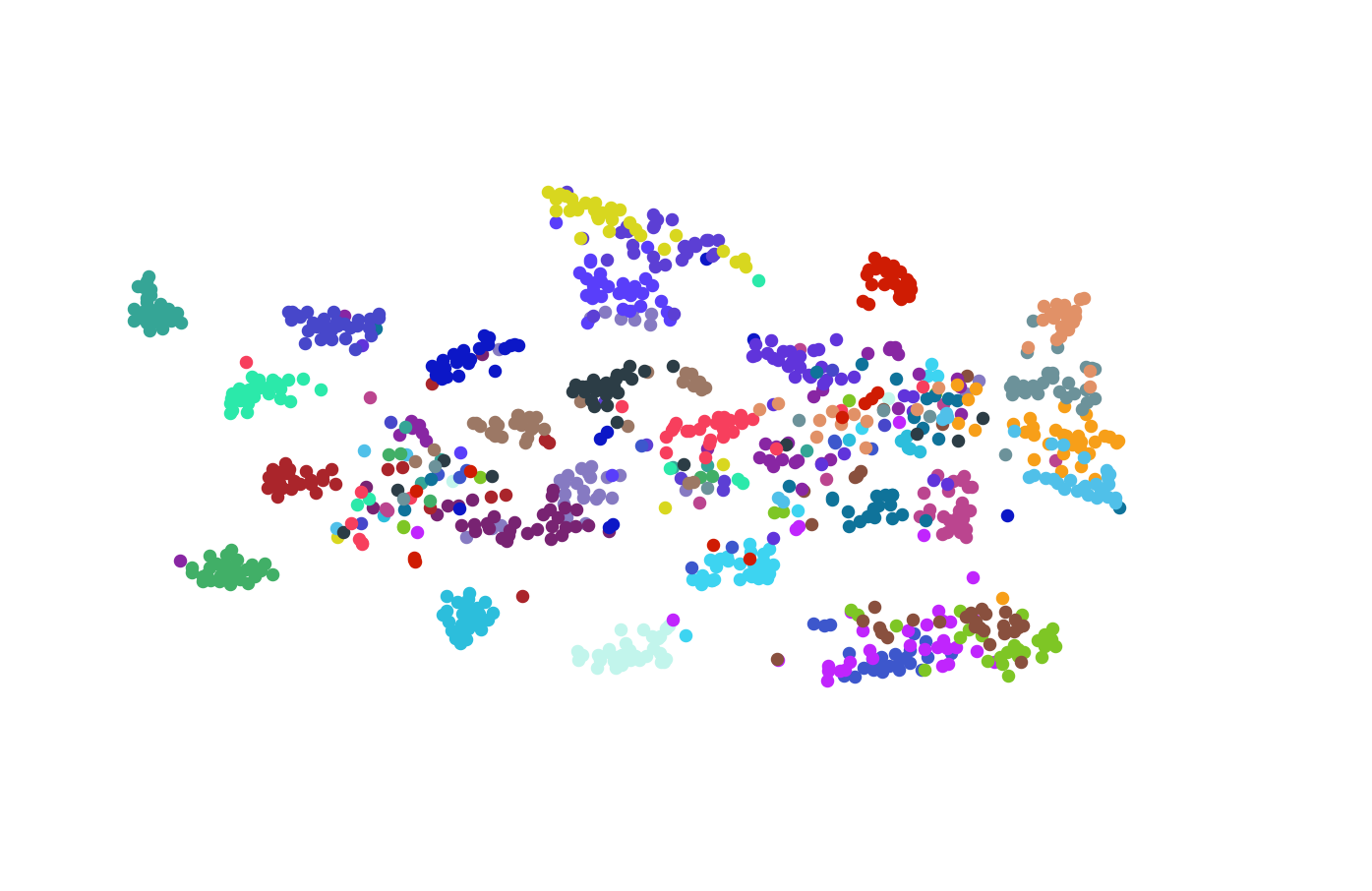}
    \includegraphics[width=0.25\textwidth] {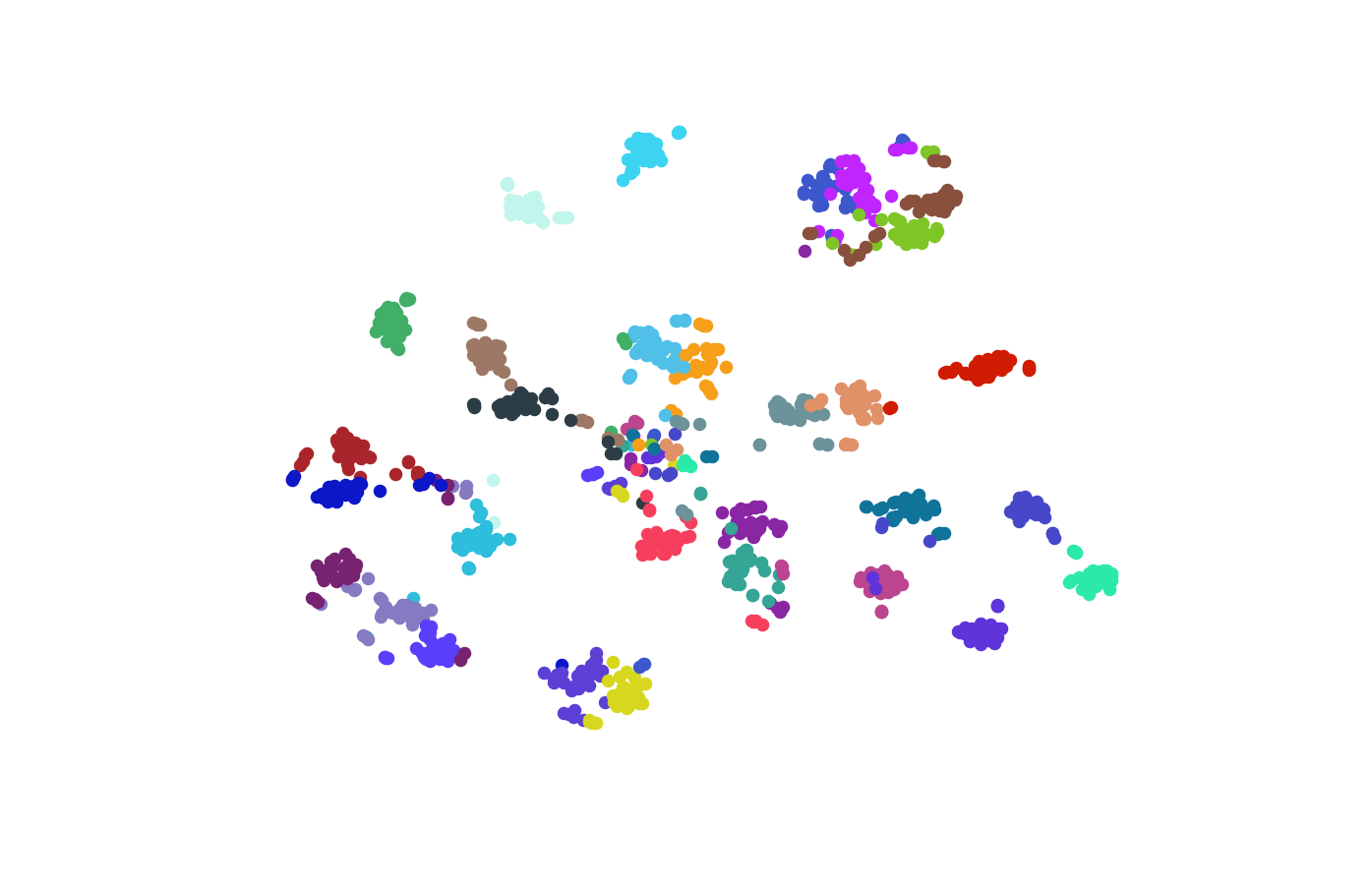}
    \includegraphics[width=0.25\textwidth] {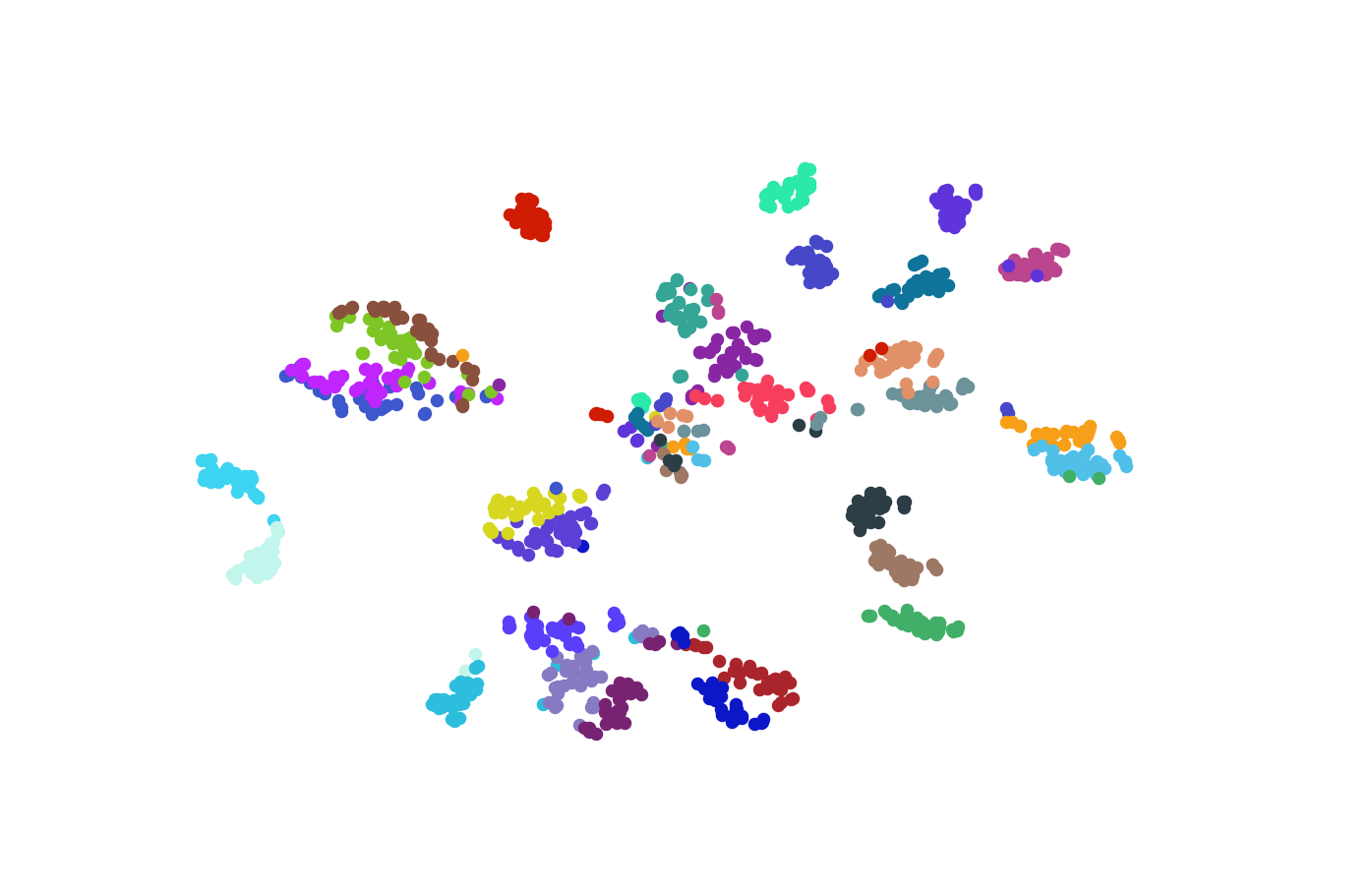}
    }
\caption{t-SNE visualization \cite{van2014accelerating} of class-specific discriminative feature 
using $H=3$ attention heads in (\ref{eqn:attn}), and $L=3$ 
in (\ref{eq_1}). 
(a) top-left: Layer 1 (left to right): head$_1$, head$_2$, head$_3$, and their concatenation;  (b) top-right: Layer 2: same order as in (a); 
  (c) bottom-left: Layer 3: same order as in (a); 
  (d) bottom-right: final representation (left to right) using 2, 3 and 4 attention heads.
Attention head-specific plots are shown in (Layer 1) $\rightarrow$ (Layer 2) $\rightarrow$ (Layer 3), representing layers from smaller regions to larger ones. It is evident that the discriminability of the features representing medium-size regions (Layer 2) $>$ small-size (Layer 1) $>$ large-size (Layer 3), and could be linked to the problem of FGVC. (d) shows the combined layers' representation using $H=$ 2, 3 and 4 attention heads. }
    \label{fig:fig_feat}
\end{figure*}
\subsection{Model complexity and qualitative analysis}
The complexity is given as Gigaflops (GFLOPS) and trainable parameters in millions (M). For a 3-layer architecture with Xception backbone, 3 attention heads and output dimension of 512, our model has 36.1M parameters and 13.2 GFLOPS. The same with an output dimension of 256, these are 29.4M and 13.2 GFLOPS. The number of regions does not impact on the parameters, but a very little ($\sim 10^{-2}$) increase in GFLOPS. This is mainly because the parameters are shared between regions belonging to a given graph. Our model is also comparable to CAP \cite{behera2021context} (34.2M) and lighter than RAN \cite{behera2020regional} (49M). Our per-image inference time is 8.5 milliseconds (ms). It is 27ms for step 3 and extra 227ms in step 2 of the method in \cite{ge2019weakly}. The inference time for FCANs \cite{liu2016fully} is reported as 150 ms. Additional information on complexity is included in the supplementary. 

We provide qualitative analysis via visualization (t-SNE \cite{van2014accelerating}) to get insight into our model. Firstly, we look into the class-specific discriminability of features representing regions using our layer-wise graph structure, and is shown in Fig. \ref{fig:fig_feat}. All test images from 30 randomly chosen classes in Aircraft are processed for this visualization. Each layer is represented as a complete graph with nodes linking regions. Three ($H=3$) attention heads are attached to a region and output from each head is visualized as well as their concatenation. From Fig. \ref{fig:fig_feat}, it is evident that multi-scale hierarchical representation does influence the discriminability.  Secondly, we look into the cluster-specific ``importance'' (i.e., higher weights) of regions towards a class during the spectral clustering-based graph pooling, and is shown in Fig. \ref{fig:clust1}. 
It clearly displays the effect of different number of clusters ($K$) to aggregate contributions from regions 
to discriminate subtle variation during the decision-making. For Aircraft, it is clear that the cluster compactness and separability improve with increasing $K$, but not much with Flowers and Pets dataset (also presented in Table \ref{tab:K}). The explanation is given in the following ablation study. More visualizations are shown in the supplementary.  
%
%
%
%
\begin{figure*}[t]
\centering
\subfigure{ \includegraphics[width= 0.95\textwidth]{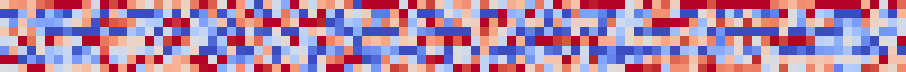}}
    \\ (a) Aircraft - 100 classes, \#cluster $K=8$ \\
 \subfigure{ \includegraphics[width= 0.95\textwidth]{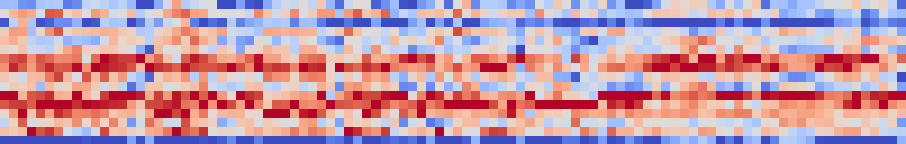}}\\ (b) Aircraft - 100 classes, \#cluster $K=16$ \\

\caption{Visualization of the cluster-specific contributions (i.e. weights, {\color{cyan}cool} to {\color{red}warm} $\Rightarrow$ less to more) from the graph representation of regions towards a given category during the spectral clustering-based graph pooling. The y-axis represents $K$ and the x-axis shows the number of classes. Each column is different, representing the feature discriminability during the decision making process. All test images from the Aircraft are used to compute weights.}
    \label{fig:clust1}
\end{figure*}

\vspace{- 1 cm}

\subsection{Ablation study}\label{sec:abl}
We study the impact of the following key aspects using Aircraft, Flowers, and Pets datasets. \textbf{1)} Effect of 
$K$ in (\ref{eqn:pred}) while forming a coarser (i.e., more abstract) graph representation to combine multiple graph structures via spectral clustering (Fig.\ref{fig:full_model}c). 
The impact of $K$ on accuracy (Table \ref{tab:K}) is dataset-dependent since the value for best accuracy on Aircraft ($K=40$) is different from Flowers and Pets ($K=8$). This can be linked to the dataset size and types. For example, the training size of Aircraft (6,667) 
is larger than the Flowers (2,040) and Pets (3,680). 
\textbf{2)} Performance linking number of regions ($R = \sum_{l=1}^LR_l$ in (\ref{eq_1})) in our multi-scale representation and is presented in Table \ref{tab:RoIs}. The number of regions is controlled by the region proposal algorithm's cell-size parameter in \cite{behera2020regional} (Section \ref{sec:regions}). \textbf{3)} The impact of the number of attention heads $H$ in (\ref{eqn:attn}) and per-head output dimension on accuracy is shown in Table \ref{tab:RoIs}. The model performs better for 3 heads with an output dimension of 512.  \textbf{4)} Finally, we evaluate the influence of the number of layers $L$ in our hierarchical representation and is provided in Table \ref{tab:layers}. A notable observation in the above ablative studies is the Aircraft's sensitivity to the values of $H$, $L$ and regions $R=\{R_l\}$. For example, accuracy drops from 94.9\% to 88.1\% when $|R|$ increases from 52 to 62, and it climbs from 83.7\% to 94.9\% for $H=2$ to $H=3$ (Table \ref{tab:RoIs}). Similarly, the accuracy increases from 88.7\% to 94.9\% for $L=2$ to $L=3$ (Table \ref{tab:layers}). Moreover, the baseline model (without our attention-driven hierarchical description) 
accuracy (79.5\%) is significantly low in comparison to Flowers (91.9\%) and Pets (91\%) FGVC datasets (Table \ref{tab:all_results}). This suggests that the Aircraft dataset is a difficult one and the sensitiveness could be due to the aeroplane shapes (higher aspect ratio) and the presence of significant perspective distortions in images (e.g. plane on ground, in mid-air, and landing/take off). As a result, the model struggles to find the optimal $H$, $L$ and $R$ parameters.    

\begin{table}[t]
    \centering
    \begin{tabular}{|l||c|c|c|c|c|c|c|c|c|}
        \hline
         \#Cluster (K) &  2 & 4& 8 & 16 & 20 & 32 & 36 & 40 & 48 \\
         \hline
         Aircraft &  - & - &  91.6 & 93.6 & 94.1 & 94.6 & 94.6 & \textbf{94.9} & 94.2  \\
         Flowers &  95.8 & 98.3 & \textbf{98.7} &\textbf{98.7} &98.5 &98.6 &- &- &-  \\
         Pets & 97.2 & 97.5 & \textbf{98.1} & 97.7 & 97.5 & 97.5 & - & - &- \\
         \hline
    \end{tabular}
    \caption{Accuracy (\%) with varying number of clusters ($K$).}
    \label{tab:K}
\end{table}

\begin{table}[]
\centering
\begin{tabular}{|l||p{5 mm}|p{5 mm}|p{7 mm}|p{5 mm}|p{5 mm} ||p{5 mm}|p{5 mm}|p{5 mm}|| p{5 mm}|p{7 mm}|p{5 mm}|}
\hline
Dataset &
\multicolumn{5}{c||}{Regions } & \multicolumn{3}{c||}{Head (ch 256)} & \multicolumn{3}{c|}{Head (ch 512) }     \\
 & 32& 43& 52 & 62& 74  & 2 & 3 & 4 & 2 & 3 & 4\\
\hline
Aircraft & 93.7 & 94.3 & \textbf{94.9} & 88.1 & 85.5  & 94.3 & 94.3 & 94.4 & 83.7 & \textbf{94.9} & 94.2 \\
Flowers &  97.2 & 98.0 & \textbf{98.7} & 96.8 & 94.9 & 98.4 &98.5 & 98.2 & 98.5 & \textbf{98.7} & 98.4 \\
Pets & 97.7 & 97.8 & \textbf{98.1} & 97.6 & 97.4 & 97.3 & 97.3 & 97.6 & 97.3 & \textbf{98.1} & 97.8 \\
\hline
    \end{tabular}
    \caption{Accuracy (\%) with varying number of regions in our multi-scale hierarchical structure and varying number attention heads and output channels per attention head.}
    \label{tab:RoIs}
\end{table}

\begin{table}[!h]

    \centering
    \begin{tabular}{|c||c|c|c|c||c|c|}
        \hline
        \#Layers $L$ & Aircraft & Flowers & Pets & CIFAR-100 & Trainable  & GFLOPs \\
         & & & & & Parameters & \\
        \hline
        1 &77.4 &89.4 &93.5 & 81.7 &27,493,316 &11.849 \\
        2 &88.7 &98.0 &97.5 & 82.9 &31,691,204 &13.213 \\
              \textbf{3} &\textbf{94.9} & \textbf{98.7} & \textbf{98.1} & \textbf{83.8} & \textbf{36,082,880} &\textbf{13.215} \\
        4 &91.9 &94.2 &93.8 & 83.5 &40,086,980 &13.671 \\
        5 &91.4 &93.8 &92.0 & 82.1 &44,284,868 &13.673 \\
        \hline
\end{tabular}
\caption{
Influence of the number of layers $L$ in ({\color{red}1}) in our hierarchical representation. We use the optimal number of attention heads ($H$=3) and its output dimension is 512. It is evident that the accuracy increases with the number of layers $L$, but after $L$=3, it starts decreasing. This trend is in all 4 datasets.  
The trainable parameters and GFLOPs increase with $L$ as well.} 
\label{tab:layers}
    
    \vspace{-0.4cm}
\end{table}

\subsection{More Experimental Results in Supplementary} 
More experimental analysis is given in the supplementary: 1) Dataset description in Table \textcolor{red}{5}; 
and 2) further results of Table \textcolor{red}{3} comparing \textit{concatenation with averaging} the outputs from multi-head attention is in Table \textcolor{red}{6}.
It also includes detailed study on complexity analysis in Table \textcolor{red}{7}, 
top-N accuracy (\%) in Table \textcolor{red}{8}, example of the regions linking various layers to visualize hierarchy in Fig.\textcolor{red}{4-5}, cluster-specific contributions of the graph-based regions in Fig.\textcolor{red}{6-8}, and t-SNE analysis of layer-wise attention heads are shown in Fig.\textcolor{red}{9-12} with better clarity. 

\vspace{-0.1cm}
\subsubsection{Further Link} A notable characteristics of our model from some recent works \cite{yang2021re, shroff2020focus, he2019and, behera2021context} is that it does not explicitly localize or search for salient regions/parts. This, in fact, is a forte of our model; making it capable of attending to all possible regions and model their short- and long-range dependencies, since this is possibly a best way to overcome a high intra-class and low inter-class variances in FGVC due to occlusions, deformation, and illuminations. We would like to emphasize that graph-based attention-driven relation-aware expressive representation gained improvement in person re-identification \cite{zhang2020relation}, visual question answering \cite{li2019relation, narasimhan2018out}, social relationship understanding \cite{wang2018deep}, and human-object interaction \cite{qi2018learning}.
\vspace{- 0.3 cm}
\section{Conclusion}
To address the problem of long-range dependencies to capture the subtle changes, we have proposed an end-to-end deep model by exploring Graph Convolutional Networks to represent regions as a set of complete graph structures. These regions are multi-scale and arranged hierarchically consisting of smaller (closer look) to larger (far look) size. To emphasize regions, the dependencies among regions are modeled using attention-driven message propagation that explores the finer to coarser graph structures. Our model's performance demonstrates its effectiveness in advancing both FGVC and generic visual classification.  

\vspace{0.3 cm}

\noindent\textbf{Acknowledgement:} This research is supported by the UKIERI (CHARM) under grant DST UKIERI-2018-19-10.
\bibliography{egbib}

\clearpage

\begin{center}
\Huge{Supplementary Document}
\end{center}
\vspace{ 1 cm}

In this supplementary document, the remaining quantitative and qualitative results are presented. A few additional supporting experimental results are also included. \\
\noindent\textbf{Dataset Description:} Details about the datasets with the state-of-the-arts (SotA), and the accuracy of proposed method are given in Table \ref{table:dataset} (cf. lines: 245-247, and 390 in the paper).
\begin{table} [ht]
\vspace{-0.4cm}
\begin{center}
\begin{tabular}{|c| c| c| c|  c| c||}
 \hline
Dataset &  \#Train / \#Test & \#Class & \text{  SotA} & Proposed \\
\hline
Aircraft-100 [\textcolor{red}{36}] & 6,667 / 3,333 &100 & CAP [\textcolor{red}{4}]: \textbf{94.9} 
& \textbf{94.9} \\ 
Flowers-102 [\textcolor{red}{38}] & 2,040 / 6,149 &102  & CAP [\textcolor{red}{4}]: 97.7 
& \textbf{98.7} \\
Oxford-IIIT Pets-37 [\textcolor{red}{39}] & 3,680 / 3,669 &37  & CAP [\textcolor{red}{4}]: 97.3 
& \textbf{98.1} \\
\hline
CIFAR-100 [\textcolor{red}{30}] & 50,000 / 10,000 &100 & BOT [\textcolor{red}{77}]: 83.5 
& \textbf{83.8} \\ 
Caltech-256 [\textcolor{red}{19}] & 15,360 / 14,420 &256  & CPM [\textcolor{red}{18}]: 94.3 
&\textbf{96.2} \\
\hline
\end{tabular}
 \caption{For evaluation, datasets consisting of fine-grained (Aircraft-100, Flowers-102, and Pets-37) and generic (CIFAR-100 and Caltech-256) visual classification are used. Accuracy (\%) of our model in comparison to the best SotA.   
 }
 \label{table:dataset}
 \end{center}
 \end{table}

\vspace{-0.5 cm}
\noindent\textbf{Impact of the number of layers $L$:} We evaluate the influence of the number of layers in performance assessment of our hierarchical representation is given in Table \ref{tab:layers} (cf. lines: 388-389 and 391-392 in the article).
\begin{table*}[ht]
    \centering
    \begin{tabular}{|c||c||c||c||c||c|c|}
        \hline
        \#Layers $L$ & Aircraft & Flowers & Pets & CIFAR-100 & Trainable  & GFLOPs \\
         & & & & & Parameters & \\
        \hline
        1 &77.4 &89.4 &93.5 & 81.7 &27,493,316 &11.849 \\
        2 &88.7 &98.0 &97.5 & 82.9 &31,691,204 &13.213 \\
              \textbf{3} &\textbf{94.9} & \textbf{98.7} & \textbf{98.1} & \textbf{83.8} & \textbf{36,082,880} &\textbf{13.215} \\
        4 &91.9 &94.2 &93.8 & 83.5 &40,086,980 &13.671 \\
        5 &91.4 &93.8 &92.0 & 82.1 &44,284,868 &13.673 \\
        \hline
    \end{tabular}
    \caption{Ablation study 
    involving the the influence of the number of layers $L$ in ({\color{red}2}) (cf. lines:145-149 in the article) in our hierarchical representation. The results presented here use optimal number attention heads ($H$=3) and its output dimension is 512. It is evident that the accuracy increases with the number of layers $L$, but after $L$=3, it starts decreasing. This trend has been observed in all 4 datasets (Aircraft-100, Flowers-102, Pets-37, and CIFAR-100). Moreover, the model's trainable parameters and GFLOPs increases with $L$ values. The model performs best for $L$=3. }
    \label{tab:layers}
\end{table*}

\noindent\textbf{Additional results of Table {\color{red}3} (concatenation vs averaging) in the main article:}  The remaining results of Table {\color{red}3} by comparing concatenation with averaging the outputs from \textit{multi-head attention} in ({\color{red}2}). It is found that the concatenation is better than the averaging. The results are provided in Table \ref{tab:attention_type} (cf. lines: 392-393 in paper). The performance of average aggregation increases with the number of heads. However, the computational complexity (number of trainable parameters and GFLOPs) also increases with the number of attention heads as shown in Table \ref{tab:model_complexity}. 
\begin{table*}[ht]
    \centering
    \begin{tabular}{|c|c|c|c|c|c|}
        \hline
        Attention Type & Attention Heads & Aircraft & Flowers & Pets \\
        \hline
     \textbf{Concatenate} & \textbf{3} & \textbf{94.9} & \textbf{98.7} & \textbf{98.1} \\ \hline
        Average & 2 & 85.5 & 97.8 & 97.3 \\
        Average & 3 & 90.2 & 98.5 & 97.6 \\
        Average & 4 & 90.8 & 98.7 & 98.0 \\
        \hline
    \end{tabular}
    \caption{More results of Table 3 in the main paper using average of different attention head's outputs versus their concatenation. The concatenation result is presented in Table 3, and the best accuracy is achieved using 3 attention heads with output dimension of 512. In this table, the accuracy with \textit{averaging} is presented. It is observed that the performance using averaging increases with the number of attention heads. However, the model complexity (number of trainable parameters and GFLOPs) also increases with the number of attention heads as shown in Table \ref{tab:model_complexity}. Thus, concatenation using an optimal number of attention heads ($H$=$3$) is preferred. This has been specified in the main paper (cf. lines: 202-203).}
    \label{tab:attention_type}
\end{table*}

\noindent\textbf{Model complexity:} We could not include more details about the model complexity of our method in the main paper (Section {\color{red}4.2}). It is presented here in Table \ref{tab:model_complexity} (cf. lines: 346 \& 393).   
\begin{table*}[ht]
    \centering
    \begin{tabular}{|c|c|c|c|c|c|}
        \hline
        Clusters  & Channels & Attention  & Trainable  & GFLOPs & Per-frame inference time \\
        $K$ &  &Heads &Parameters & &  in milliseconds (ms)\\
        \hline
        8  &256 &2 &27,473,088 &13.206 &8.0 \\
        8  &256 &3 &29,428,928 &13.208 &8.5 \\
        8  &256 &4 &31,515,840 &13.210 &8.6 \\
        8  &512 &2 &31,515,840 &13.210 &8.5 \\
        8  &512 &3 &36,082,880 &13.215 &8.5 \\
        8  &512 &4 &41,174,208 &13.220 &8.6 \\
        16 &512 &3 &36,095,176 &13.219 &8.5 \\
        20 &512 &3 &36,101,324 &13.222 &8.6 \\
        32 &512 &3 &36,119,768 &13.229 &8.6 \\
        36 &512 &3 &36,125,916 &13.231 &8.6 \\
        40 &512 &3 &36,132,064 &13.233 &8.6 \\
        48 &512 &3 &36,144,360 &13.238 &8.7 \\
         \hline
    \end{tabular}
    \caption{Statistics about how the various hyper-parameters ($\#K, \#H$, and the dimension of $H$) affect the complexity of our model. This has been mentioned in line-346 of the main article. The number of clusters $K$ in soft clustering-based graph pooling does have a little impact on the model complexity (bottom six rows). The number of attention heads and their output dimensions (256 or 512) influence the complexity i.e., higher number of attention heads combined with larger dimension increase the complexity. However, there is a little impact of these values on GFLOPs and inference time in milliseconds.}
    \label{tab:model_complexity}
\end{table*}

\vspace{1cm}\noindent\textbf{Top-N Accuracy (\%):} We have also evaluated the proposed approach using top-N accuracy metric on Aircraft-100 [{\color{red}36}], Oxford-Flowers-102 [{\color{red}38}], Oxford-IIIT Pets [{\color{red}39}], CIFAR-100 [{\color{red}30}], and Caltech-256 [{\color{red}19}] datasets. Our model's performance is presented in Table \ref{tab:top_n_acc} (cf. lines: 394-395 in the paper). All datasets except CIFAR-100, the top-2 accuracy is around 99\%. Moreover, their top-5 accuracy is nearly 100\%. It  clearly reflects the efficiency of our proposed metohd to enhance the performance of both FGVC and generic object recognition.

\begin{table*}[h]
    \centering
    \begin{tabular}{|c|c|c|c|c|}
        \hline
        Dataset & Top-1 Acc & Top-2 Acc & Top-5 Acc & Top-10 Acc \\
        \hline
        Aircraft-100 & 94.9 & 98.8 & 99.6 & 99.8 \\
        Flowers-102 & 98.7 & 99.6 & 99.9 & 100.0 \\
        Pets-37 & 98.1 & 99.8 & 100.0 & 100.0 \\
        Caltech-256 & 96.2 & 99.0 & 99.7 & 99.8 \\ 
        CIFAR-100 & 83.8 & 89.3 & 92.0 & 93.6 \\
         \hline
    \end{tabular}
    \caption{Top-N accuracy (in \%) of the proposed model using optimal number of attention heads $H$=3 with output dimensions of 512 and $L$=3 layers in the hierarchical representation. The top-2 accuracy is around 99\% except CIFAR-100. Similarly, the top-5 accuracy is nearly 100\% (except CIFAR-100). This shows the effectiveness of the proposed model.}
    \label{tab:top_n_acc}
\end{table*}

\vspace{ 0.5 cm}
\begin{center}
    \textbf{Visualization and Analysis}
\end{center}

We have provided additional qualitative results of our method which are mentioned in the Section \textcolor{red}{4.2}. \\

\noindent {1) Example of the regions linking various layers to visualize the hierarchical structure is shown in Fig. \textcolor{red}{4-5}.} \\

\noindent {2) Cluster-specific contributions of the graph-based regions are shown in Fig. \textcolor{red}{6-8}.} \\

\noindent {3) t-SNE [{\color{red}50}] analysis of layer-wise attention heads are shown  in Fig. \textcolor{red}{9-12}.}

\clearpage

\begin{figure*}[ht]
\centering
\subfigure{\includegraphics[width=0.3\textwidth] {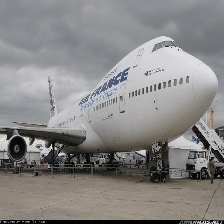}}\\ (a) An example image from the Aircraft dataset \\
\subfigure{\begin{tabular}[b]{c}
    \includegraphics[width=0.1\textwidth] {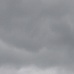}
    \includegraphics[width=0.1\textwidth] {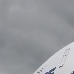}
    \includegraphics[width=0.1\textwidth] {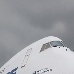}
    \includegraphics[width=0.1\textwidth] {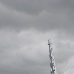}
    \includegraphics[width=0.1\textwidth] {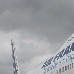}
    \includegraphics[width=0.1\textwidth] {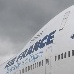}
    \includegraphics[width=0.1\textwidth] {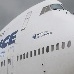}
    \includegraphics[width=0.1\textwidth] {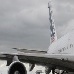}
    \includegraphics[width=0.1\textwidth] {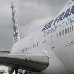}\\
    \includegraphics[width=0.1\textwidth] {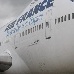}
    \includegraphics[width=0.1\textwidth] {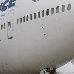}
    \includegraphics[width=0.1\textwidth] {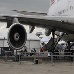}
    \includegraphics[width=0.1\textwidth] {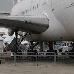}
    \includegraphics[width=0.1\textwidth] {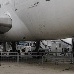}
    \includegraphics[width=0.1\textwidth] {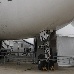}
    \includegraphics[width=0.1\textwidth] {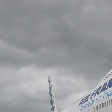}
    \includegraphics[width=0.1\textwidth] {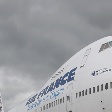}
    \includegraphics[width=0.1\textwidth] {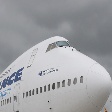}\\
    \includegraphics[width=0.1\textwidth] {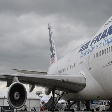}
    \includegraphics[width=0.1\textwidth] {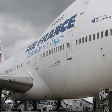}
    \includegraphics[width=0.1\textwidth] {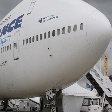}
    \includegraphics[width=0.1\textwidth] {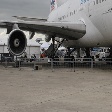}
    \includegraphics[width=0.1\textwidth] {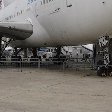}
    \includegraphics[width=0.1\textwidth] {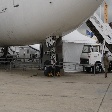}
    \includegraphics[width=0.1\textwidth] {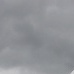}
    \end{tabular}}\\ (b) Layer 1 regions in our hierarchical structure \\
\subfigure{\begin{tabular}[b]{c}
    \includegraphics[width=0.145\textwidth] {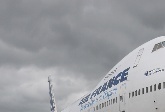}
    \includegraphics[width=0.15\textwidth] {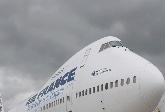}
    \includegraphics[width=0.15\textwidth] {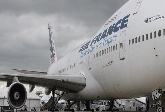}
    \includegraphics[width=0.15\textwidth] {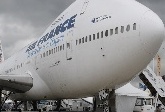}
    \includegraphics[width=0.15\textwidth] {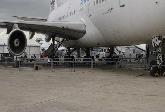}
    \includegraphics[width=0.15\textwidth] {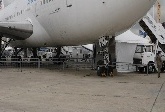}\\
    \includegraphics[height=0.15\textwidth] {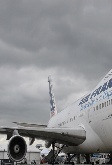}
    \includegraphics[width=0.1\textwidth] {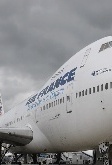}
    \includegraphics[width=0.1\textwidth] {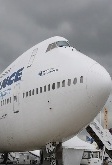}
    \includegraphics[width=0.1\textwidth] {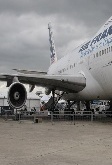}
    \includegraphics[width=0.1\textwidth] {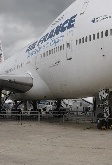}
    \includegraphics[width=0.1\textwidth] {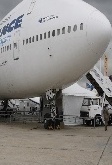}
     \includegraphics[height=0.15\textwidth] {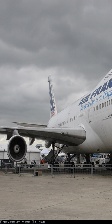}
    \includegraphics[height=0.15\textwidth] {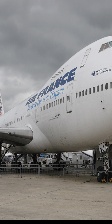}
    \includegraphics[height=0.15\textwidth] {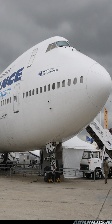}\\
    \includegraphics[width=0.09\textwidth] {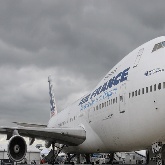}
    \includegraphics[width=0.09\textwidth] {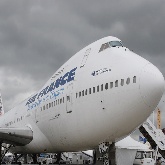}
    \includegraphics[width=0.09\textwidth] {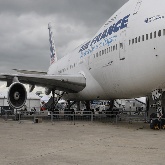}
    \includegraphics[width=0.09\textwidth] {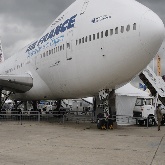}
    \includegraphics[width=0.18\textwidth] {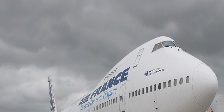}
    \includegraphics[width=0.18\textwidth] {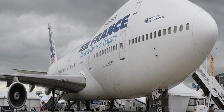}
    \includegraphics[width=0.18\textwidth] {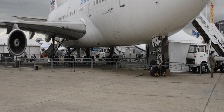}
    \end{tabular}} \\
    (c) Layer 2 regions in our hierarchical structure
    
    \caption{Layer-wise regions of fixed area but with different aspect ratios corresponding to a given hierarchical layer are generated using the region proposal algorithm in [{\color{red}3}]. In this example, we consider 3-layer hierarchical structure consisting of 52 regions. The original image is shown in (a).}
    \label{fig:fig_feat1}
\end{figure*}
\begin{figure*}[t]
\centering
\subfigure [ Layer 3 regions in our hierarchical structure ]{
    \includegraphics[width=0.22\textwidth] {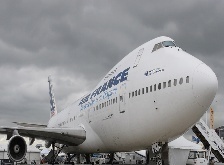}
    \includegraphics[width=0.22\textwidth] {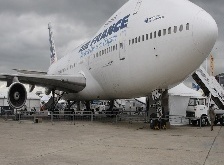}
    \includegraphics[width=0.12\textwidth] {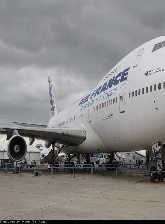}
    \includegraphics[width=0.12\textwidth] {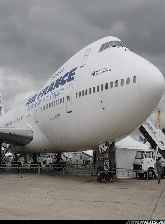}
    \includegraphics[width=0.165\textwidth] {Fig/aircraft_rois/aircraft_full_img.jpg}

    }
    \caption{Layer-wise regions of fixed area but with different aspect ratios corresponding to a given hierarchical layer are generated using the region proposal algorithm in [{\color{red}3}]. In this example, we consider 3-layer hierarchical structure consisting of 52 regions. The original image is shown in Fig. \ref{fig:fig_feat1} (a).}
\end{figure*}

\begin{figure*}[t]
    \centering
    \subfigure{ \includegraphics[width=\textwidth]{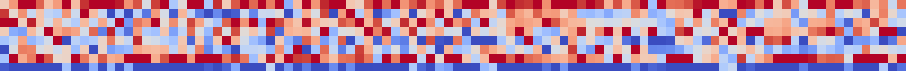}}
    \\ (a) Flowers: 102 classes, \#cluster $K=8$ \\
    \subfigure{ \includegraphics[width=\textwidth]{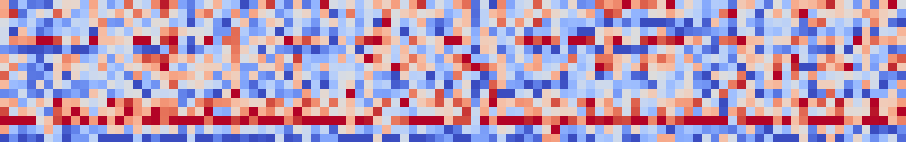}}\\  (b) Flowers: 102 classes, \#cluster $K=16$ \\
    \subfigure{\includegraphics[width=\textwidth]{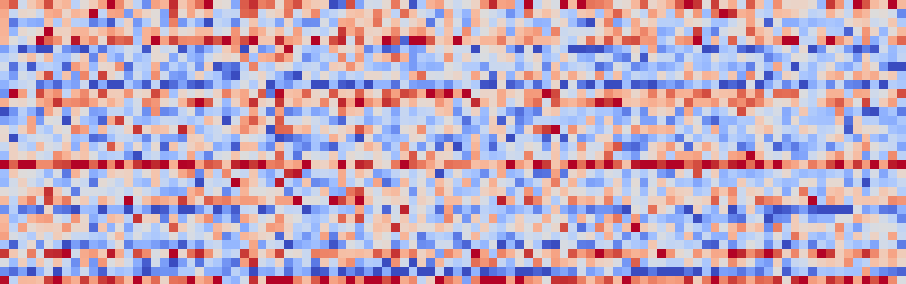}} \\ (c) Flowers: 102 classes, \#cluster $K=32$ \\
    \caption{
    Visualization of the  cluster-specific contributions (i.e. weights, {\color{cyan}cool} to {\color{red}warm} $\Rightarrow$ less to more) from the graph representation of regions towards a given category during the spectral clustering-based graph pooling. The y-axis (rows) represents $K$ (coarser representation) and the x-axis (cols) shows the number of classes. Each column is different, representing the feature discriminability during the decision making process. All test images from the \textbf{Oxford-Flowers-102} dataset are used to compute weights.}
    \label{fig:clust1}
    \vspace{-0.5cm}
\end{figure*}
\begin{figure*}[t]
    \centering
    \subfigure{ \includegraphics[width=\textwidth]{Fig/Aircraft_100class_8clusters_normalised_rot.png}}
    \\ (a) Aircraft - 100 classes, \#cluster $K=8$ \\
    \subfigure{ \includegraphics[width=\textwidth]{Fig/Aircraft_100class_16clusters_normalised_rot.png}}\\  (b) Aircraft - 100 classes, \#cluster $K=16$ \\
    \subfigure{\includegraphics[width=\textwidth]{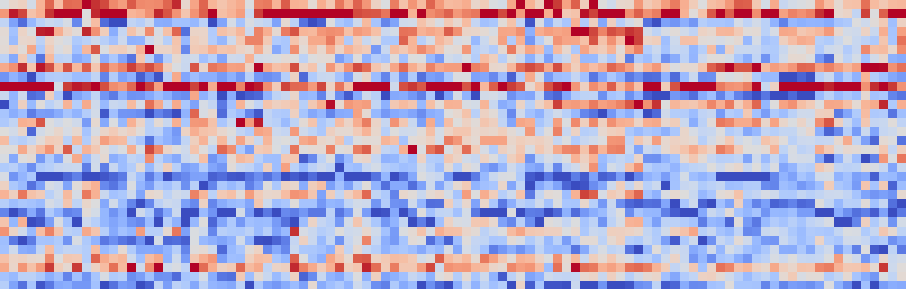}} \\ (c) Aircraft - 100 classes, \#cluster $K=32$ \\
    \caption{
    Visualization of the cluster-specific contributions (i.e. weights, {\color{cyan}cool} to {\color{red}warm} $\Rightarrow$ less to more) from the graph representation of regions towards a given category during the spectral clustering-based graph pooling. The y-axis (rows) represents $K$ (coarser representation) and the x-axis (cols) shows the number of classes. Each column is different, representing the feature discriminability during the decision making process. All test images from the \textbf{Aircraft-100} dataset are used to compute weights.}
    \label{fig:clust1}
    \vspace{-0.5cm}
\end{figure*}
\begin{figure*}[t]
    \centering
    \subfigure{ \includegraphics[width=0.73\textwidth]{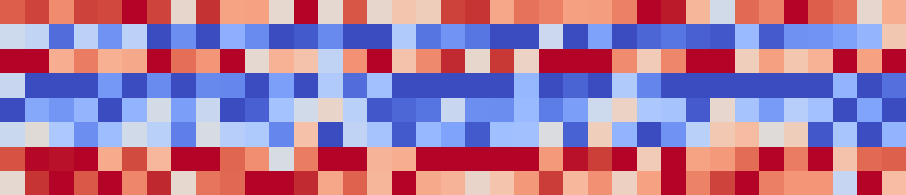}}
    \\ (a) Pets - 37 classes, \#cluster $K=8$ \\
    \subfigure{ \includegraphics[width=0.73\textwidth]{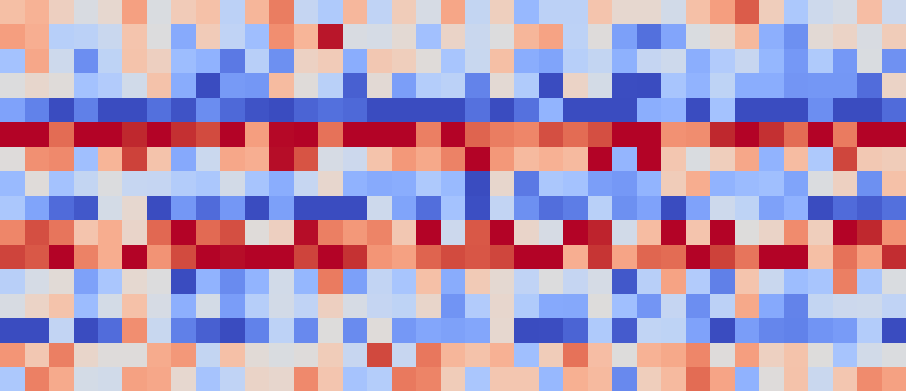}}\\  (b) Pets - 37 classes, \#cluster $K=16$ \\
    \subfigure{\includegraphics[width=0.73\textwidth]{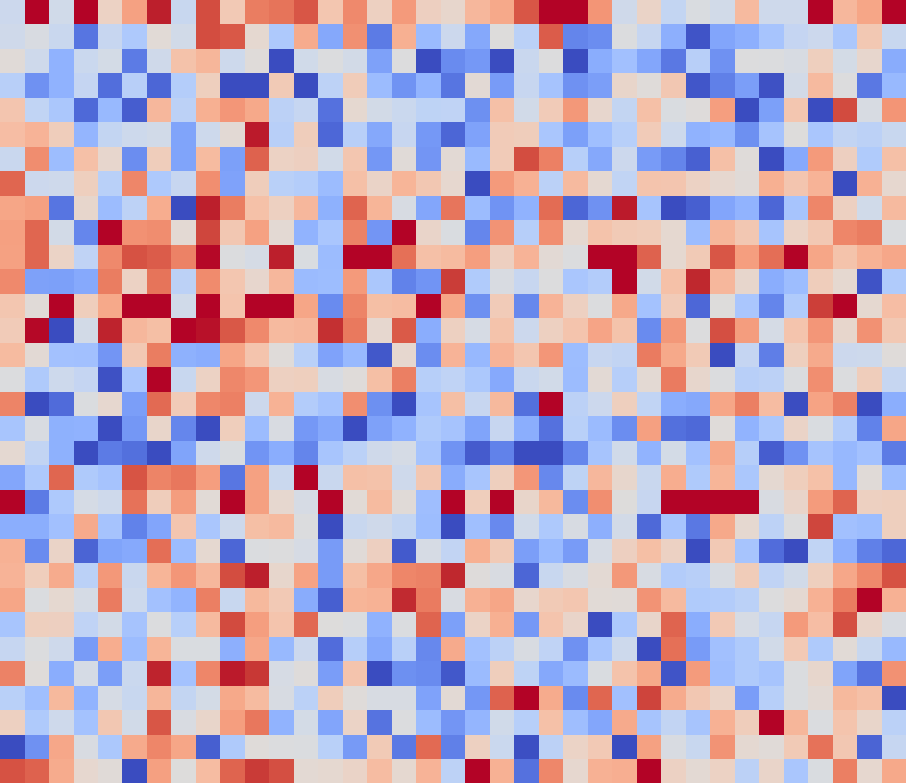}} \\ (c) Pets - 37 classes, \#cluster $K=32$ \\
    \caption{
    Visualization of the cluster-specific contributions (i.e. weights, {\color{cyan}cool} to {\color{red}warm} $\Rightarrow$ less to more) from the graph representation of regions towards a given category during the spectral clustering-based graph pooling. The y-axis (rows) represents $K$ (coarser representation) and the x-axis (cols) shows the number of classes. Each column is different, representing the feature discriminability during the decision making process. All test images from the \textbf{Oxford-IIIT Pets-37} dataset are used to compute weights.}
    \label{fig:clust1}
    \vspace{-0.5cm}
\end{figure*}
\begin{figure*}[t]
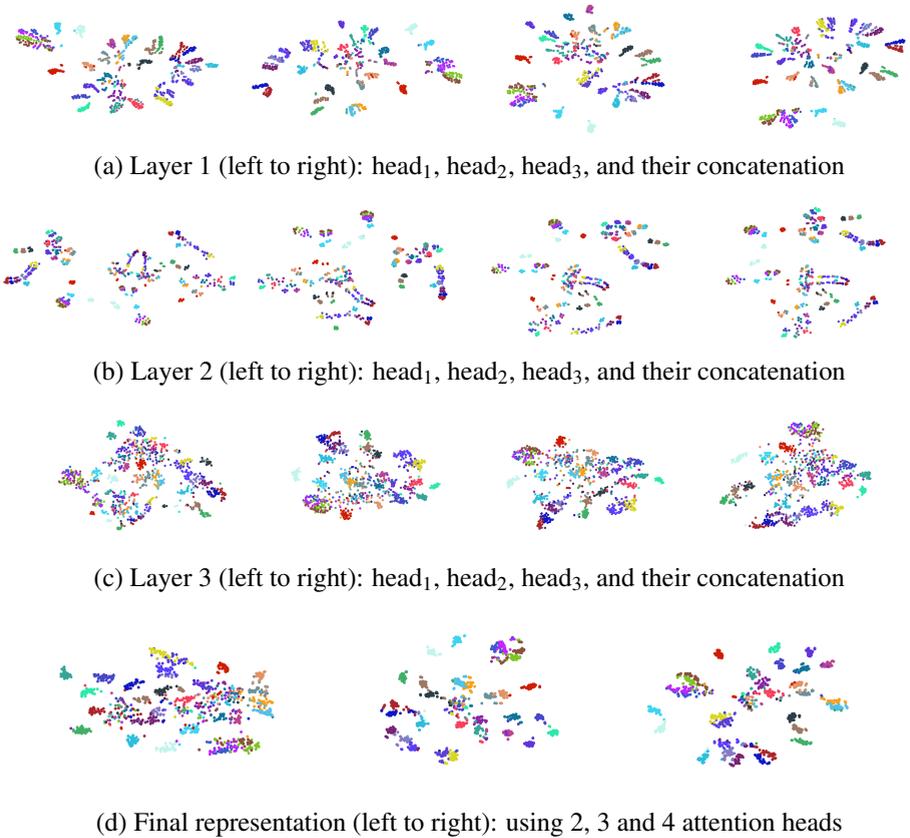

\centering
    \subfigure{\includegraphics[width=0.24\textwidth] {Fig/Aircraft_GATCONV_1_pooled_head1.png}\hspace{-0.3em}
    \includegraphics[width=0.24\textwidth] {Fig/Aircraft_GATCONV_1_pooled_head2.png}\hspace{-0.3em}
    \includegraphics[width=0.24\textwidth] {Fig/Aircraft_GATCONV_1_pooled_head3.png}\hspace{-0.3em}
    \includegraphics[width=0.24\textwidth] {Fig/Aircraft_GATCONV_1_pooled_all_heads.png}
    }\\  (a) Layer 1 (left to right): head$_1$, head$_2$, head$_3$, and their concatenation  \\
     \subfigure{\includegraphics[width=0.24\textwidth] {Fig/Aircraft_GATCONV_2_pooled_head1.png}\hspace{-.3em}
    \includegraphics[width=0.24\textwidth] {Fig/Aircraft_GATCONV_2_pooled_head2.png}\hspace{-0.5em}
    \includegraphics[width=0.24\textwidth] {Fig/Aircraft_GATCONV_2_pooled_head3.png}\hspace{-0.5em}
    \includegraphics[width=0.24\textwidth] {Fig/Aircraft_GATCONV_2_pooled_all_heads.png}
    }\\ (b) Layer 2 (left to right): head$_1$, head$_2$, head$_3$, and their concatenation \\ 
    \subfigure{\includegraphics[width=0.24\textwidth] {Fig/Aircraft_GATCONV_3_pooled_head1.png}\hspace{-0.3cm}
    \includegraphics[width=0.24\textwidth] {Fig/Aircraft_GATCONV_3_pooled_head2.png}\hspace{-0.3cm}
    \includegraphics[width=0.24\textwidth] {Fig/Aircraft_GATCONV_3_pooled_head3.png}\hspace{-0.3cm}
    \includegraphics[width=0.24\textwidth] {Fig/Aircraft_GATCONV_3_pooled_all_heads.png}
    }\\ (c) Layer 3 (left to right): head$_1$, head$_2$, head$_3$, and their concatenation \\
    \subfigure{\includegraphics[width=0.3\textwidth] {Fig/Aircraft_GATCONV_all_pooled_2_heads.png}\hspace{-0.5em}
    \includegraphics[width=0.3\textwidth] {Fig/Aircraft_GATCONV_all_pooled_3_heads.png}\hspace{-0.5em}
    \includegraphics[width=0.3\textwidth] {Fig/Aircraft_GATCONV_all_pooled_4_heads.png}
    } \\ (d) Final representation (left to right): using 2, 3 and 4 attention heads \\
    \caption{\textbf{For clarity, repetition of Fig. {\color{red}2} in the main article with larger size}. t-SNE [{\color{red}50}] visualization of class-specific discriminative feature representation of multi-scale hierarchical regions 
    using $H=3$ attention heads in ({\color{red}2}), and $L=3$ layers hierarchical structure in ({\color{red}1}). All test images from 30 randomly chosen classes within Aircraft dataset are used. 
    Attention head-specific plots are shown in $(a)\rightarrow (c)$, representing layers from smaller regions (a) to larger ones (c). It is evident that the discriminability of the features representing medium-size regions (b) $>$ small-size (a) $>$ large-size (c). 
    (d) shows the combined layers' representation using 2, 3 and 4 attention heads. More than 2 attention heads has shown better discriminability.}
    \label{fig:fig_feat}
    \vspace{-.5cm}
\end{figure*}

\begin{figure*}[t]
\centering
\subfigure{\includegraphics[width=0.3\textwidth] {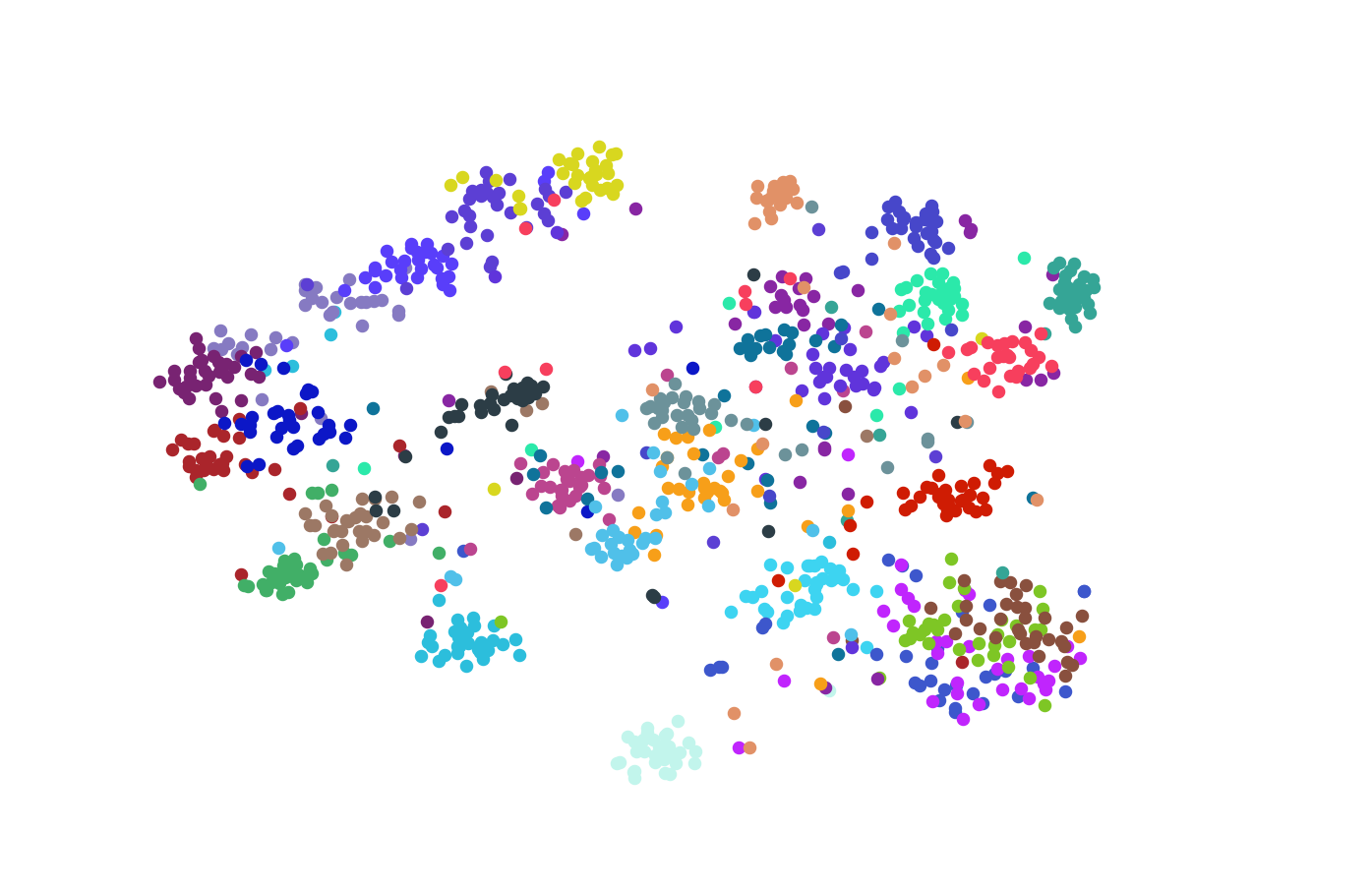}\hspace{-0.3em}
\includegraphics[width=0.3\textwidth] {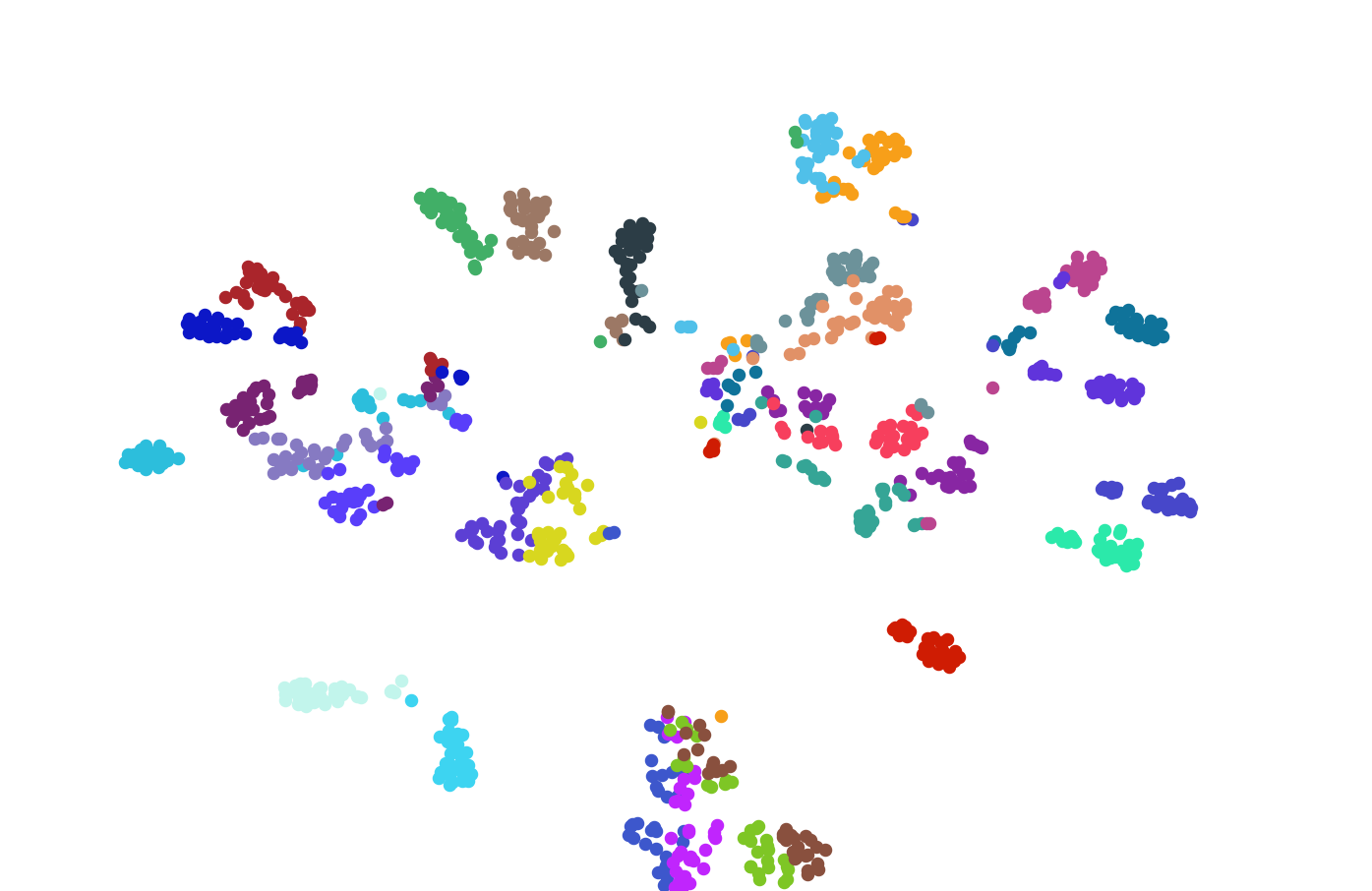}\hspace{-0.3em}
\includegraphics[width=0.3\textwidth] {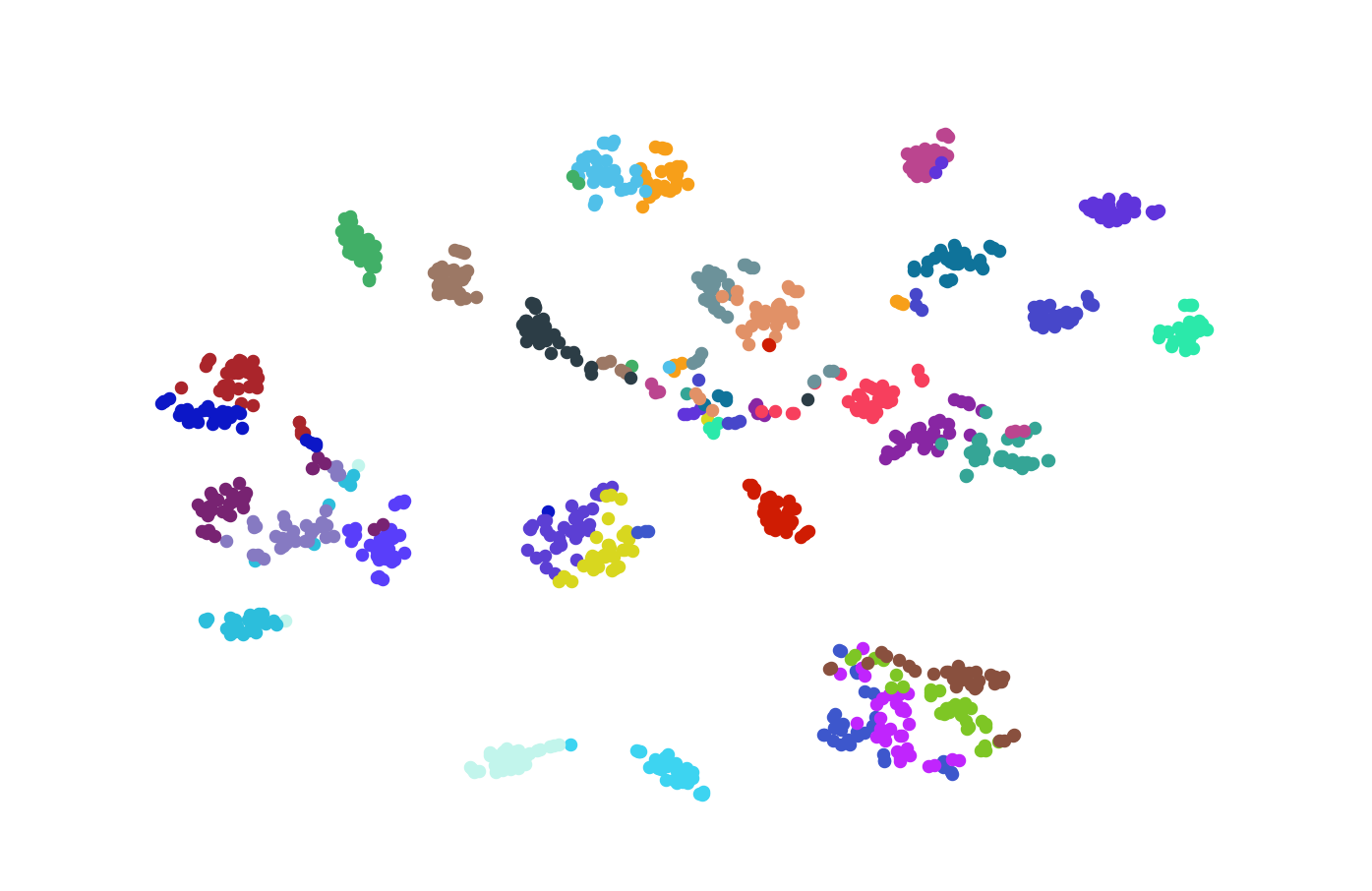}\hspace{-0.3em}
    }\\ (a) Aircraft (left to right): $K=8, 16, \text{and } 32$ \\
\subfigure{\includegraphics[width=0.3\textwidth] {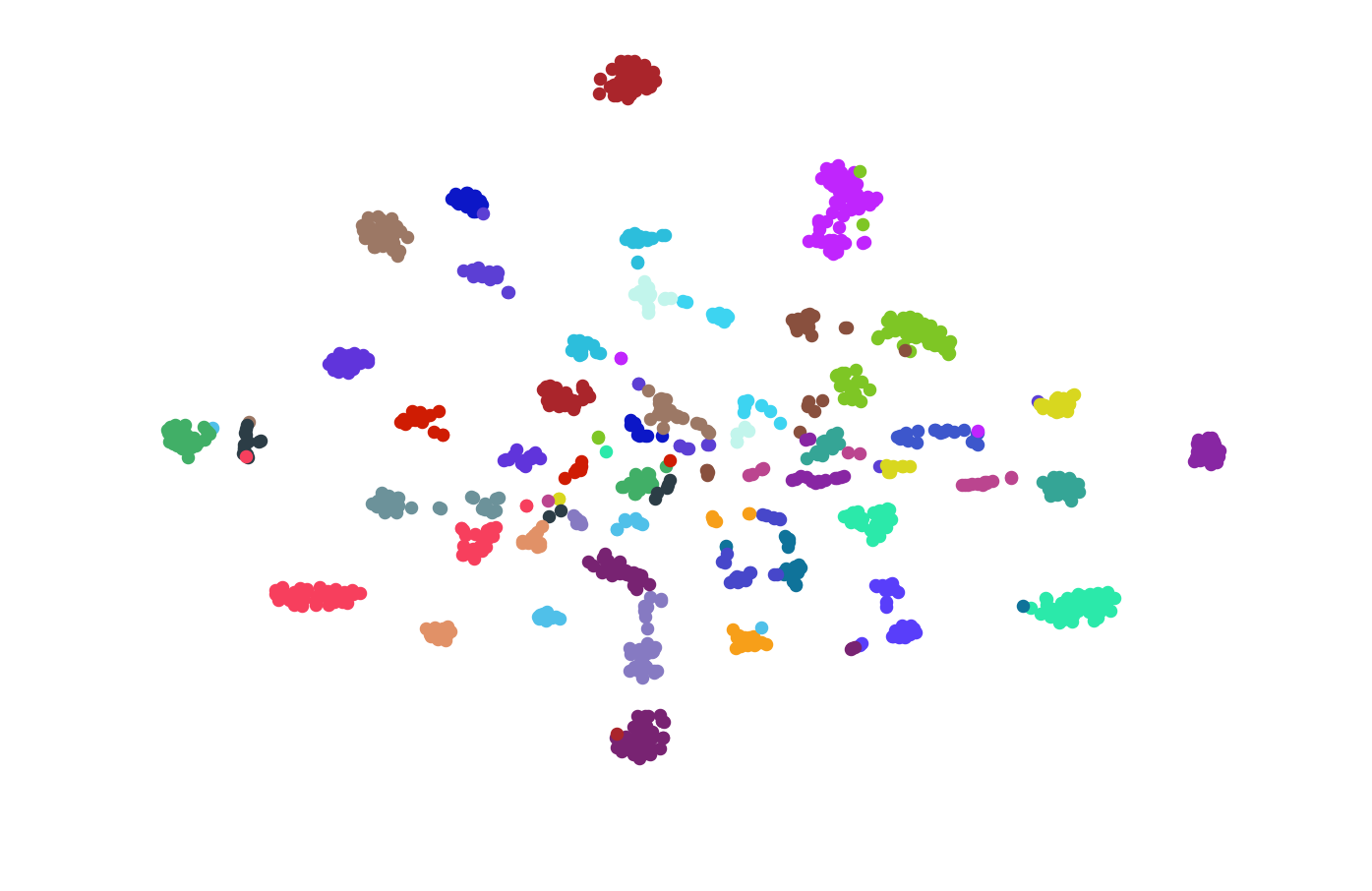}\hspace{-0.3em}
\includegraphics[width=0.3\textwidth] {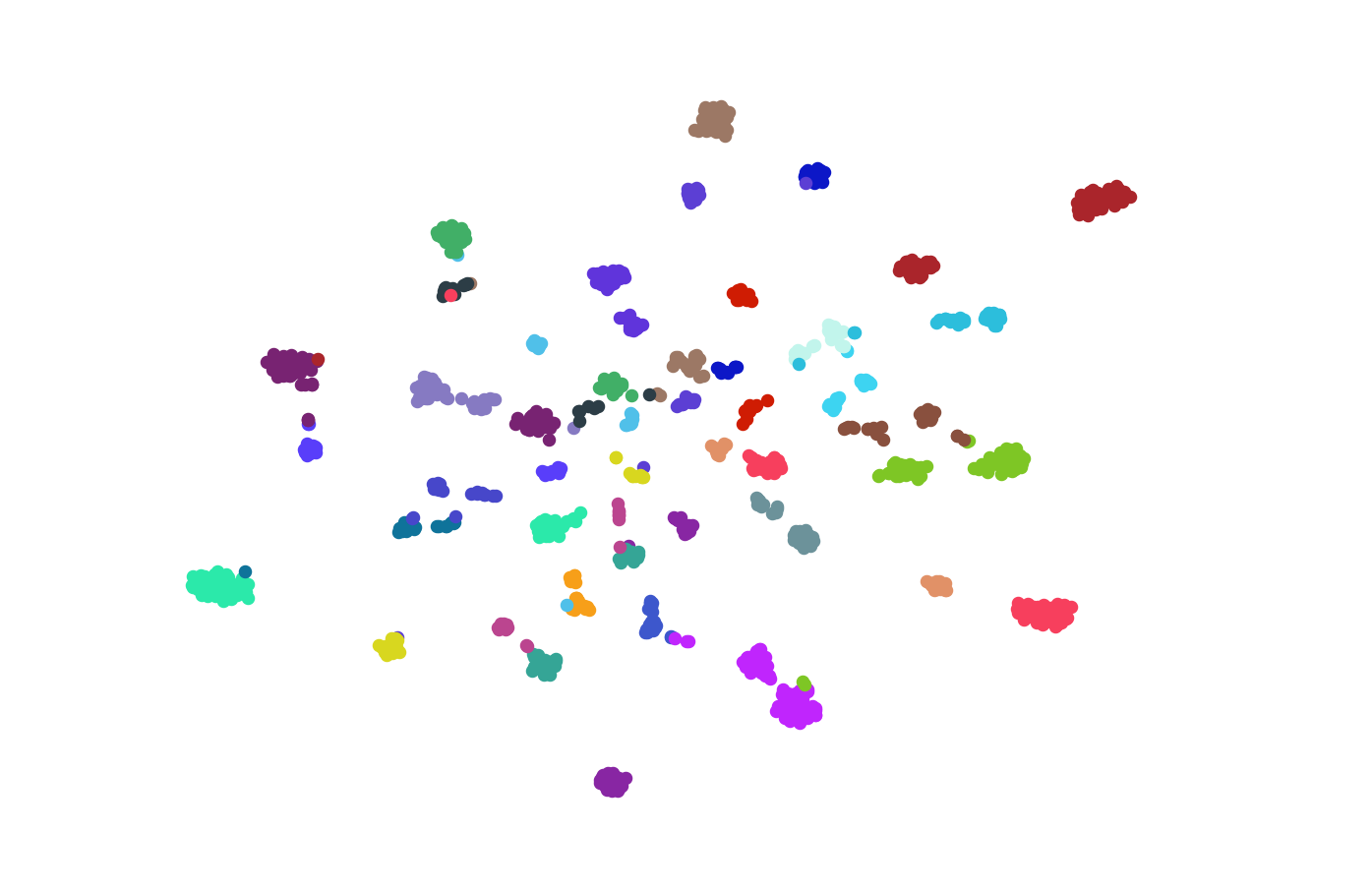}\hspace{-0.3em}
\includegraphics[width=0.3\textwidth] {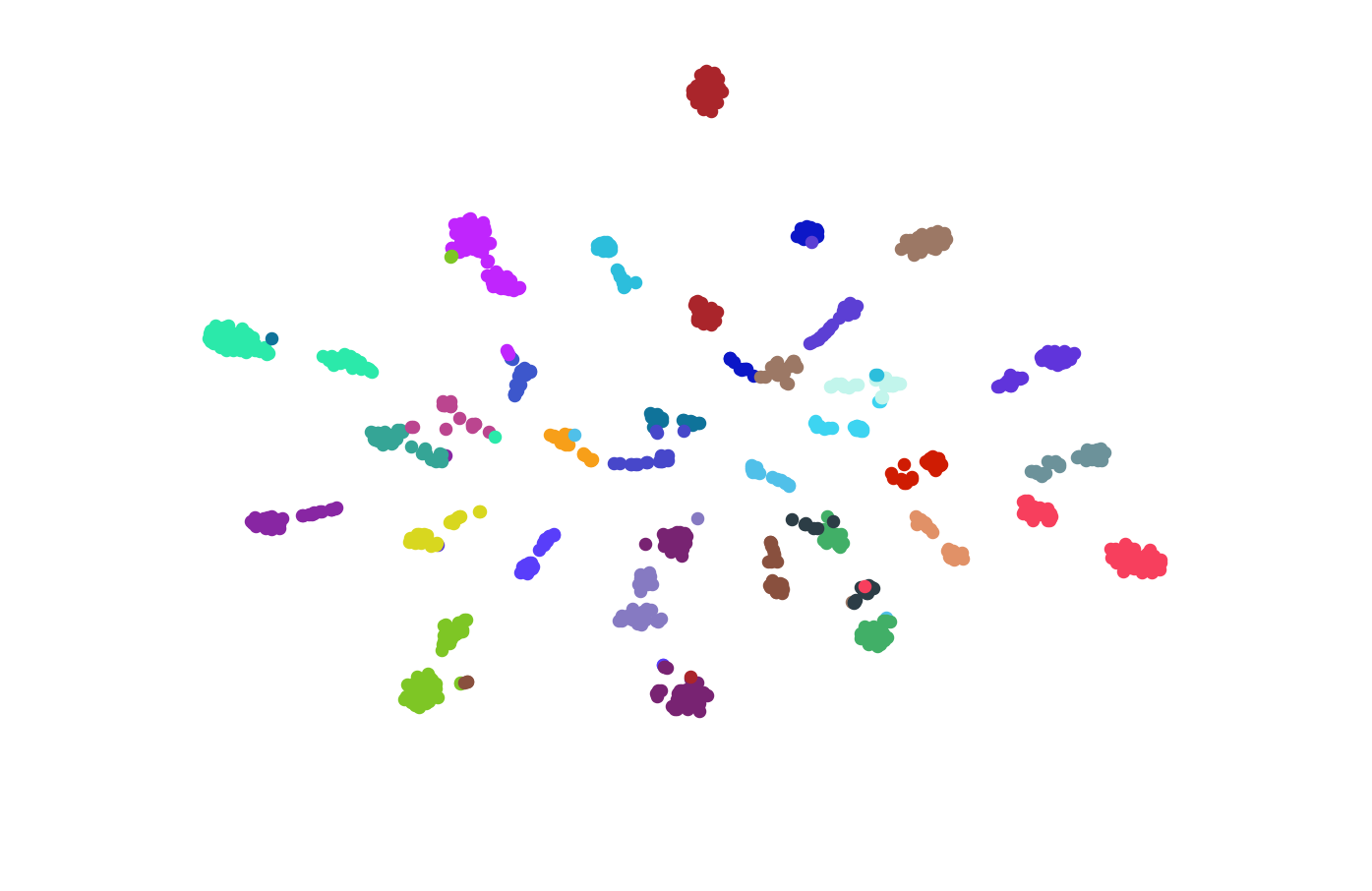}\hspace{-0.3em}
    }\\ (b) Flowers (left to right): $K=8, 16, \text{and } 32$ \\
\subfigure{\includegraphics[width=0.3\textwidth] {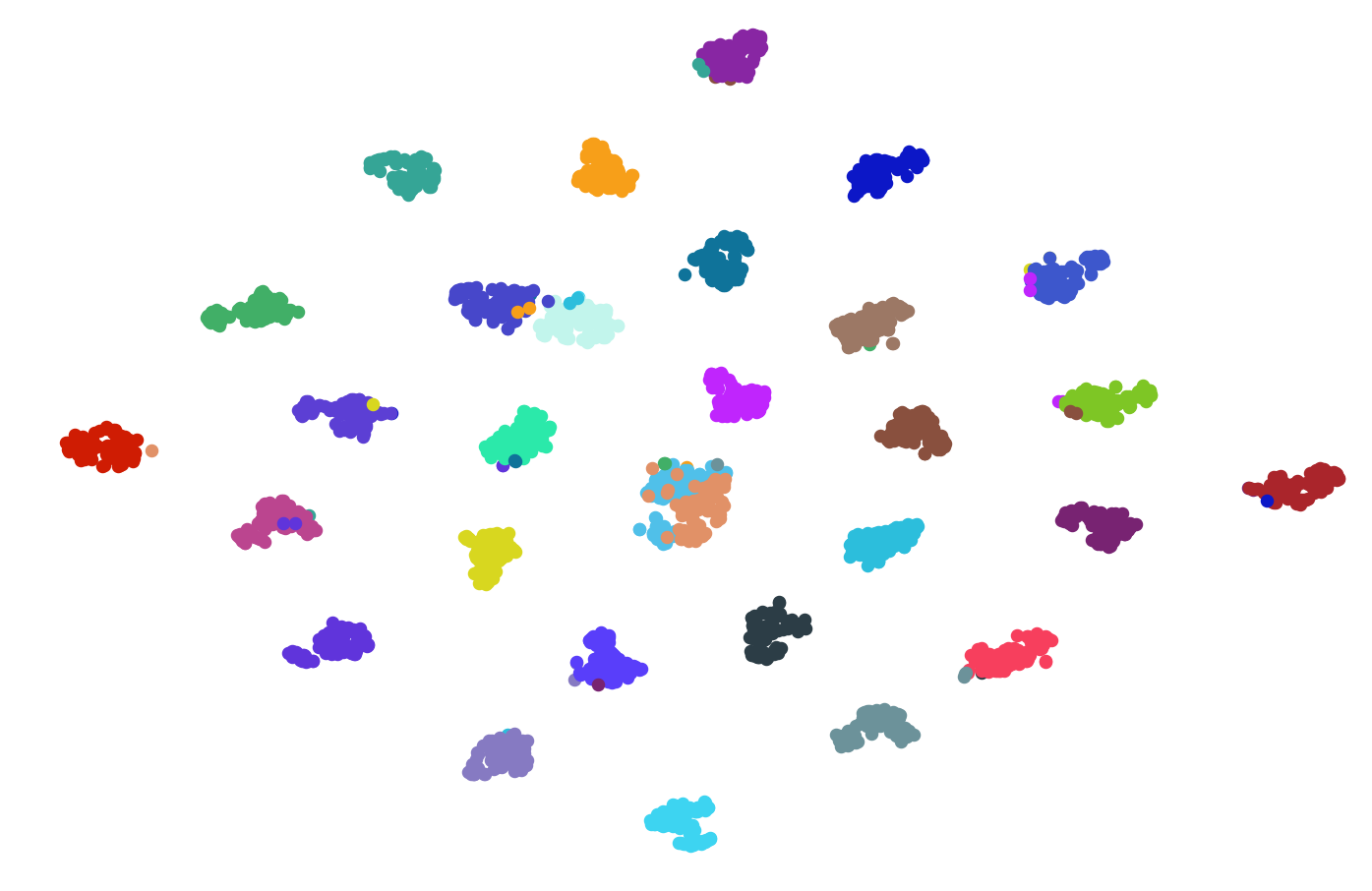}\hspace{-0.0em}
\includegraphics[width=0.3\textwidth] {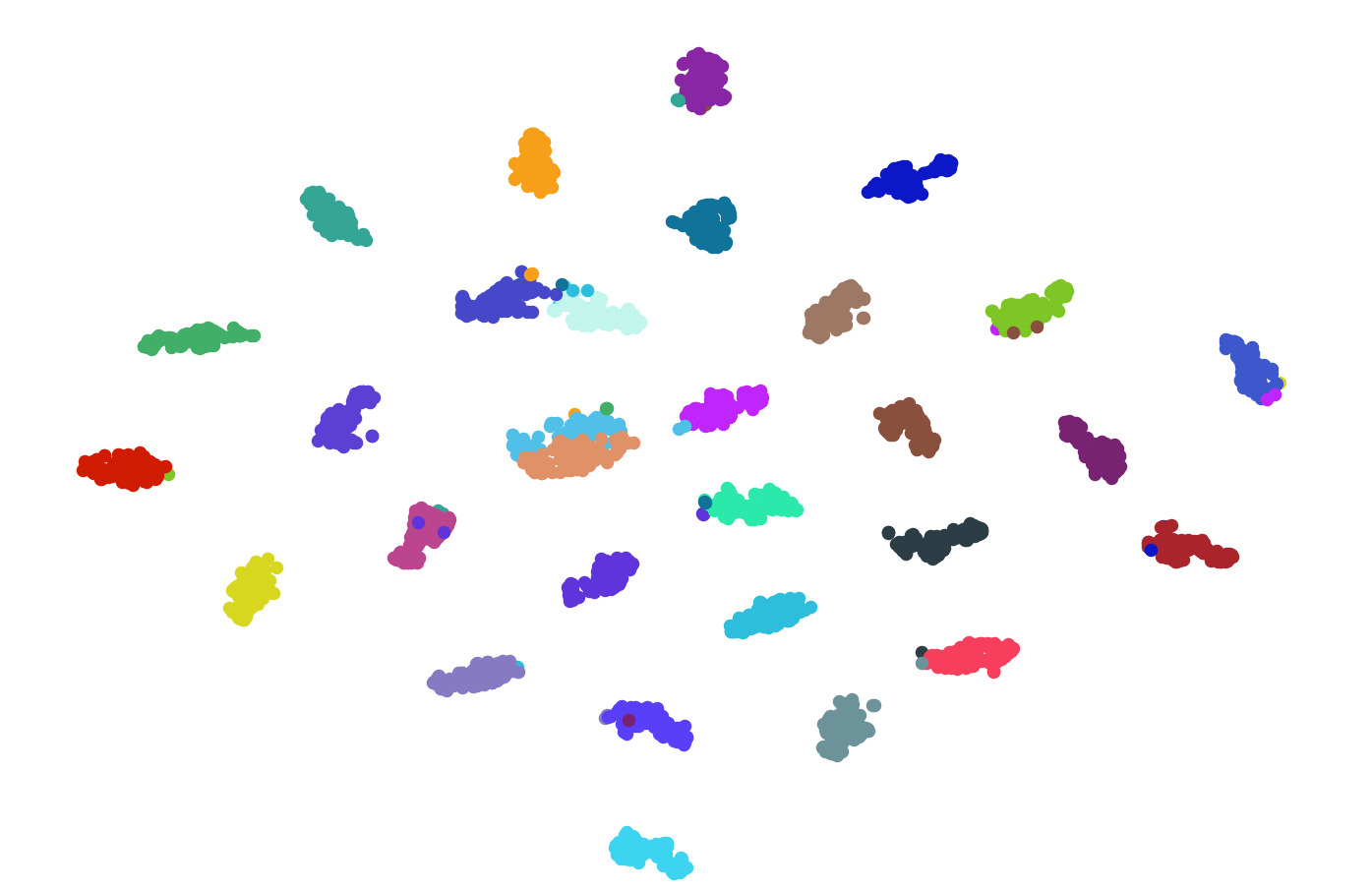}\hspace{-0.0em}
\includegraphics[width=0.3\textwidth] {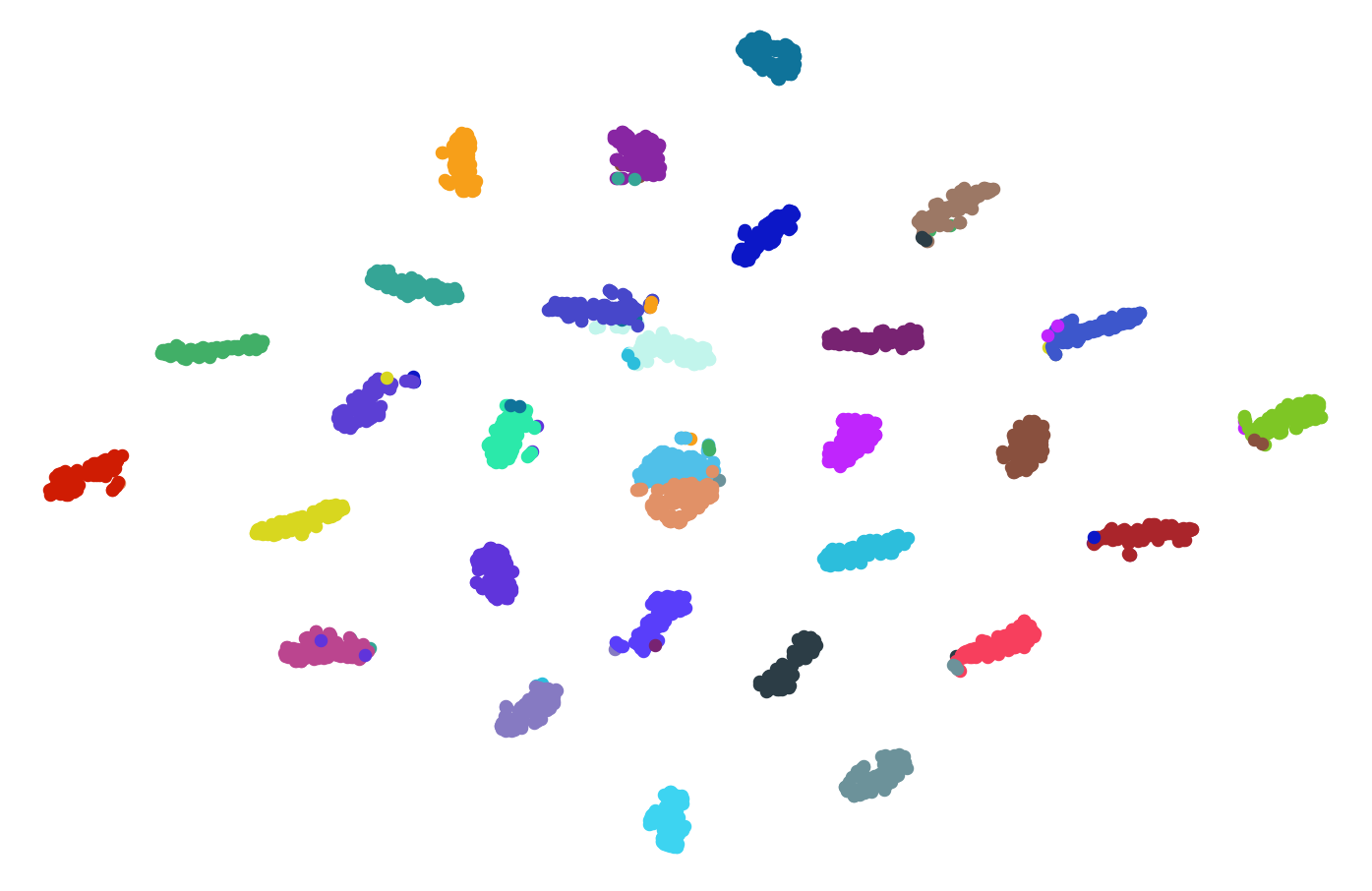}\hspace{-0.0em}
    } \\ (c) Pets (left to right): $K=8, 16, \text{and } 32$ \\
    \caption{
    t-SNE [{\color{red}50}] visualization of class-specific discriminative feature representing different clusters $K$ (coarser representation) to aggregate graph structure-driven regions via spectral clustering-based graph pooling (Fig. 1c). All test images from 30 randomly chosen classes within a dataset are used for the visualization.
    } 
    \label{fig:fig_tsne-k}
\end{figure*}
\begin{figure*}[t]
\centering
    \subfigure{\includegraphics[width=0.23\textwidth] {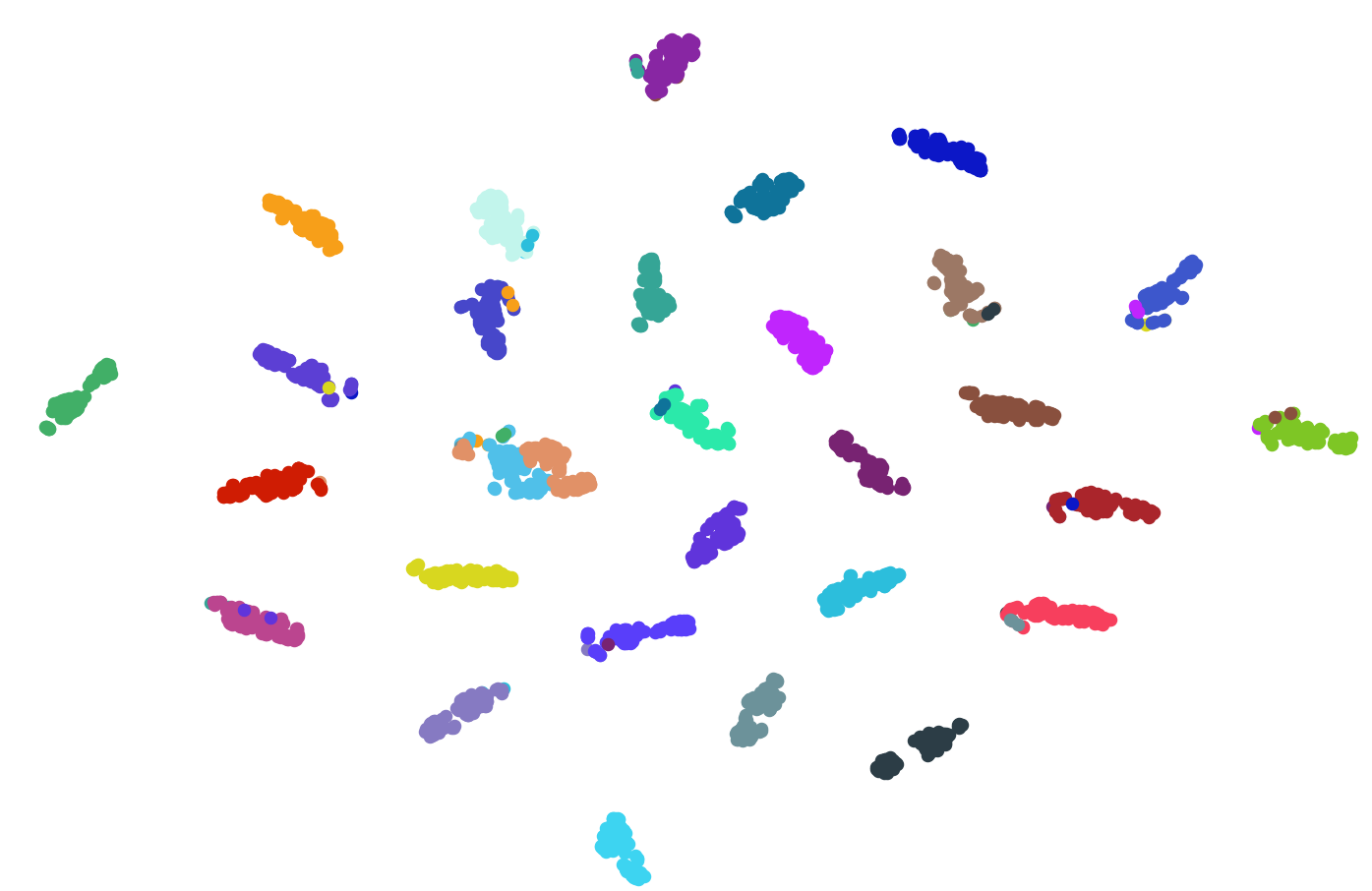}\hspace{-0.21em}
    \includegraphics[width=0.23\textwidth] {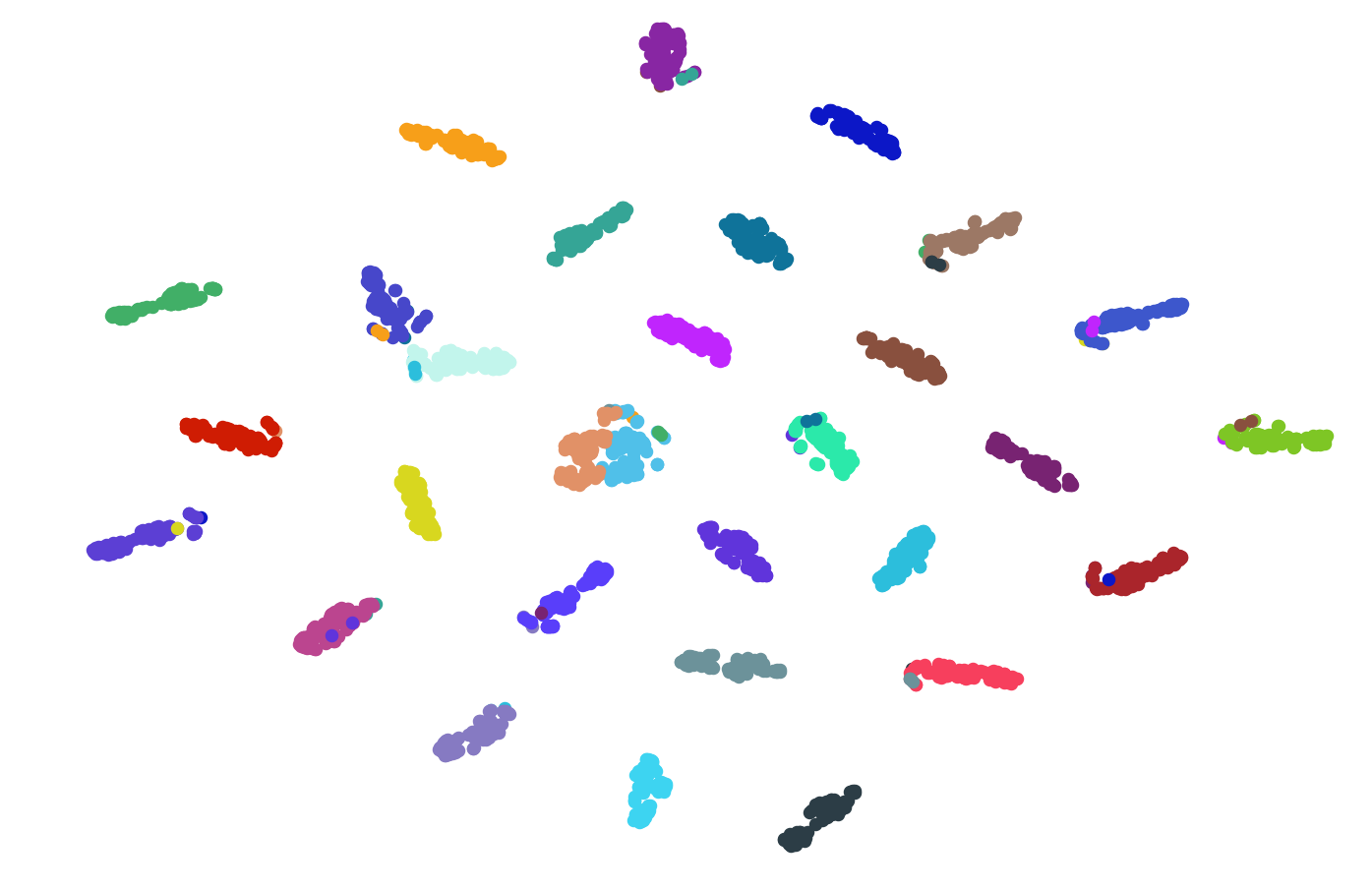}\hspace{-0.21em}
    \includegraphics[width=0.23\textwidth] {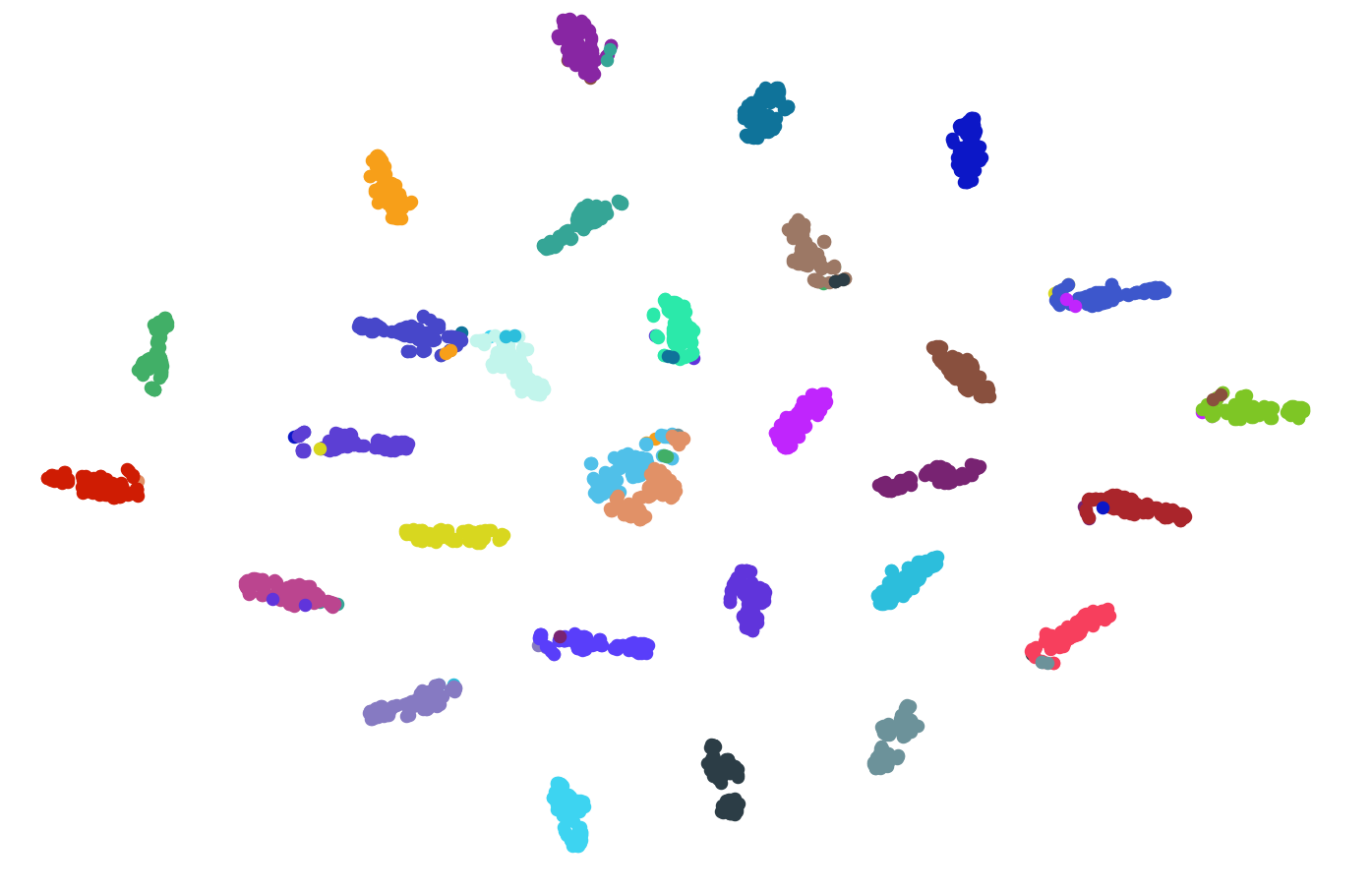}\hspace{-0.2em}
    \includegraphics[width=0.23\textwidth] {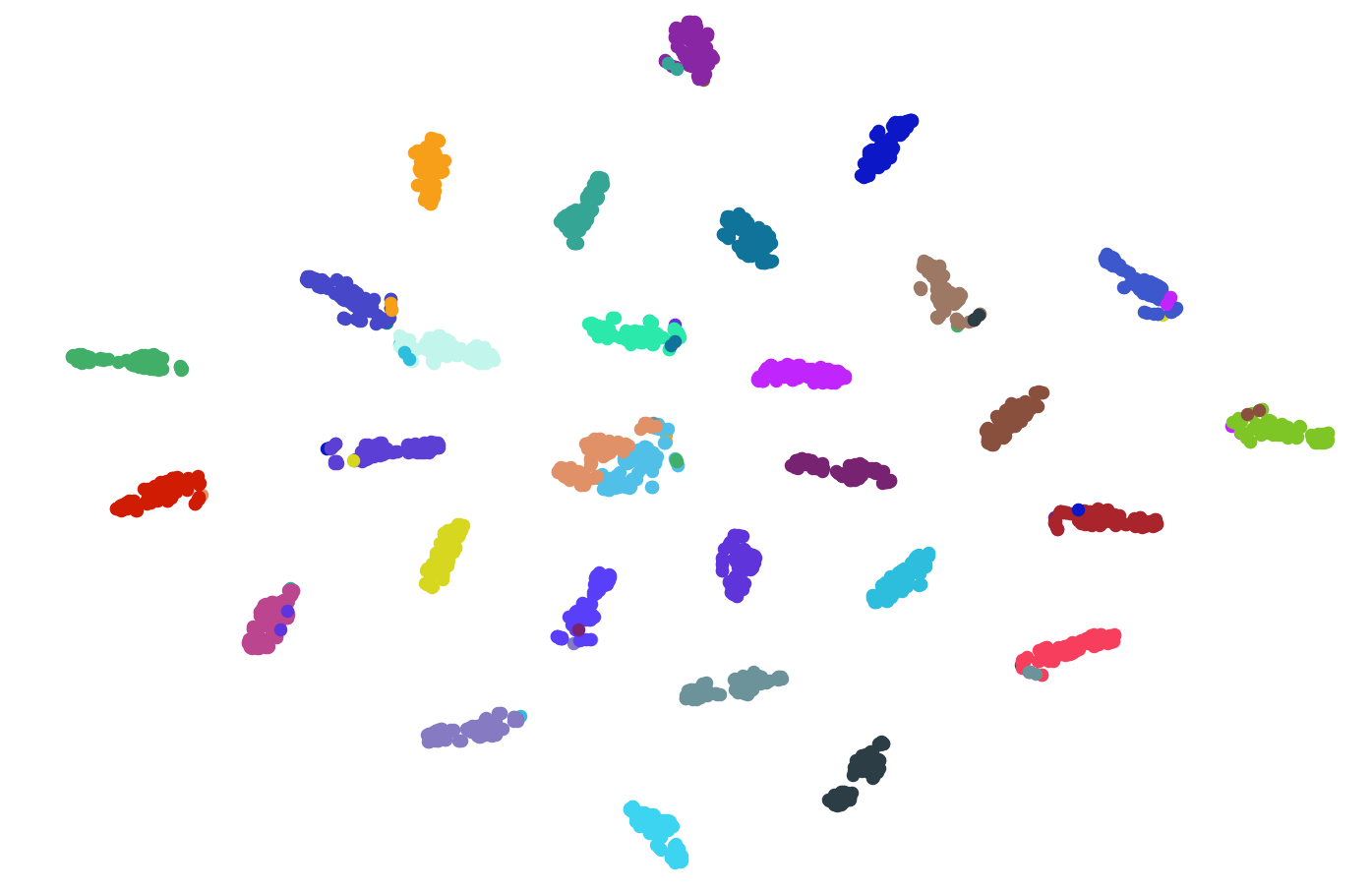}
    }\\ (a) Layer 1 (left to right): head$_1$, head$_2$, head$_3$, and their concatenation  \\
     \subfigure{\includegraphics[width=0.23\textwidth] {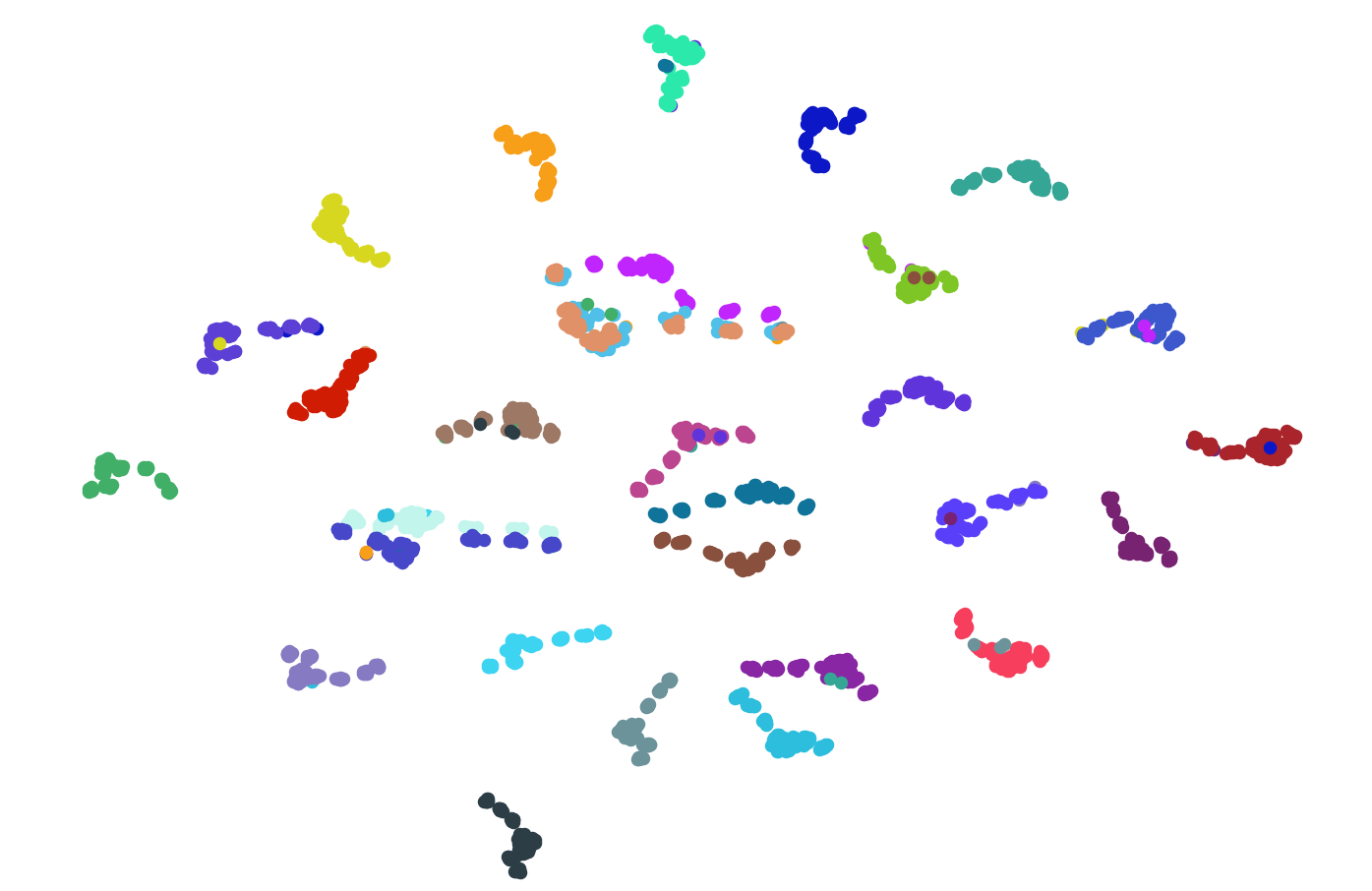}\hspace{-.3em}
    \includegraphics[width=0.23\textwidth] {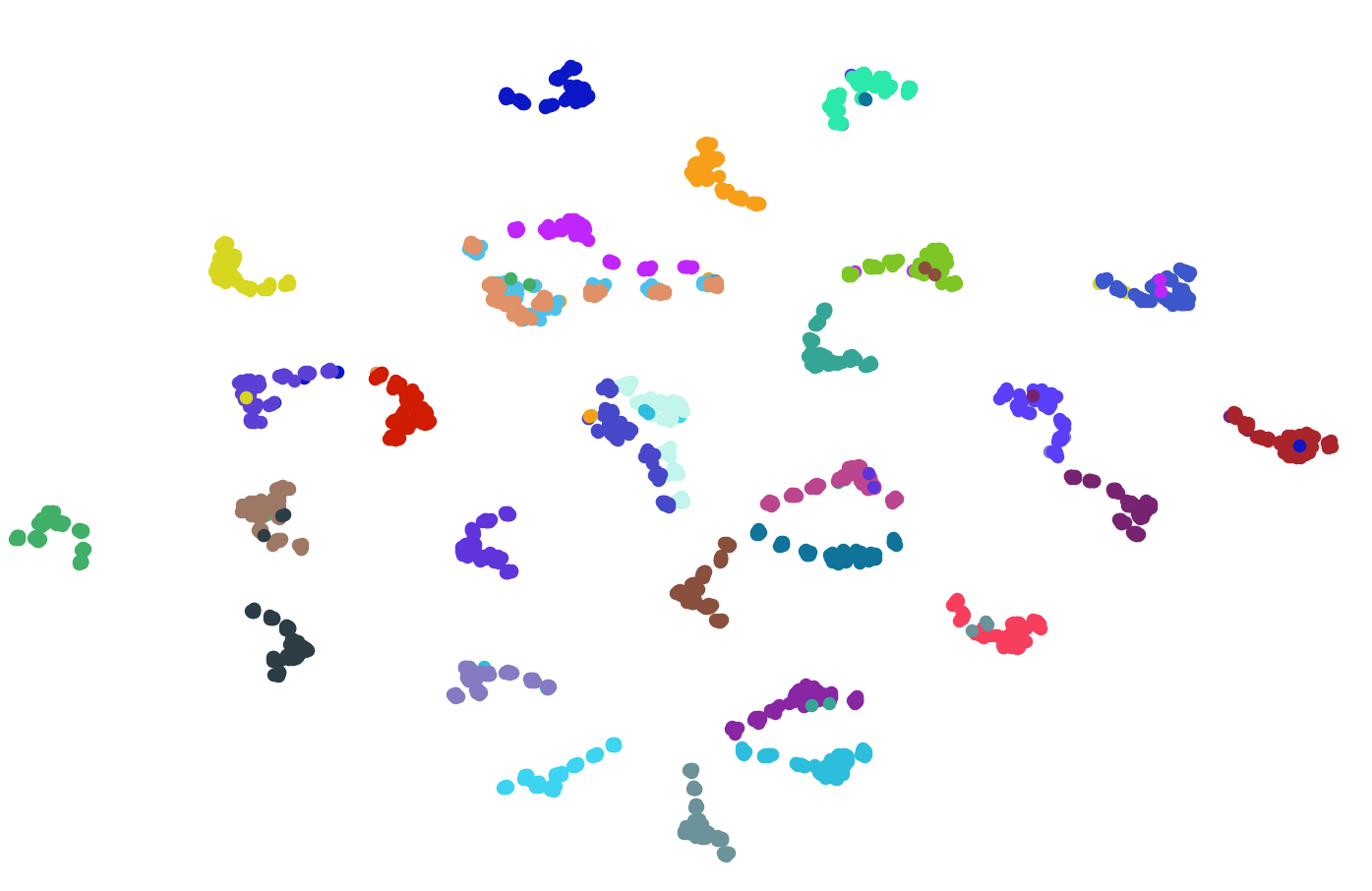}\hspace{-0.3em}
    \includegraphics[width=0.23\textwidth] {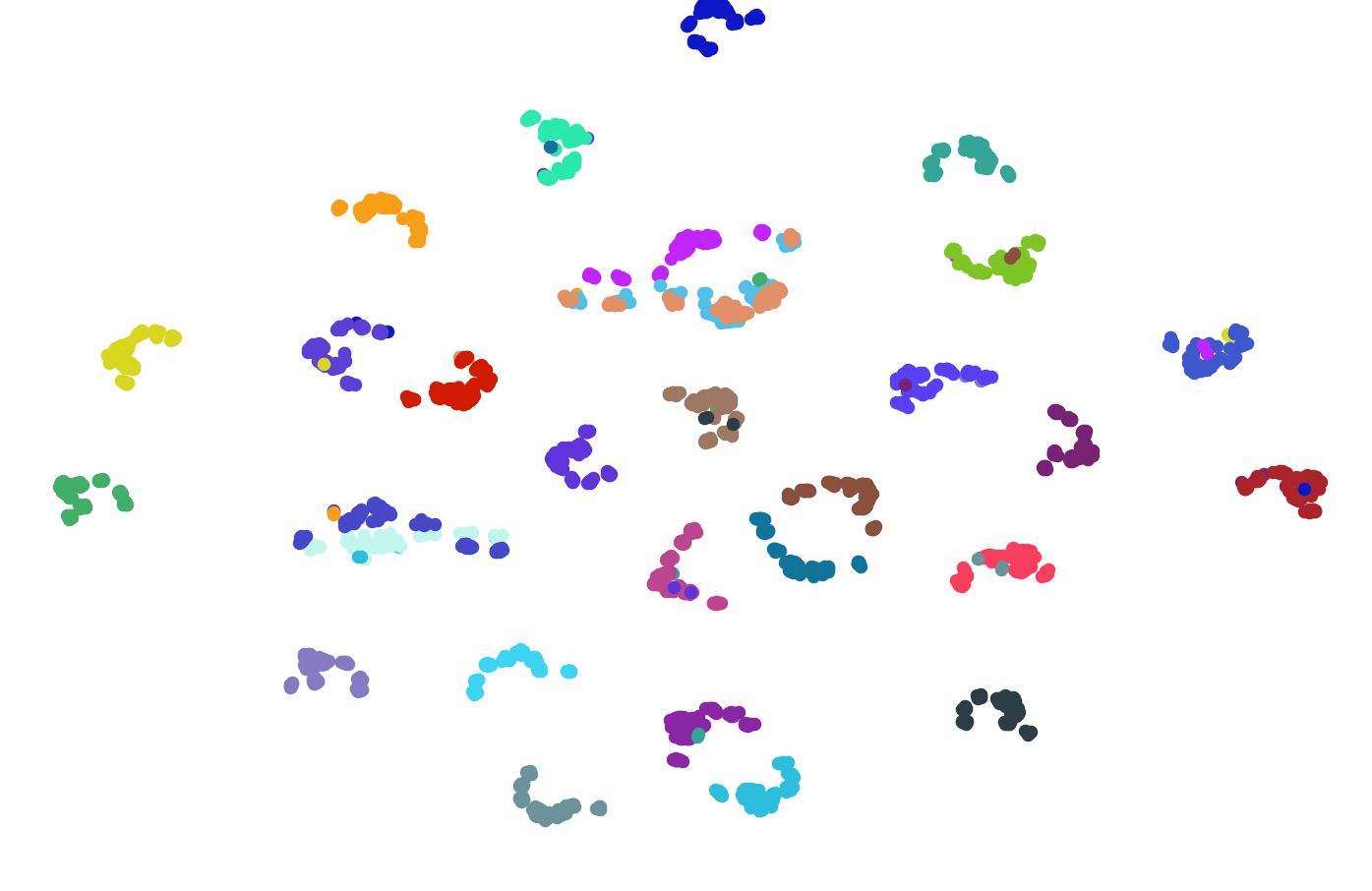}\hspace{-0.2em}
    \includegraphics[width=0.23\textwidth] {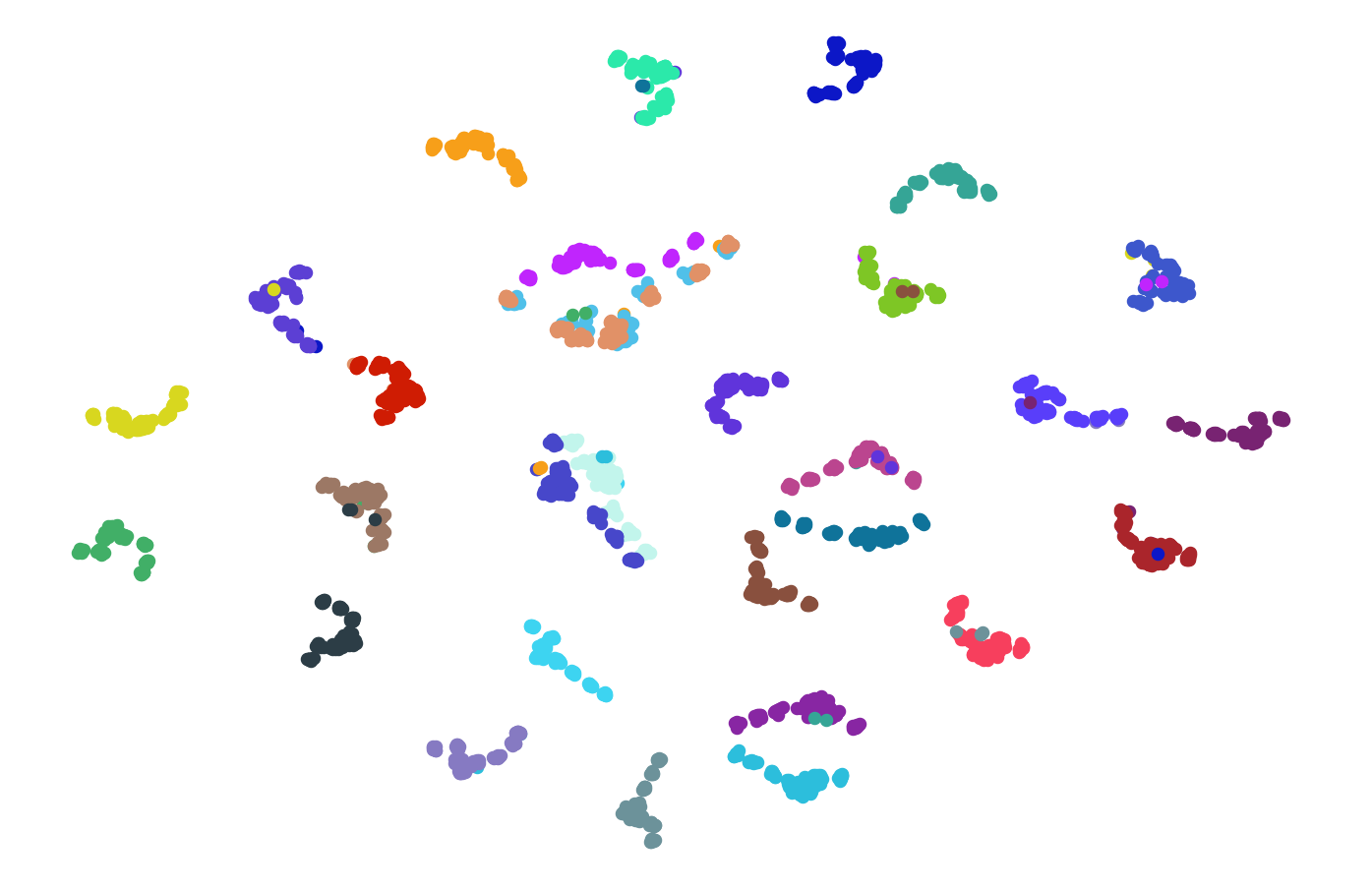}
    }\\  (b) Layer 2 (left to right): head$_1$, head$_2$, head$_3$, and their concatenation \\ 
    \subfigure{\includegraphics[width=0.26\textwidth] {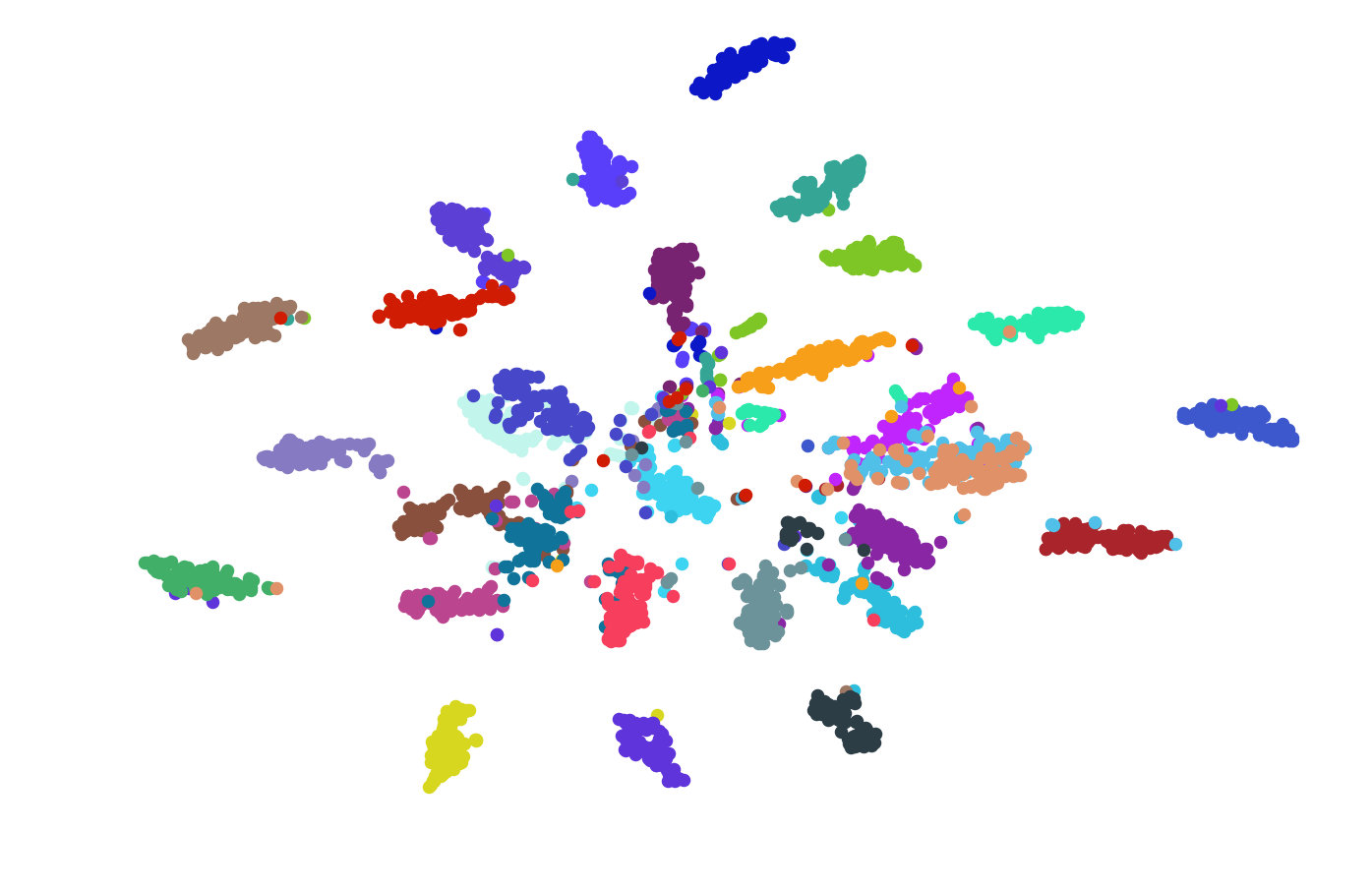}\hspace{-0.2cm}
    \includegraphics[width=0.26\textwidth] {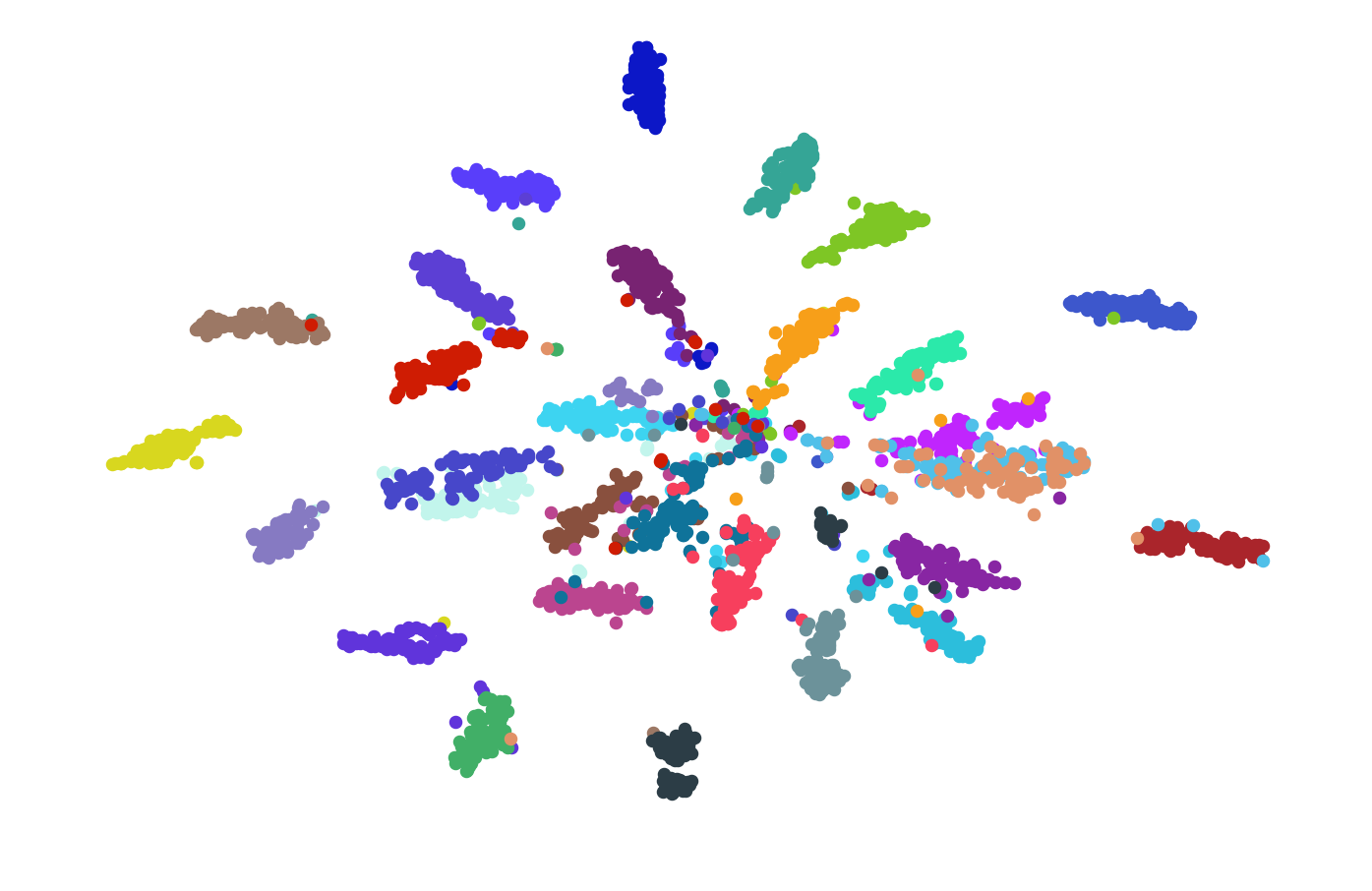}\hspace{-0.3cm}
    \includegraphics[width=0.26\textwidth] {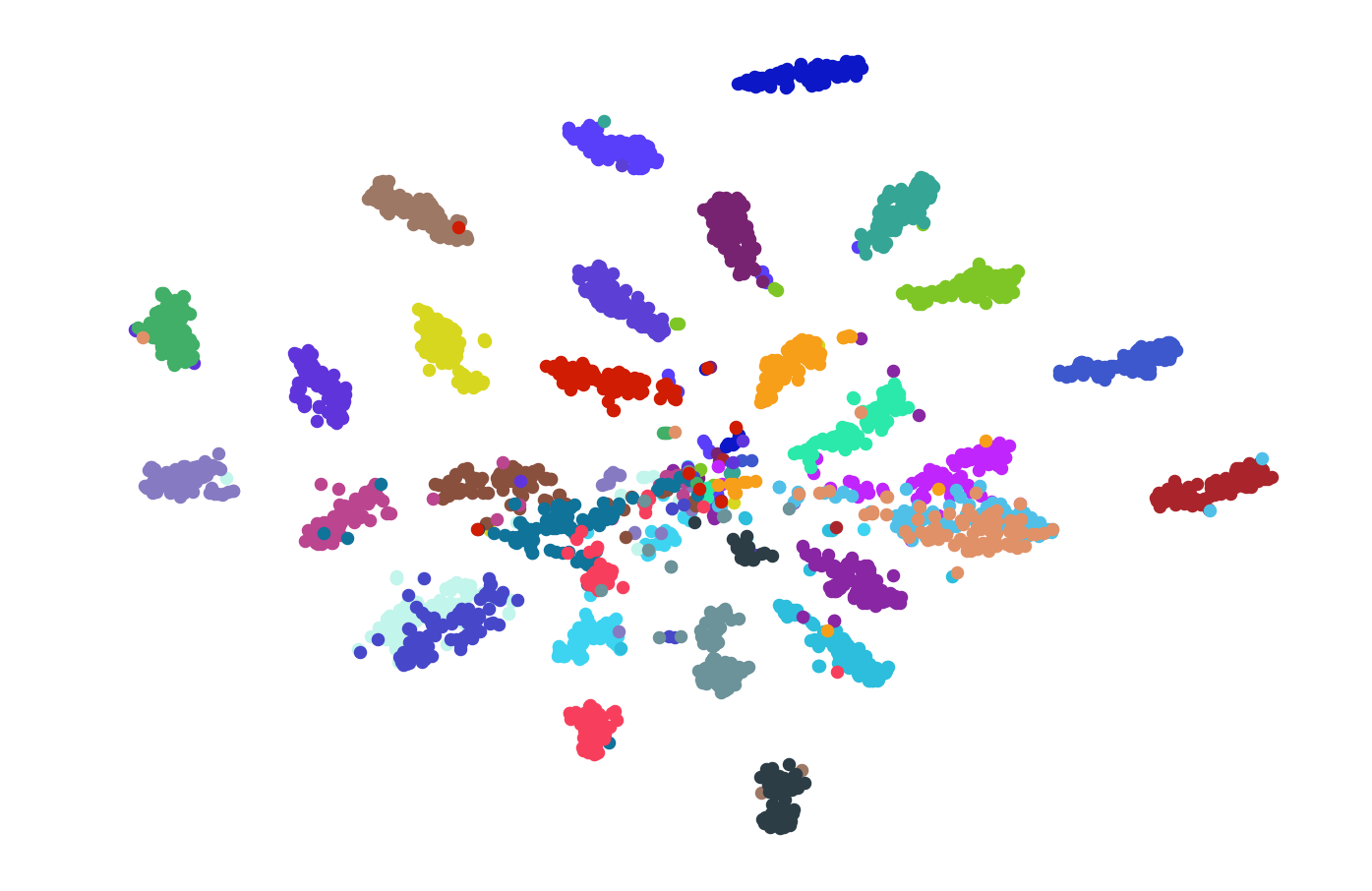}\hspace{-0.3cm}
    \includegraphics[width=0.26\textwidth] {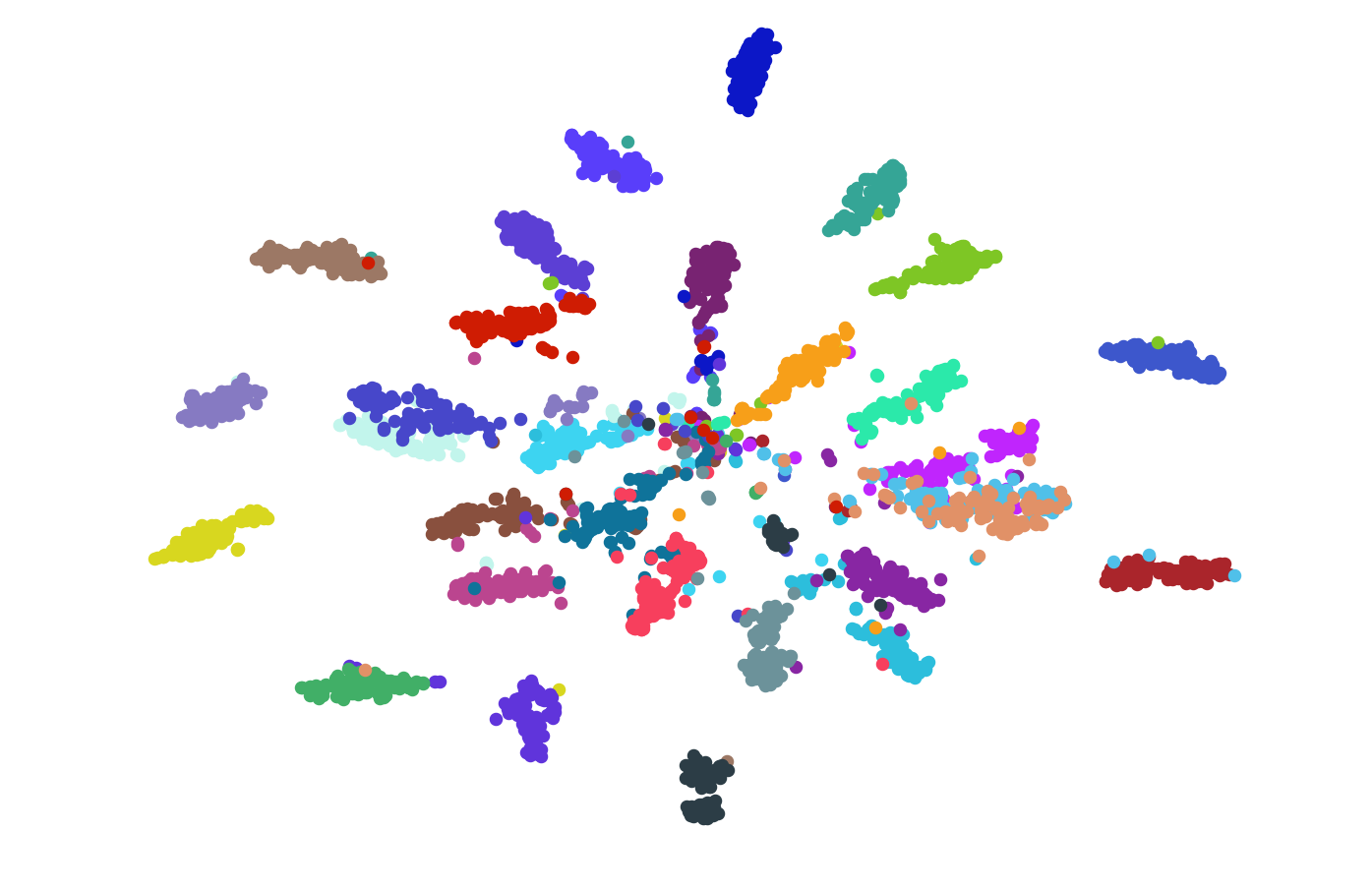}
    }\\ (c) Layer 3 (left to right): head$_1$, head$_2$, head$_3$, and their concatenation \\
    \subfigure{\includegraphics[width=0.33\textwidth] {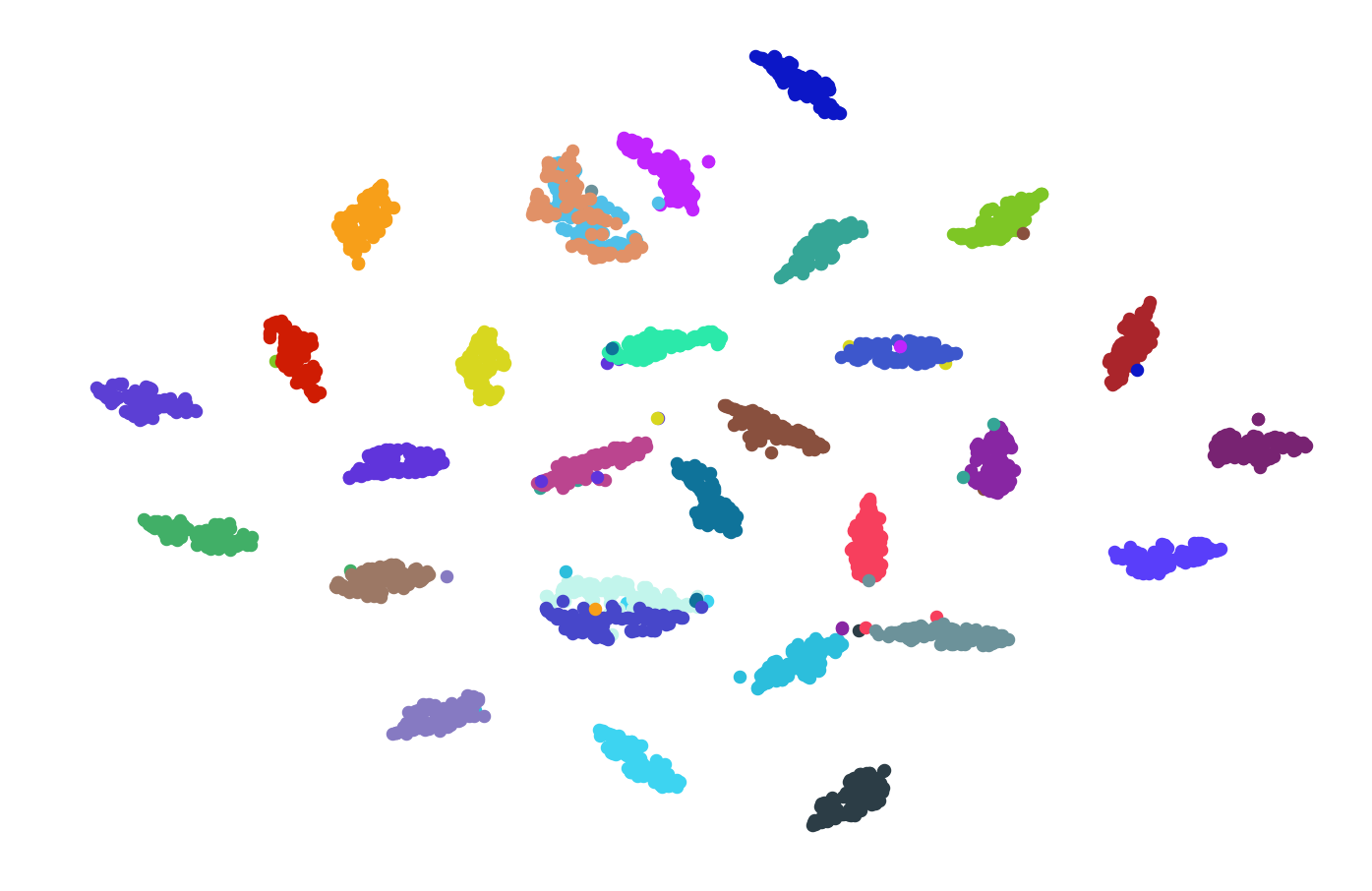}\hspace{-0.5em}
    \includegraphics[width=0.33\textwidth] {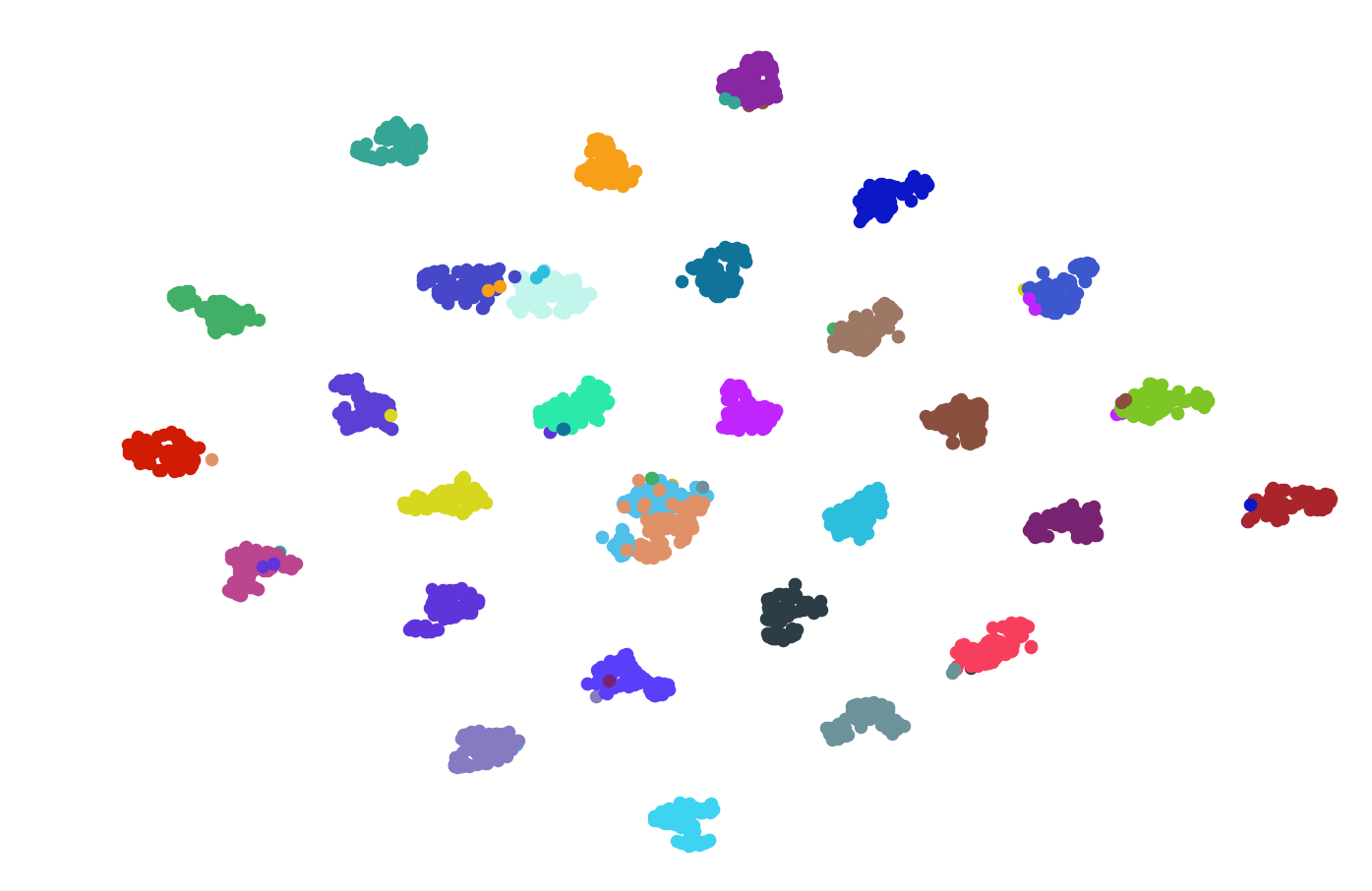}\hspace{-0.5em}
    \includegraphics[width=0.33\textwidth] {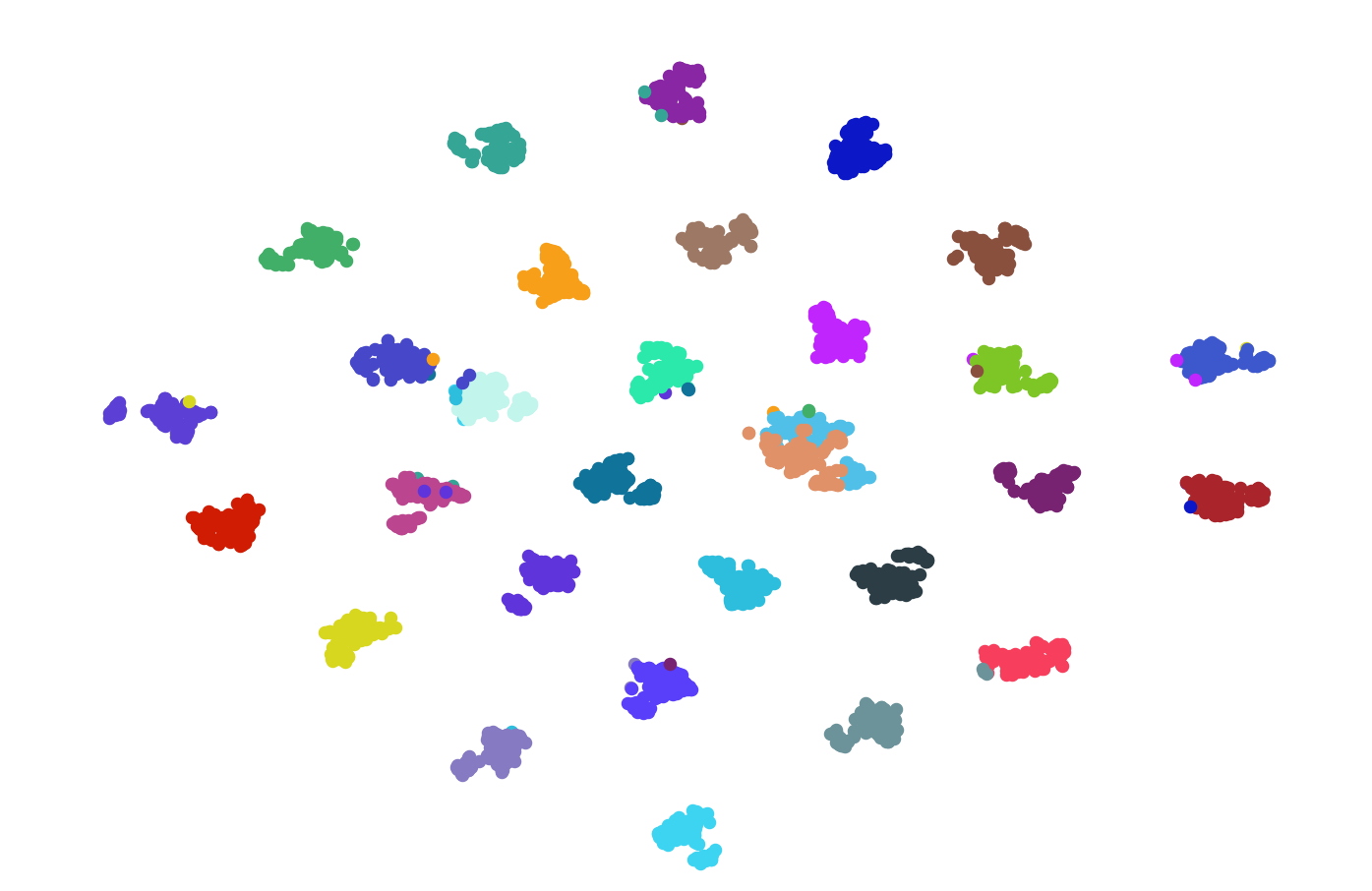}
    } \\ (d) Final representation (left to right): using 2, 3 and 4 attention heads \\
    \caption{t-SNE [{\color{red}50}] visualization of class-specific discriminative feature representation of multi-scale hierarchical regions 
    using $H$=3 attention heads in ({\color{red}2}), and $L$=3 layers hierarchical structure in ({\color{red}1}). All test images from 30 randomly chosen classes within \textbf{Oxford-IIIT Pets-37} dataset are used. 
    Attention head-specific plots are shown in $(a)\rightarrow (c)$, representing layers from smaller regions (a) to larger ones (c). It is evident that the discriminability of the features representing medium-size regions (b) $>$ small-size (a) $>$ large-size (c). 
    (d) shows the combined layers' representation using 2, 3 and 4 attention heads. More than 2 attention heads has shown better discriminability.}
    \label{fig:fig_pets}
    \vspace{-.5cm}
\end{figure*}
\begin{figure*}[t]
\centering
    \subfigure{\includegraphics[width=0.25\textwidth] {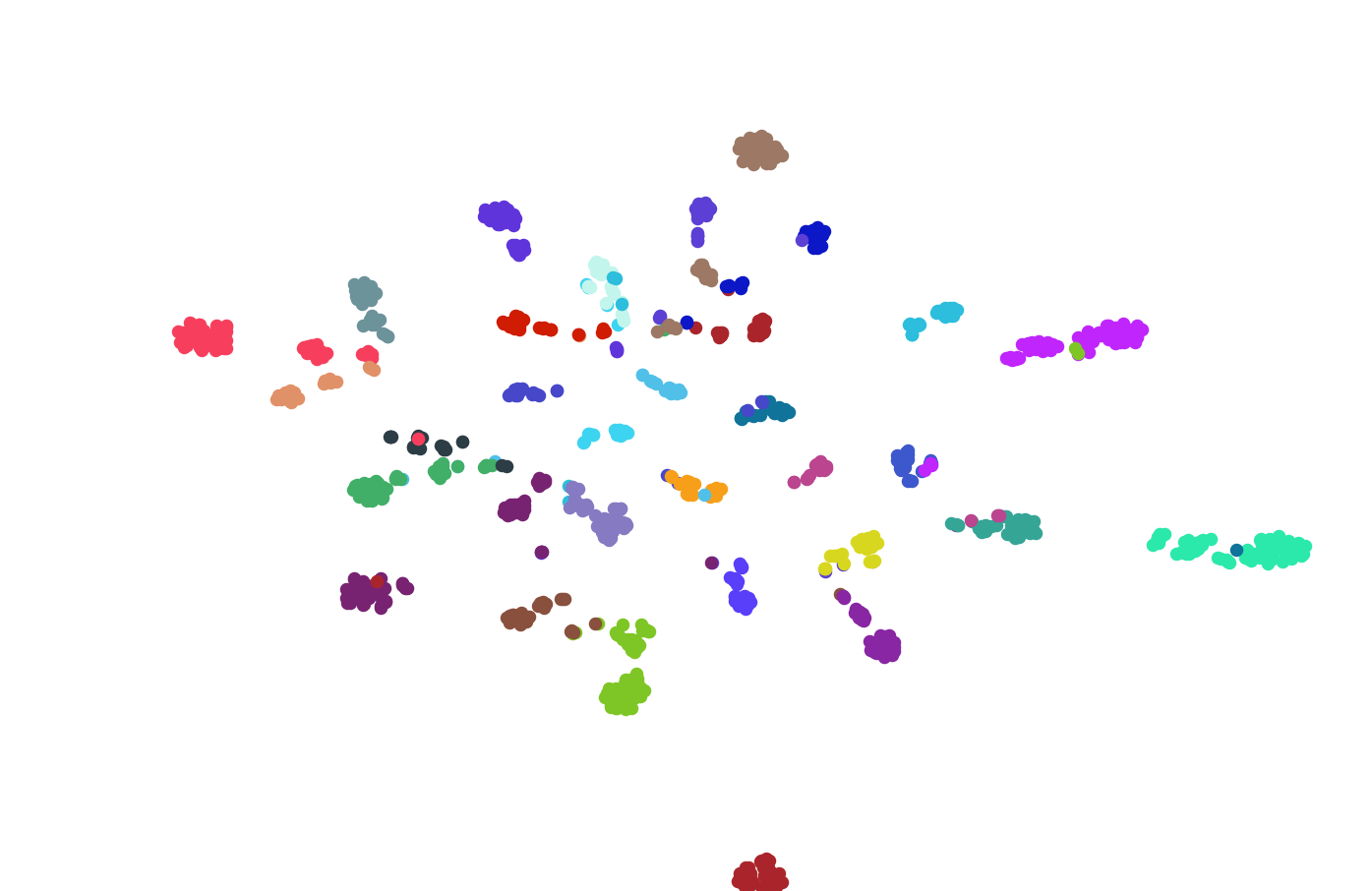}\hspace{-1em}
    \includegraphics[width=0.25\textwidth] {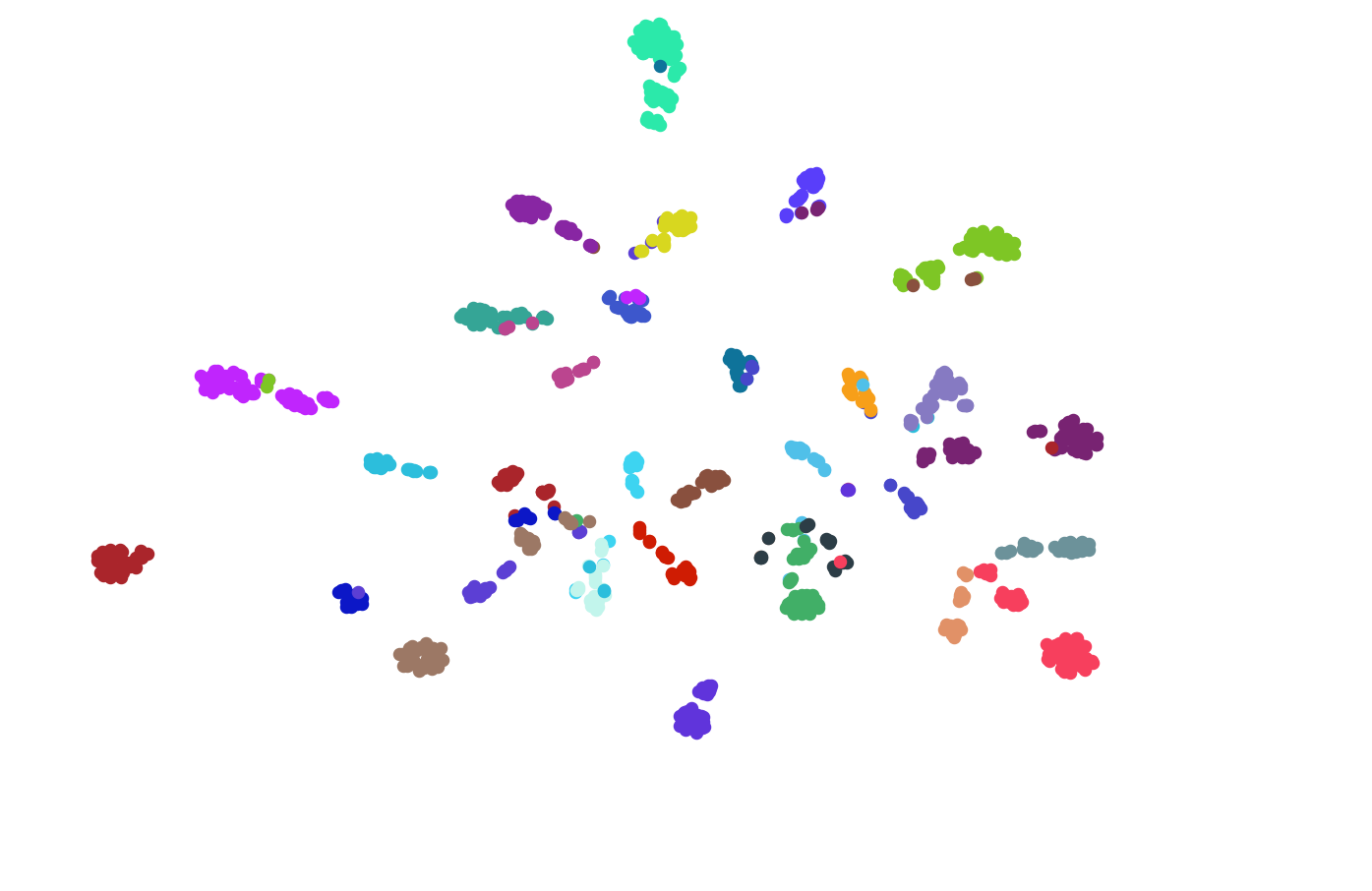}\hspace{-1em}
    \includegraphics[width=0.25\textwidth] {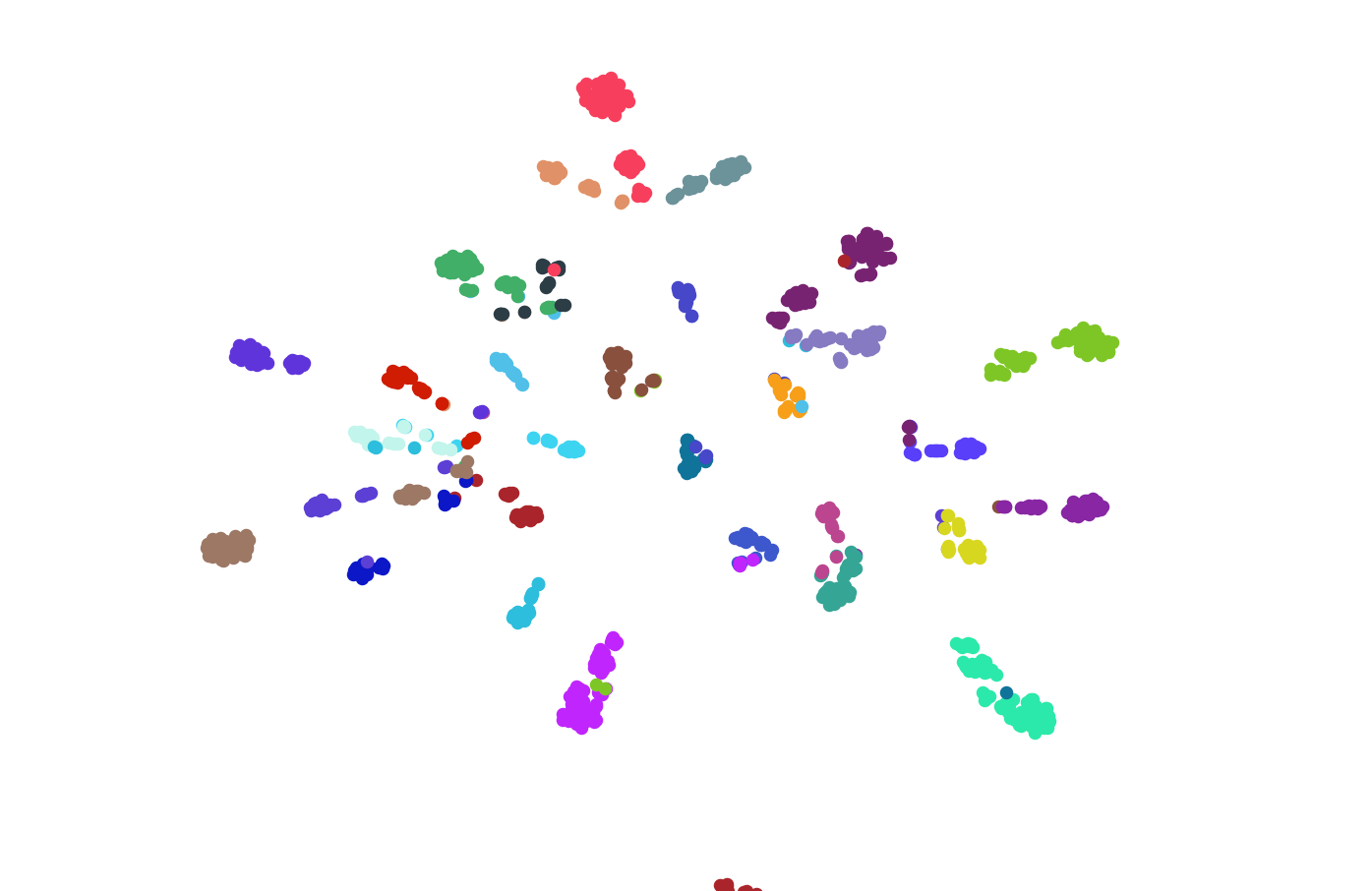}\hspace{-2.5em}
    \includegraphics[width=0.25\textwidth] {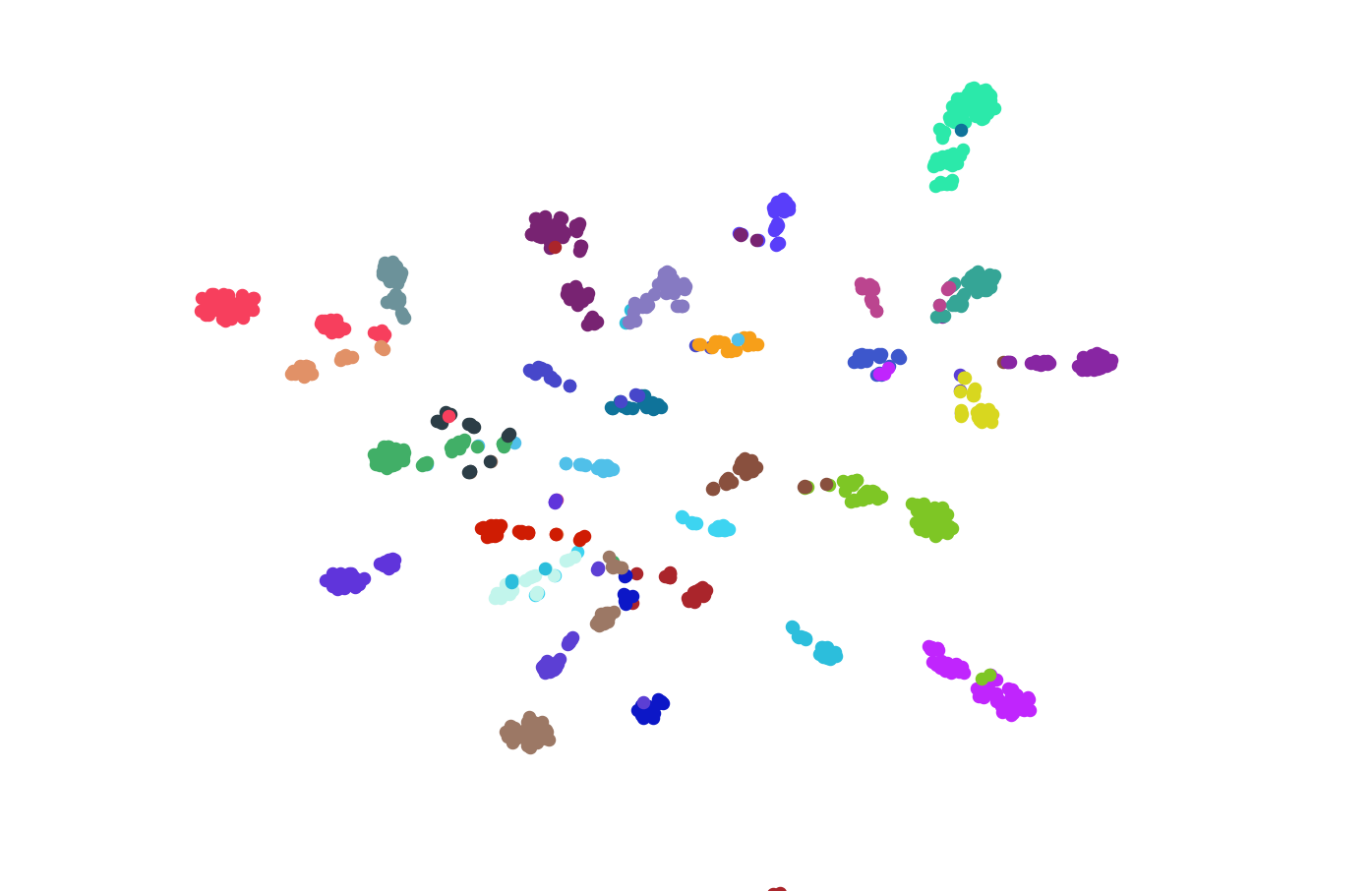}
    }\\ (a) Layer 1 (left to right): head$_1$, head$_2$, head$_3$, and their concatenation  \\
     \subfigure{\includegraphics[width=0.23\textwidth] {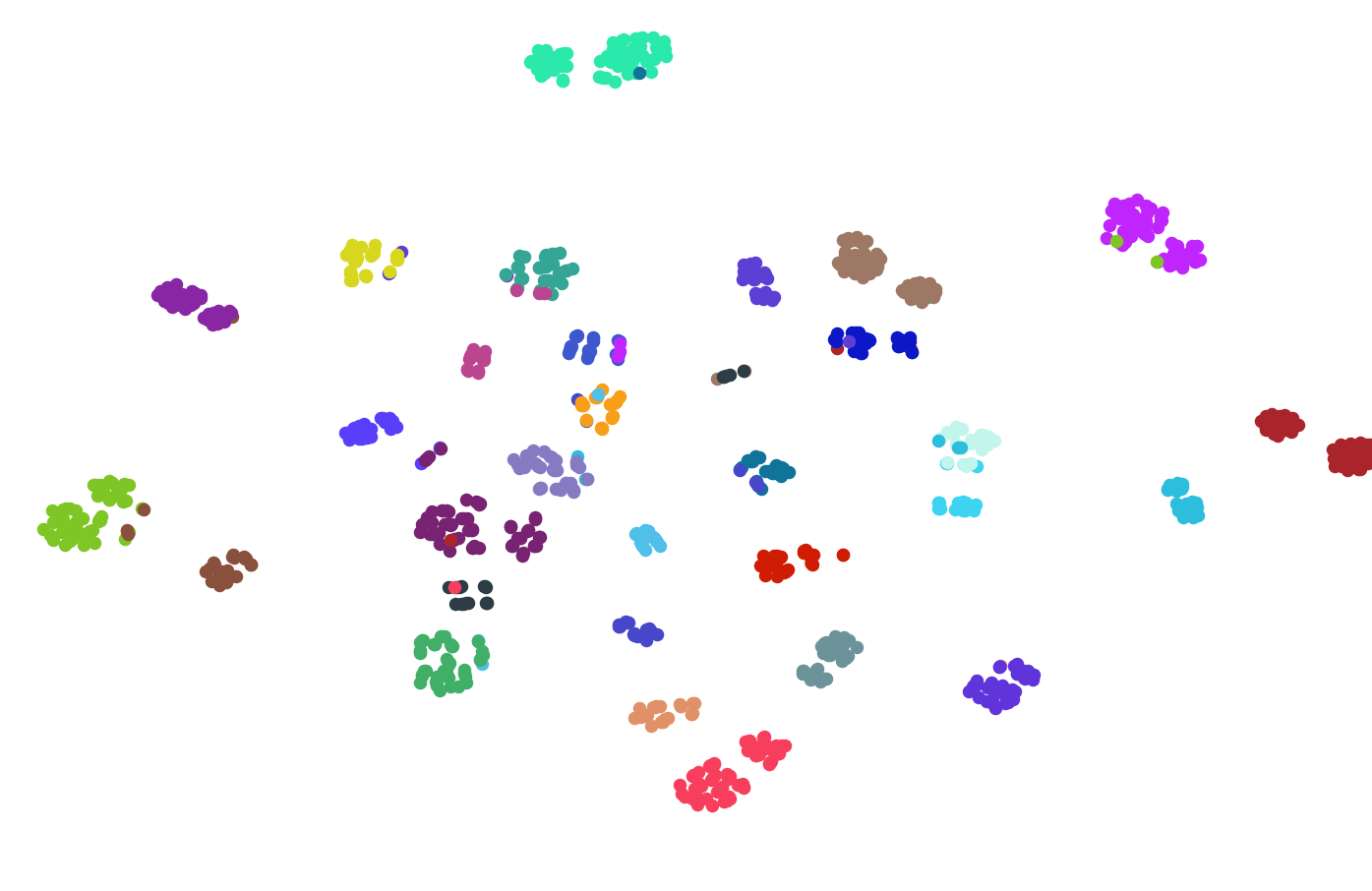}\hspace{-.3em}
    \includegraphics[width=0.23\textwidth] {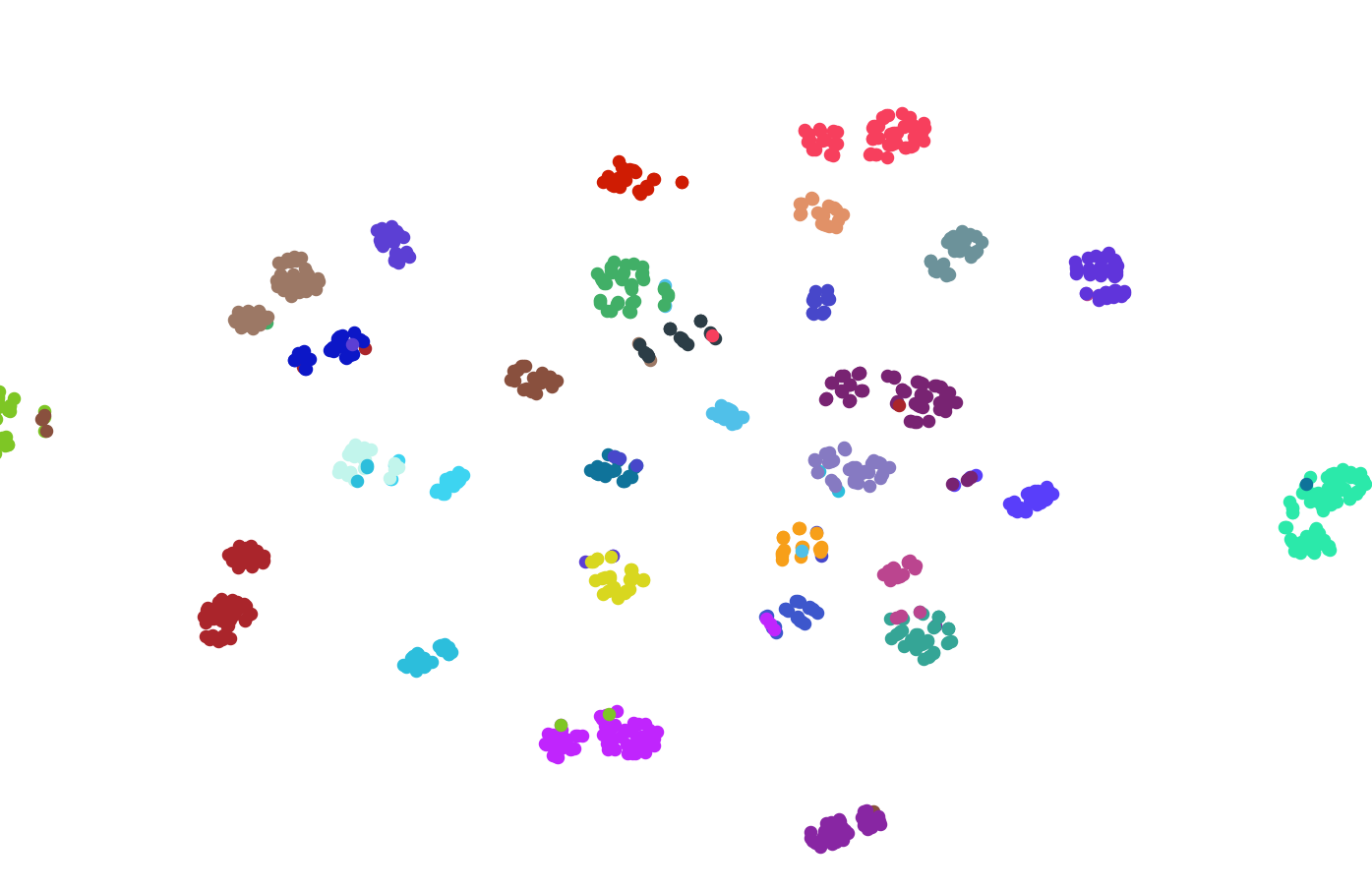}\hspace{-1.5em}
    \includegraphics[width=0.23\textwidth] {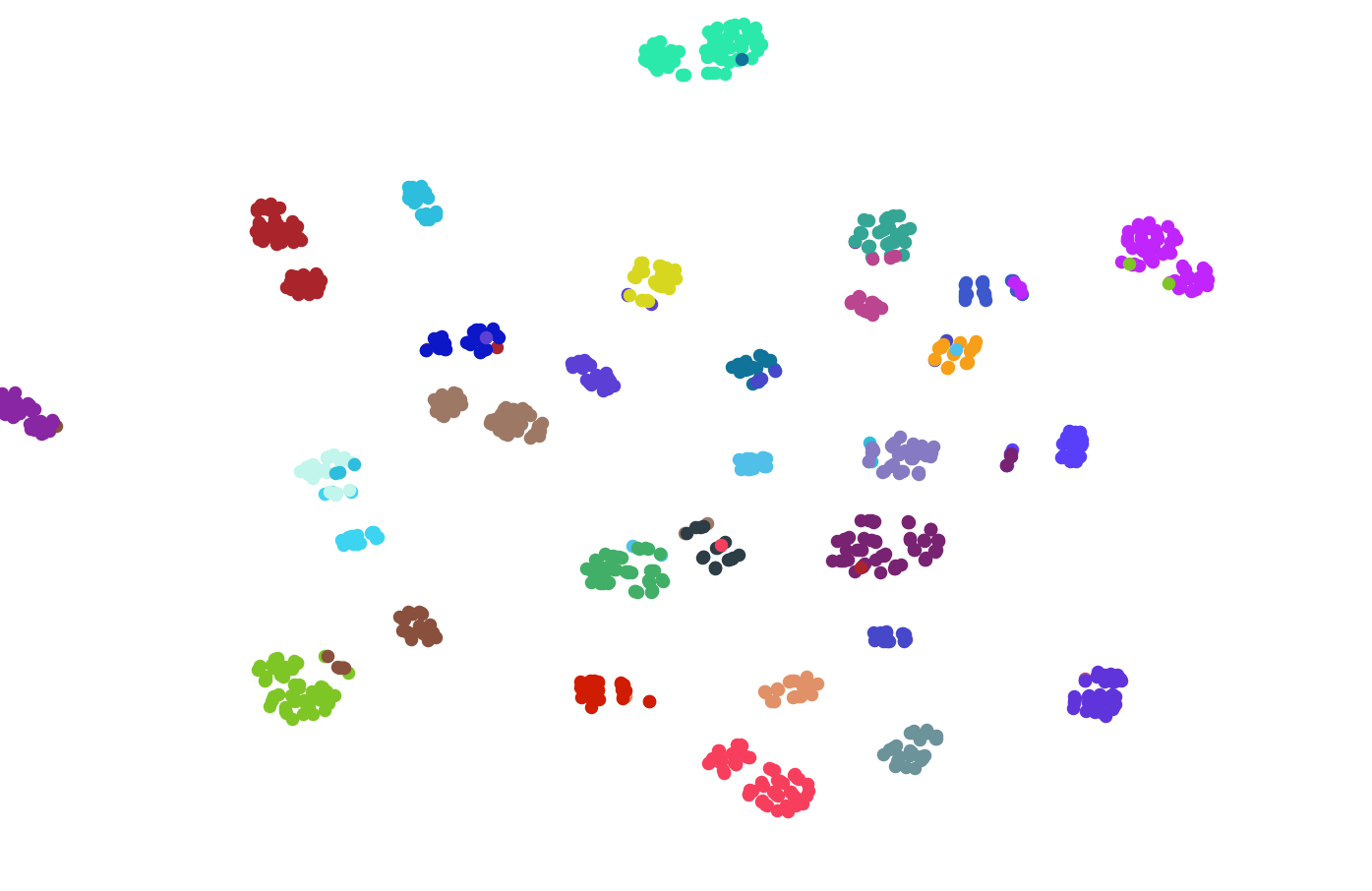}\hspace{-2em}
    \includegraphics[width=0.23\textwidth] {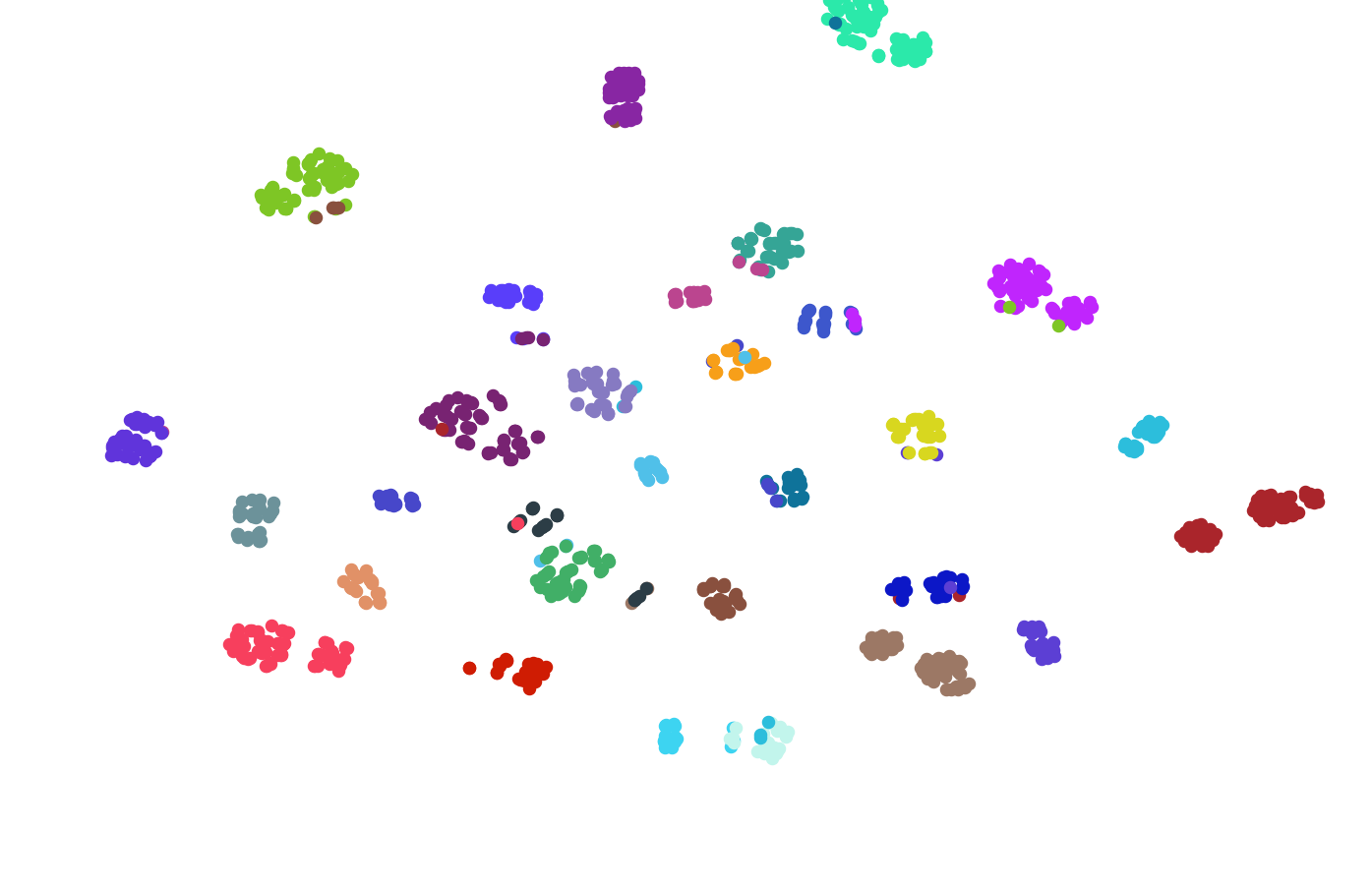}
    }\\ (b)  Layer 2 (left to right): head$_1$, head$_2$, head$_3$, and their concatenation \\ 
    \subfigure{\includegraphics[width=0.3\textwidth] {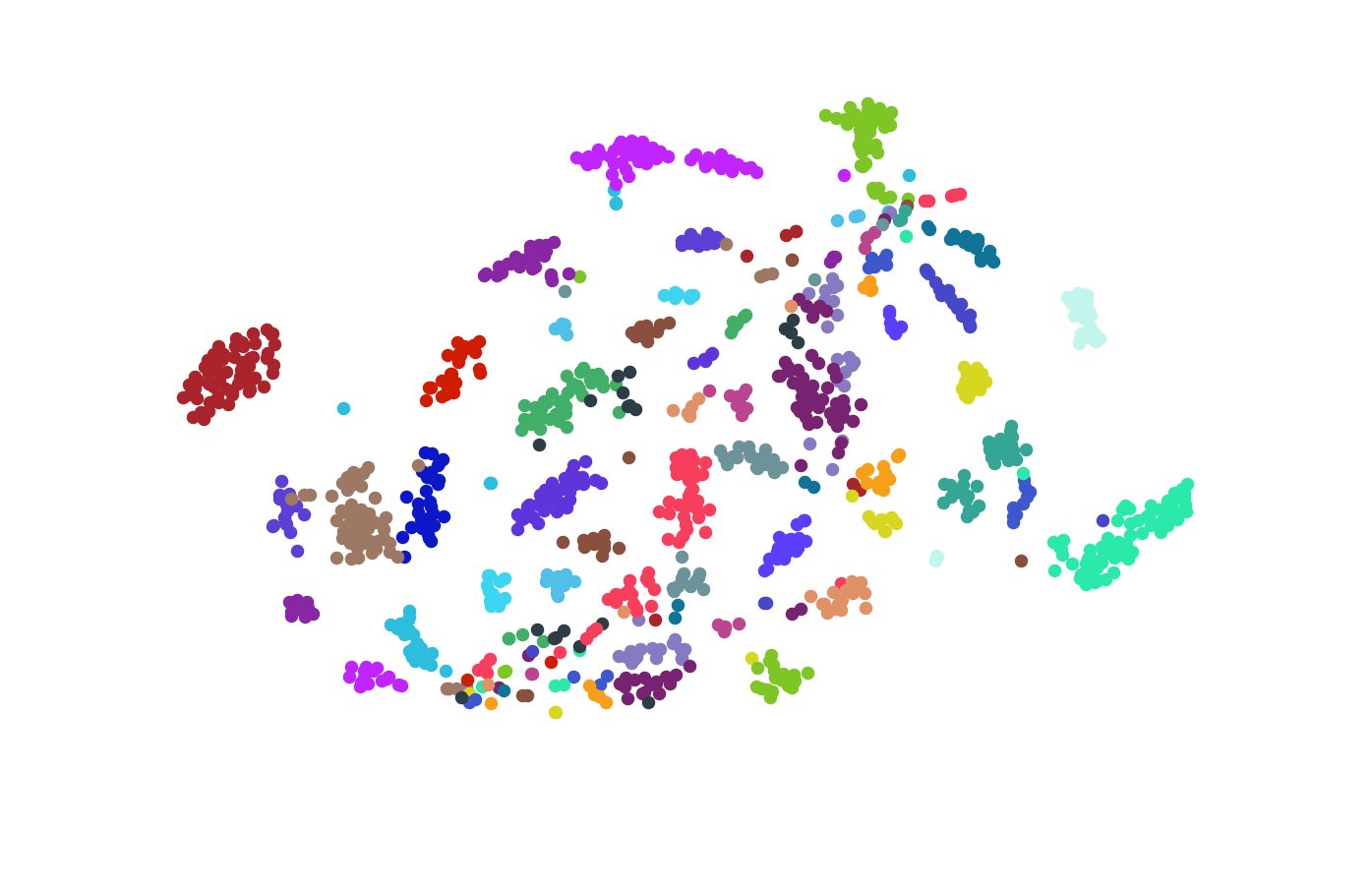}\hspace{-1.1cm}
    \includegraphics[width=0.3\textwidth] {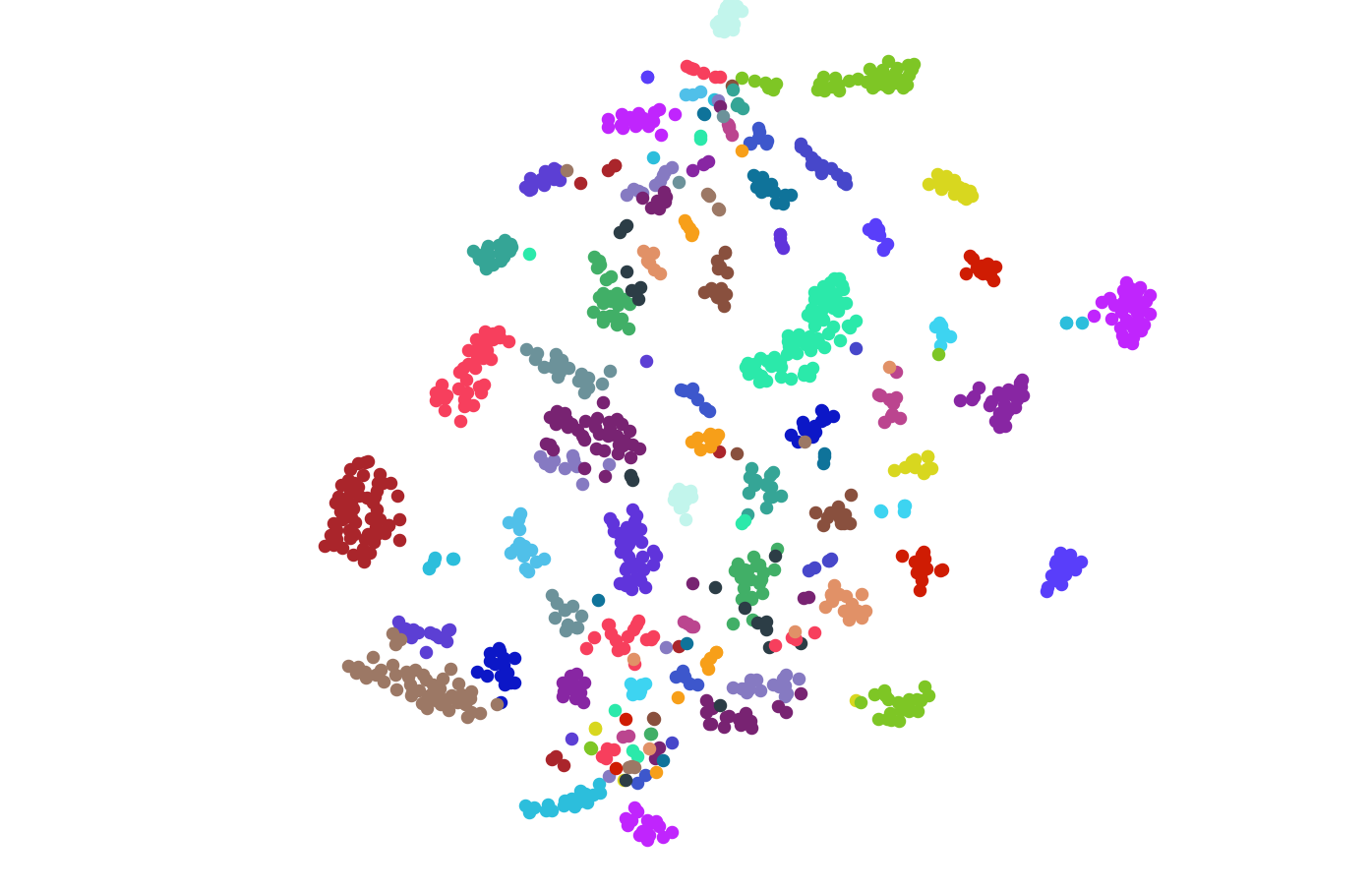}\hspace{-1.2cm}
    \includegraphics[width=0.3\textwidth] {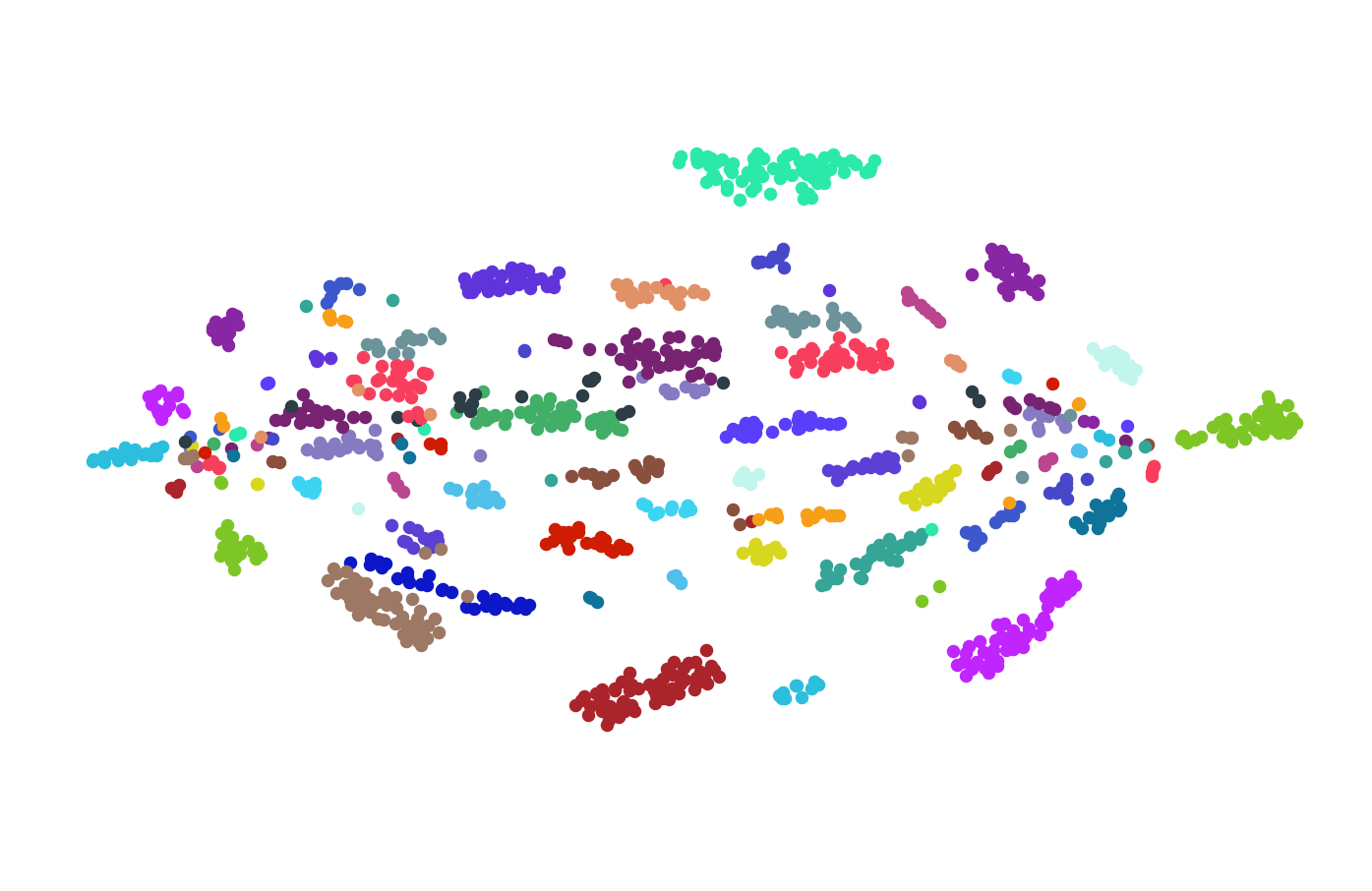}\hspace{-0.8cm}
    \includegraphics[width=0.3\textwidth] {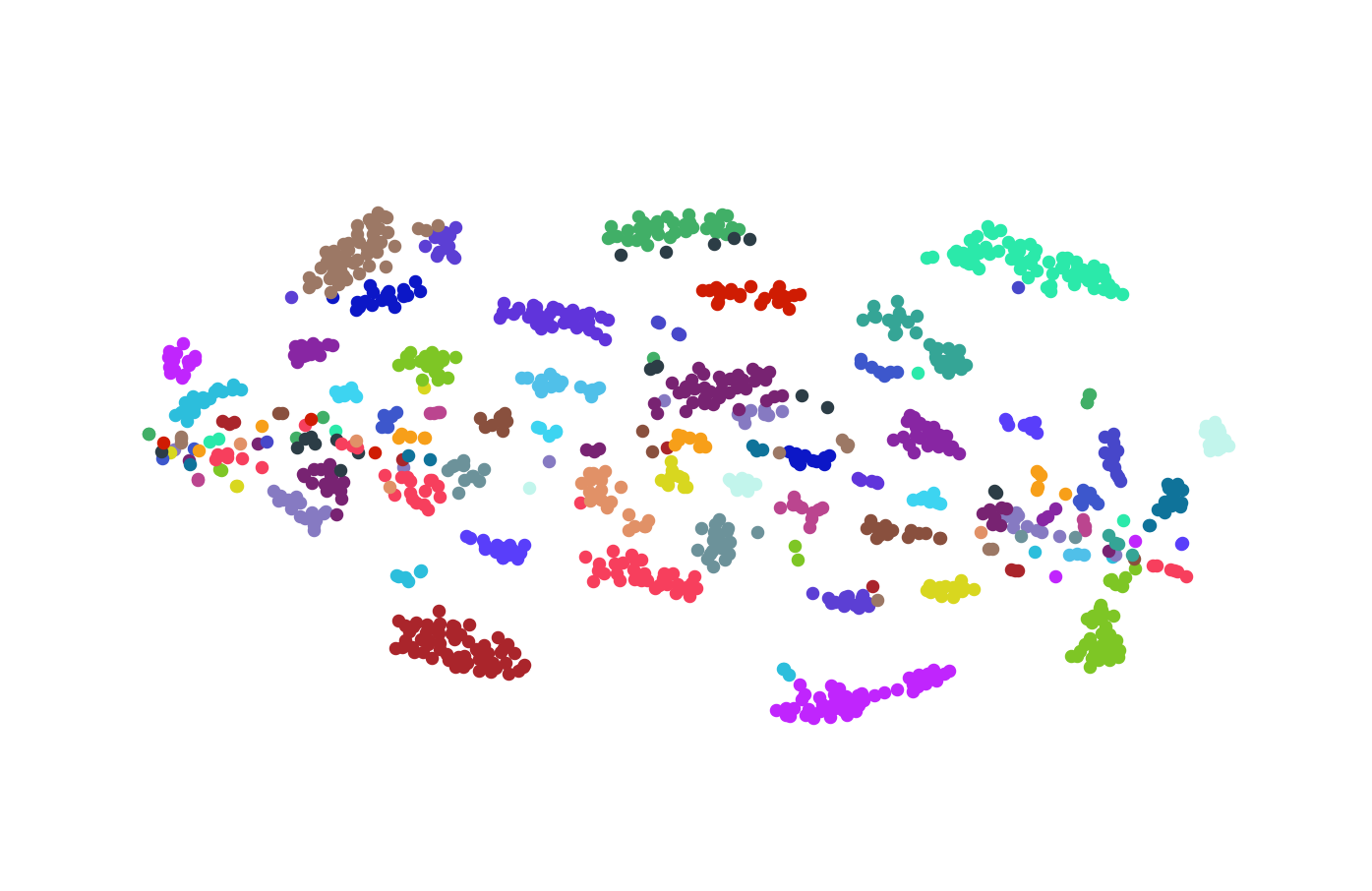}
    }\\ (c) Layer 3 (left to right): head$_1$, head$_2$, head$_3$, and their concatenation \\
    \subfigure{\includegraphics[width=0.3\textwidth] {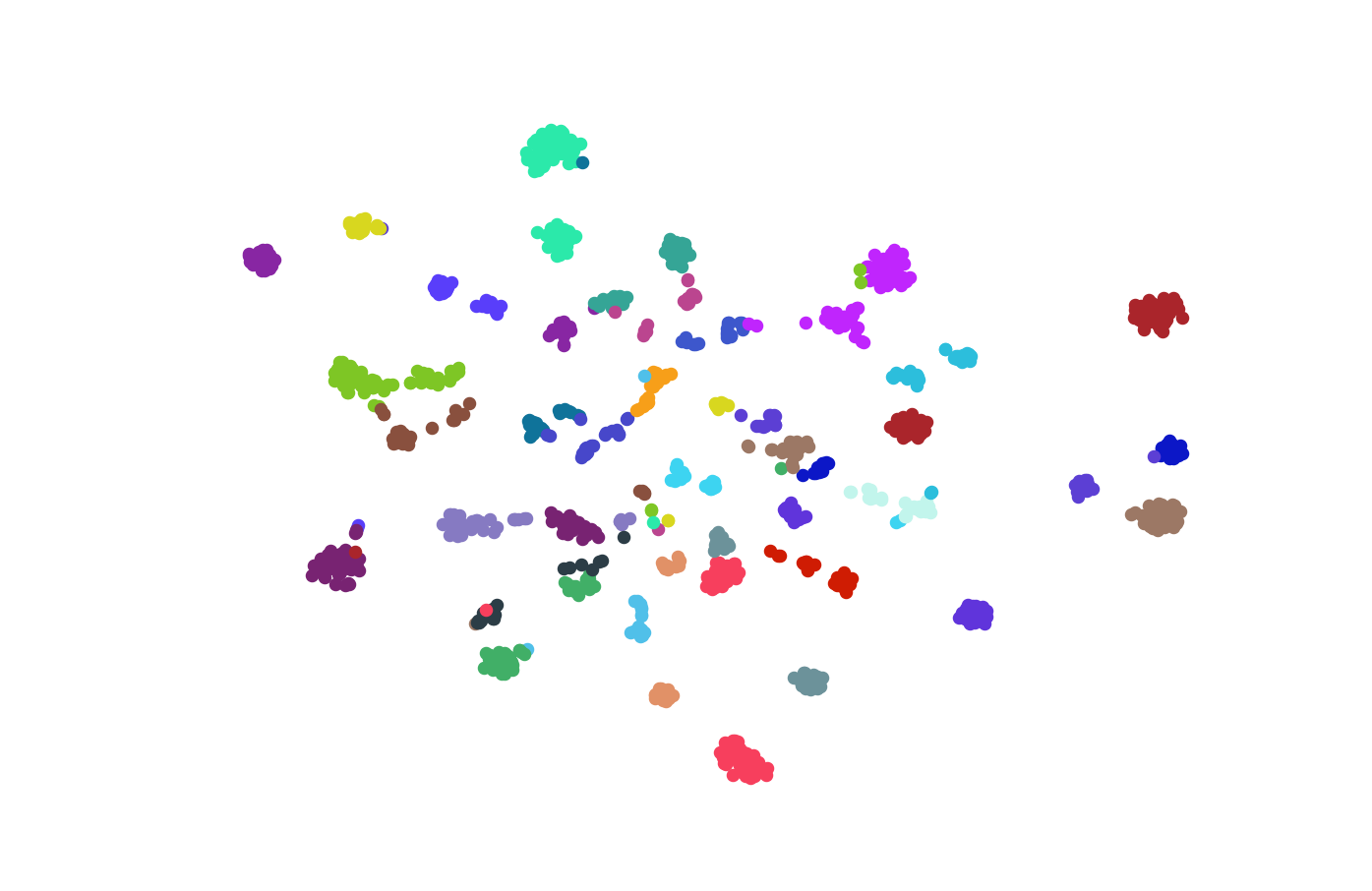}\hspace{-0.5em}
    \includegraphics[width=0.3\textwidth] {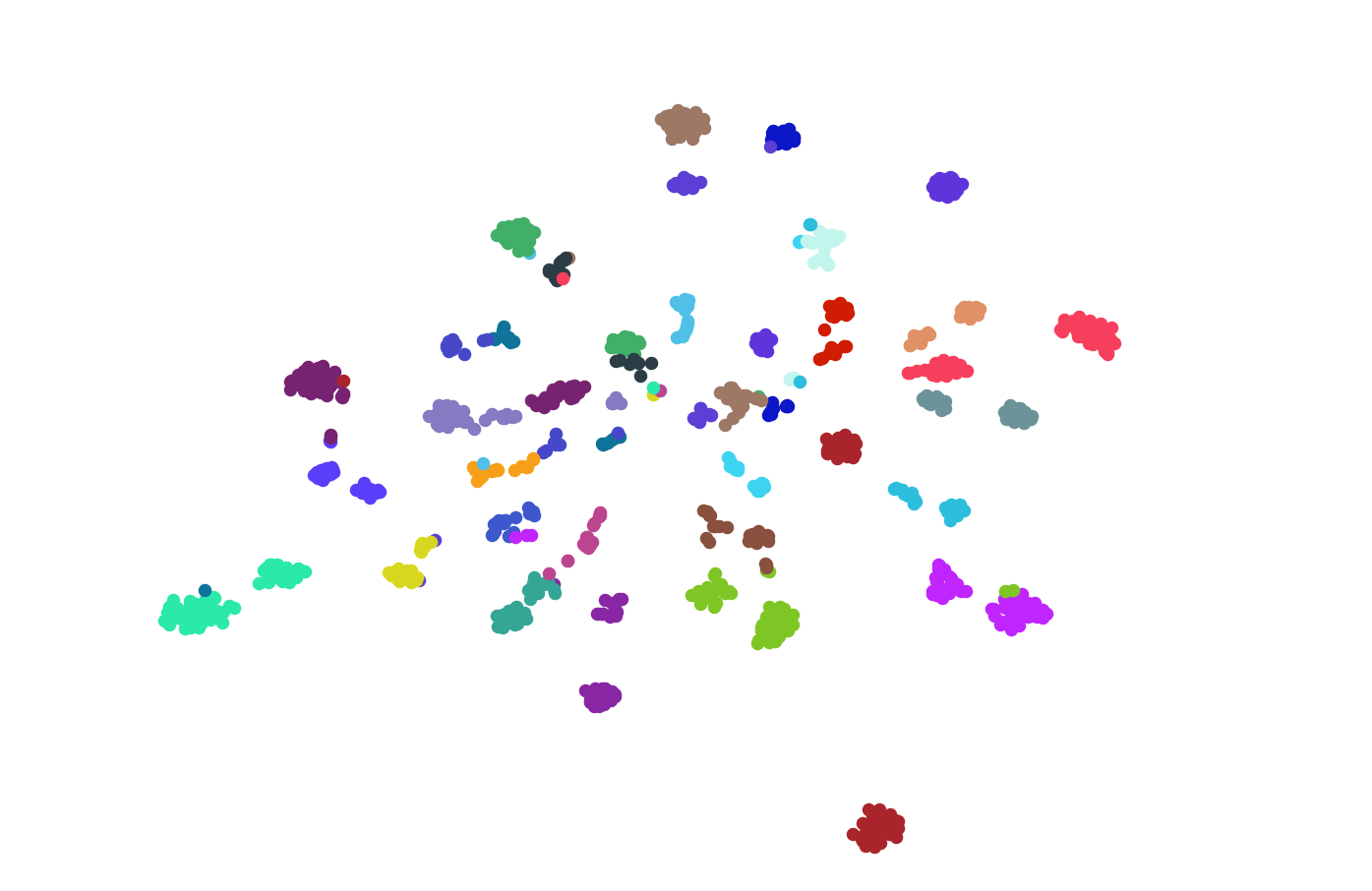}\hspace{-0.5em}
    \includegraphics[width=0.3\textwidth] {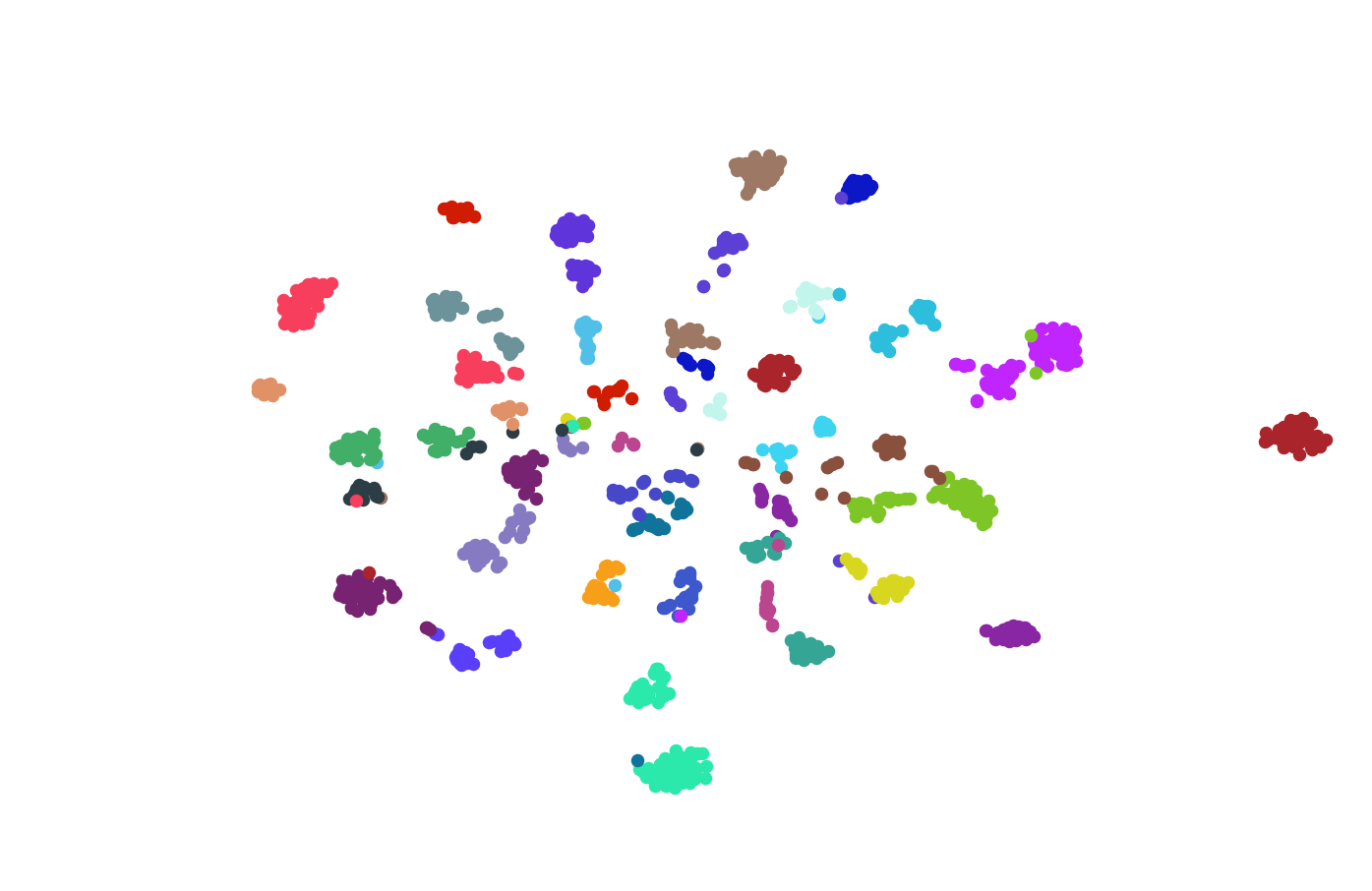}
    } \\ (d) Final representation (left to right): using 2, 3 and 4 attention heads \\
    \caption{t-SNE [{\color{red}50}] visualization of class-specific discriminative feature representation of multi-scale hierarchical regions 
    using $H$=3 attention heads in ({\color{red}2}), and $L$=3 layers hierarchical structure in ({\color{red}1}). All test images from 30 randomly chosen classes within \textbf{Oxford-Flowers-102} dataset are used. 
    Attention head-specific plots are shown in $(a)\rightarrow (c)$, representing layers from smaller regions (a) to larger ones (c). It is evident that the discriminability of the features representing medium-size regions (b) $>$ small-size (a) $>$ large-size (c). 
    (d) shows the combined layers' representation using 2, 3 and 4 attention heads. More than 2 attention heads has shown better discriminability.}
    \label{fig:fig_pets}
    \vspace{-.5cm}
\end{figure*}


\end{document}